%% file: main.tex
\documentclass[runningheads]{llncs}

\usepackage{eccv}

\usepackage{eccvabbrv}

\usepackage{graphicx}
\usepackage{booktabs}

\usepackage[accsupp]{axessibility}  

\usepackage[pagebackref,breaklinks,colorlinks,citecolor=eccvblue]{hyperref}

\usepackage{orcidlink}

\usepackage[table]{xcolor}
\usepackage{pifont}
\usepackage{makecell}

\newcommand{\cmark}{\color{teal}{\ding{51}}}
\newcommand{\xmark}{\color{red}{\ding{55}}}
\usepackage{caption}
\usepackage{tabularx}
\usepackage{arydshln}
\usepackage{adjustbox} 
\usepackage{multirow}
\usepackage{wrapfig}

\renewcommand{\paragraph}[1]{\vspace{.5em}\noindent\textbf{#1 }}

\begin{document}

\title{MA-EgoQA: Question Answering over Egocentric Videos from Multiple Embodied Agents} 

\titlerunning{MA-EgoQA}

\author{Kangsan Kim\inst{1} \and
Yanlai Yang\inst{2} \and
Suji Kim\inst{1,3} \and
Woongyeong Yeo\inst{1} \and \\
Youngwan Lee\inst{1,4} \and
Mengye Ren\inst{2}$^\dagger$ \and
Sung Ju Hwang\inst{1,5}$^\dagger$}

\renewcommand\thefootnote{$\dagger$}
\footnotetext{Equal advising.}

\authorrunning{K. Kim et al.}


\institute{KAIST~~ \and
New York University \\ \and
Samsung Electronics~~ \and
ETRI~~ \and
DeepAuto.ai\\
\email{kangsan.kim@kaist.ac.kr}}

\maketitle

\input{sec/0_abstract}
\input{sec/1_introduction}
\input{sec/2_related_work}
\input{sec/3_ma-egoqa}
\input{sec/4_benchmark_construction}
\input{sec/5_egomas}
\input{sec/6_experiment}
\input{sec/7_analysis}

\input{sec/8_conclusion}



%
%
\bibliographystyle{splncs04}
\bibliography{main}

\input{sec/X_appendix}
\end{document}

%% file: sec/0_abstract.tex
\begin{abstract}
As embodied models become powerful, humans will collaborate with multiple embodied AI agents at their workplace or home in the future. To ensure better communication between human users and the multi-agent system, it is crucial to interpret incoming information from agents in parallel and refer to the appropriate context for each query. Existing challenges include effectively compressing and communicating high volumes of individual sensory inputs in the form of video and correctly aggregating multiple egocentric videos to construct system-level memory. In this work, we first formally define a novel problem of understanding multiple long-horizon egocentric videos simultaneously collected from embodied agents. To facilitate research in this direction, we introduce MultiAgent-EgoQA (MA-EgoQA), a benchmark designed to systemically evaluate existing models in our scenario. MA-EgoQA provides 1.7k questions unique to multiple egocentric streams, spanning five categories: social interaction, task coordination, theory-of-mind, temporal reasoning, and environmental interaction. We further propose a simple baseline model for MA-EgoQA named EgoMAS, which leverages shared memory across embodied agents and agent-wise dynamic retrieval. Through comprehensive evaluation across diverse baselines and EgoMAS on MA-EgoQA, we find that current approaches are unable to effectively handle multiple egocentric streams, highlighting the need for future advances in system-level understanding across the agents. The code and benchmark are available at \url{https://ma-egoqa.github.io}.

\end{abstract}

%% file: sec/1_introduction.tex
\section{Introduction}
\label{sec:intro}

Recent advances are increasingly transferring progress from text-based environments to physical embodied domains, such as autonomous driving~\cite{fu2024drive, huang2024drivlme} and industrial robotics~\cite{zhao2025cot, kannan2024smart}. Just as large language models (LLMs) like ChatGPT have reshaped how humans interact with and reason over knowledge, we can expect a future where intelligent agents increasingly assist and augment our physical activities.
As these agents become common and multiple agents begin to operate within a shared environment, a growing line of research explores how they can interact and collaborate effectively as a system~\cite{Multi-agent-embodied-ai}. Multi-agent system (MAS) can achieve the goals efficiently by decomposing and executing tasks in parallel, and collective reasoning enables agents to leverage broader context and arrive at more optimal solutions~\cite{chen2024emos, hu2022where2comm}.
At the same time, multi-agent settings pose unique challenges, such as designing robust communication protocols and developing strategies for sub-task generation and allocation. Addressing these challenges is essential for developing embodied agent systems that can operate at scale and effectively handle the complexities of real-world environments.

\input{fig/relwork_comparison}

However, contextual understanding and question answering (QA) remain largely underexplored in the study of multiple embodied agents. Prior works have predominantly focused on goal-directed algorithms, task allocation, and completion, particularly within the robotics domain~\cite{CoELA, remac, guo2024embodied}. While these directions are critical for system operation, effective actions ultimately depend on accurately interpreting event histories, environmental state, and inter-agent communication. This capability is especially critical in QA scenarios, such as when a human manager queries the system to monitor progress, as shown in \cref{fig:concept}. Importantly, 
the system must be able to integrate experiences across agents and retrieve relevant episodes to provide a correct answer. Such functionality has broad application in the real world, for example, querying anomalies across multiple bodycam feeds from police officers, or checking when household robots last cleaned the bathroom. As QA is fundamental for making multi-agent systems transparent, controllable, and manageable, it calls for systematic investigation and comprehensive evaluation to inform better system design.

A main challenge in video QA with multiple agents lies in managing extremely long egocentric video streams and aggregating the experience of each agent. Since embodied agents continuously capture their egocentric video streams during operation, models must be capable of reasoning over very long temporal horizons~\cite{pei2024egovideo, tian2025egor1, yeo2025worldmm}. In realistic scenarios, embodied agents may operate for days, with each agent generating massive video histories, making it difficult to locate the query-related events and deliver accurate answers. In practice, current video LLMs are still limited to processing video~\cite{kim2025videoicl} at the scale of only a few hours~\cite{wang2024lvbench, wu2024longvideobench}, falling short of the requirements for multi-agent QA. Furthermore, the system should go beyond understanding the history of a single agent and integrate events across multiple agents to achieve system-level comprehension.


Therefore, we construct a new benchmark evaluating the models' understanding capability in multi-agent egocentric videos, named MultiAgent-EgoQA (MA-EgoQA). MA-EgoQA consists of 1.7k question-answer pairs based on the EgoLife egocentric video dataset~\cite{egolife}. Captured by 6 people living in a shared house for 7 days,  the dataset provides five categories unique and fundamental in multi-agent scenarios: Social Interaction, Task Coordination, Theory of Mind, Temporal Reasoning, and Environmental Interaction. Questions and answers are generated using GPT-based pipelines, refined through LLM filtering, and ultimately validated by human annotators to ensure quality. We also propose a baseline model called EgoMAS, designed to illustrate the potential of multi-agent video QA. EgoMAS aggregates information from all agents into a shared memory, and upon receiving a query, it selectively retrieves the relevant information from the appropriate agents. Compared to the naive method concatenating all information from agents, selective retrieval based on the shared memory of EgoMAS is remarkably token-efficient and superior in QA performance.

We evaluate a range of existing LLMs and video LLMs on MA-EgoQA, including our EgoMAS model. Results reveal that current models struggle to handle multiple egocentric video streams effectively: even state-of-the-art LLMs and video LLMs fail to capture the complexities of multi-agent egocentric video understanding. In contrast, EgoMAS, despite its simplicity, outperforms the baseline by 4.48\% within the same Gemini-2.5-Flash~\cite{comanici2025gemini25} backbone, and even EgoMAS with Qwen3VL-8B-Thinking~\cite{bai2025qwen3vl} achieves a performance comparable to the Gemini baseline operating at a 1M-token context. 
These findings underscore both the difficulty of the task and the promise of specialized approaches like EgoMAS.

%% file: fig/relwork_comparison.tex
\begin{figure*}[t!]
\centering
\begin{minipage}[t]{0.46\textwidth}
\vspace{0pt}
\centering
\captionsetup{type=table, skip=2pt}
\captionof{table}{Comparison between MA-EgoQA and existing benchmarks.}
\label{tab:relworks}
\resizebox{\textwidth}{!}{
\begin{tabular}{lcccccc}
\toprule
\textbf{Benchmarks} & \textbf{\#QA} & \makecell[c]{\textbf{Avg}\\\textbf{Len}} & \textbf{Ego} 
& \makecell[c]{\textbf{Days}-\\\textbf{Long}} & \makecell[c]{\textbf{Cross-}\\\textbf{Video}}
& \makecell[c]{\textbf{ToM}} \\
\midrule
EgoMemoria~{\scriptsize \cite{ye2024mm}} & 7k & $\sim$60m & \cmark & \xmark & \xmark & \xmark \\
MuMA-ToM~{\scriptsize \cite{muma-tom}} & 0.9k & 36s & \xmark & \xmark & \xmark & \cmark \\
EgoSchema~{\scriptsize \cite{egoschema}} & 5.1k & 180s & \cmark & \xmark & \xmark & \xmark \\ 
EgoThink~{\scriptsize \cite{egothink}} & 0.7k & - & \cmark & \xmark & \xmark & \cmark \\
EgoToM~{\scriptsize \cite{li2025egotom}} & 1k & 300s & \cmark & \xmark & \xmark & \cmark \\ 
EgoLifeQA~{\scriptsize \cite{egolife}} & 6k & 44h & \cmark & \cmark & \xmark & \xmark \\ 
CVBench~{\scriptsize \cite{zhu2025cvbench}} & 1k & 106s & \xmark & \xmark & \cmark & \xmark \\ 
EgoExoLearn~{\scriptsize \cite{huang2024egoexolearn}} & 2.2k & 13m & \cmark & \xmark & \cmark & \xmark \\ 
\midrule
\textbf{MA-EgoQA} & 1.7k & 266h & \cmark & \cmark & \cmark & \cmark \\
\bottomrule
\end{tabular}
}
\end{minipage}
\hfill
\begin{minipage}[t]{0.51\textwidth}
\vspace{0pt}
\centering
\includegraphics[width=\linewidth]{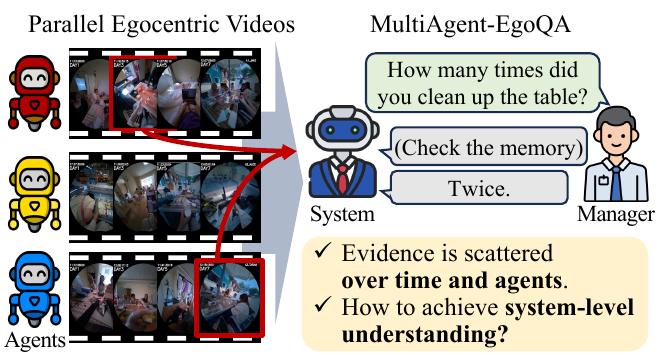}
\vspace{-0.15in}
\captionsetup{type=figure, skip=2pt}
\captionof{figure}{Problem formulation of MultiAgent-EgoQA and the associated challenges.}
\label{fig:concept}
\end{minipage}
\vspace{-0.2in}
\end{figure*}

%% file: sec/2_related_work.tex
\section{Related Work}

\subsection{Multiple Embodied Agents System}
\label{sec:rel_work_mas}
Existing works on multiple embodied agent systems have been developed, focusing on studying effective approaches to cooperate in diverse environments. CoELA~\cite{CoELA} integrates perception, memory, and execution in a modular framework and coordinates plans across agents in a natural language using LLM, and Co-NavGPT~\cite{co-navgpt} employs vision-language model (VLM) as a global planner to enable multiple robots to explore complex environments. Meanwhile, several papers evaluate models under limited communication resources and partial observability, which is closer to real-world settings~\cite{ace, furnmove}. It is also shown that structured and leadership prompting enhances teamwork efficiency and reduces unnecessary communication~\cite{guo2024embodied}, and PARTNR~\cite{partnr} built a large-scale human–robot collaboration benchmark, demonstrating that even state-of-the-art LLM-based systems still face limitations in planning and coordination. However, while these studies have primarily focused on optimizing action execution and cooperation strategies, the problem of integrating egocentric experiences collected over long periods by multiple agents to perform QA remains insufficiently addressed.

\subsection{Egocentric Video Understanding Benchmarks}
Since egocentric video understanding is crucial for real-world applications, a number of benchmarks~\cite{ye2024mm, huang2024egoexolearn} have been introduced to evaluate video QA in this setting. Ego4D~\cite{ego4d} released 3,670 hours of egocentric video along with multiple tasks, including episodic memory and hand–object interaction, establishing a fundamental resource for egocentric video research. Building on this foundation, EgoSchema~\cite{egoschema} evaluates minute-level video understanding with a focus on long-term context, while EgoThink~\cite{egothink} defines six core capabilities and twelve dimensions of egocentric reasoning to assess the interpretive and inferential capacity of VQA models. Extending beyond recognition, EgoPlan-Bench~\cite{egoplan} introduced tasks for embodied planning, examining how models can connect visual observations with action planning in egocentric settings. However, in all of these benchmarks, the video duration per sample remains shorter than one hour, which is not reflective of embodied agents that operate continuously for days. 

To address this limitation, EgoLife~\cite{egolife} constructed a super-long egocentric video dataset in which six individuals wore camera-equipped glasses to capture their daily experiences over seven consecutive days in a shared house. While this dataset breaks the length barrier of prior work, its QA benchmark, EgoLifeQA, is designed under a single-agent assumption, and questions can be answered by referencing only one individual’s memory. 
Meanwhile, MA-EgoQA is the first to evaluate the QA task on multiple, super-long, and temporally aligned egocentric videos. \cref{tab:relworks} compares existing benchmarks with MA-EgoQA.

%% file: sec/3_ma-egoqa.tex
\input{fig/examples}

\section{MA-EgoQA Benchmark}
\label{sec:ma-egoqa}
We now introduce MA-EgoQA, a new benchmark designed to evaluate the ability to comprehend and reason over multiple egocentric video streams from embodied agents. 
We first formally define the task in \cref{sec:task_definition} and describe the categories in \cref{sec:benchmark_categories}. 
Finally, we provide a detailed analysis of MA-EgoQA, presenting dataset statistics and qualitative examples in \cref{sec:benchmark_analysis}. 
We present QA examples for each category in \cref{fig:examples}, highlighting that the questions are unique to multi-agent settings and require references to multiple timestamps.

\subsection{Task Definition}
\label{sec:task_definition}

We formally define the \textbf{Multi-Agent Egocentric Video Question Answering} task as follows. Let there be $N$ embodied agents, where each agent $A_i$ continuously records an egocentric video stream $V_i$ for $T$ hours. The complete multi-agent video collection is denoted as $\mathcal{V} = \{V_1, V_2, \dots, V_N\}$, resulting in a total of $N \times T$ hours of video. Given a user query, the system should generate a response based on information from $\mathcal{V}$. In this task, a query requires information from more than two agents, causing two main challenges. First, the system must build a coherent global understanding across agents. To achieve this, each agent’s information should be stored in a way that is temporally and contextually aligned with others. Second, the model must effectively retrieve relevant events from the saved representations to answer each query. Since queries may require referencing different timestamps across multiple agents, retrieval becomes significantly more complex and challenging than in a single-agent setting.


\subsection{Benchmark Categories}
\label{sec:benchmark_categories}
We design five categories in the \textbf{MA-EgoQA benchmark} prior to generating the question-answer pairs, ensuring that the data generation pipeline is optimized for each. In selecting these categories, our primary goal is to capture aspects that are unique to the multi-agent setting and essential for human-agent system interactions in the real world. The categories are described as follows.

\textbf{Social Interaction (SI)}
This category evaluates the ability to accurately localize and ground casual conversations or affiliative behaviors across video streams.
This category contains questions related to how people engage or respond to others, group behaviors without specific goal sequences involving multiple people interacting, and meaningful information during the conversation.

\textbf{Task Coordination (TC)}   
As highlighted in \cref{sec:rel_work_mas}, research on MAS has largely focused on collaboration to achieve shared goals.
To reflect this practical importance, we define this category. Questions in this category address how roles were assigned, responsibilities divided, and actions sequenced toward goal completion, and how decisions were made throughout the execution of tasks. 

\textbf{Theory of Mind (ToM)}
ToM refers to the cognitive ability to reason about the mental states of others, including their thoughts, beliefs, desires, and emotions. 
Effective use of diverse perspectives is crucial for capturing contextual cues and improving performance on ToM questions. 
This category includes queries about what an agent believed or misunderstood, what information they could or could not perceive, and the intentions underlying their actions.

\textbf{Temporal Reasoning (TR)}
Accurate understanding of multi-agent experiences requires aligning timelines from different egocentric video streams into a coherent global view. 
To evaluate this ability, the TR category is divided into two subcategories: \textit{concurrency}, which focuses on what one agent was doing while others performed different activities at the same time, and \textit{comparison}, which examines the relative temporal ordering of events across agents.

\textbf{Environmental Interaction (EI)} Agents also engage with their surrounding environment, such as operating a vacuum cleaner or turning on a faucet. Since these interactions are distributed across multiple agents, aggregating them is essential for tracking environmental states and planning appropriate actions to achieve shared goals. Questions in this category explore object usage, including frequency, first-time use, and which agent engaged with them most.

For SI and TC categories, we additionally design multi-span questions that require reasoning across multiple non-contiguous temporal windows, thereby extending MA-EgoQA into a genuinely long-horizon benchmark. Benchmark examples for each category are shown in \cref{fig:examples} and Suppl.~\ref{appendix:samples}.
\input{fig/statistics}

\subsection{Benchmark Statistics}
\label{sec:benchmark_analysis}
MA-EgoQA contains 1,741 questions across five categories, and each question must be answered based on multiple egocentric videos captured by six agents over seven days. We analyze the distribution of questions in MA-EgoQA across question types, categories, days, and agents, and present the statistics in \cref{fig:stats}. The results show that the questions are well distributed across the five categories, the seven recorded days, and the six agents. This balance ensures that the evaluation is not dominated by a single skill or event but provides a diverse set of challenges and requires the model to track and reason about all agents effectively. The benchmark includes questions with diverse linguistic formulations. The majority of questions begin with \texttt{What} and \texttt{Who}, while questions beginning with \texttt{Why}, \texttt{How}, \texttt{When}, and \texttt{Which} broaden the range of question types. Overall, the broad coverage of questions in MA-EgoQA ensures a comprehensive evaluation of multimodal reasoning in egocentric multi-agent scenarios.

%% file: fig/examples.tex
\begin{figure*}[t!]
    \vspace{-0.1in}
    \centering
    \includegraphics[width=0.95\textwidth]{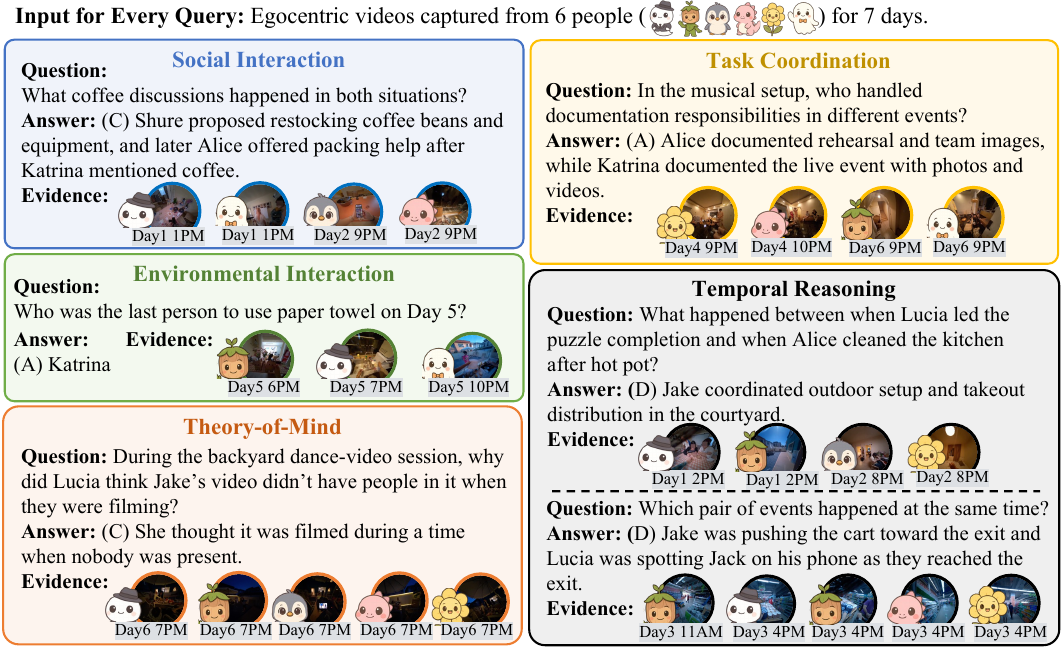}
    \vspace{-0.1in}
    \caption{Examples from MA-EgoQA across five categories. MA-EgoQA is the first multiple embodied agents egocentric video QA benchmark, requiring comprehension of six egocentric videos spanning seven days per query. False options are omitted. }
    \label{fig:examples}
    \vspace{-0.2in}
\end{figure*}

%% file: fig/statistics.tex
\vspace{-0.1in}
\begin{figure}[t!]
    \centering
    \begin{minipage}[t]{0.24\textwidth}
    \centering
    \includegraphics[width=\linewidth]{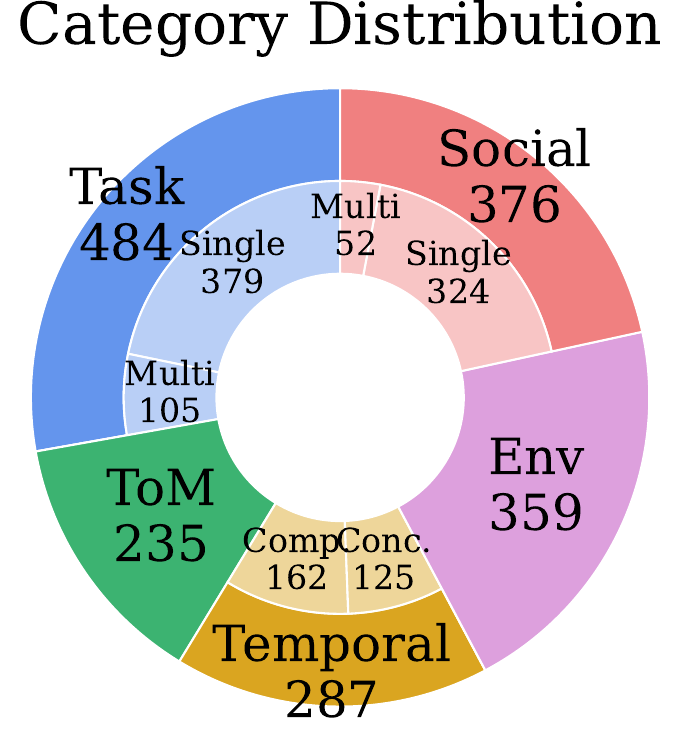}
    \end{minipage}
    \hfill
    \begin{minipage}[t]{0.24\textwidth}
    \centering
    \includegraphics[width=\linewidth]{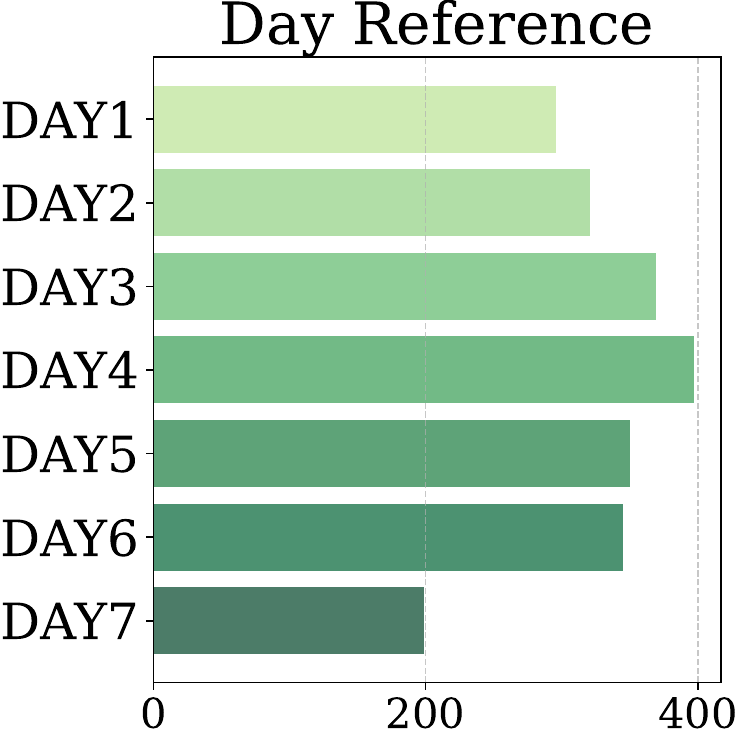}
    \end{minipage}
    \hfill
    \begin{minipage}[t]{0.24\textwidth}
    \centering
    \includegraphics[width=\linewidth]{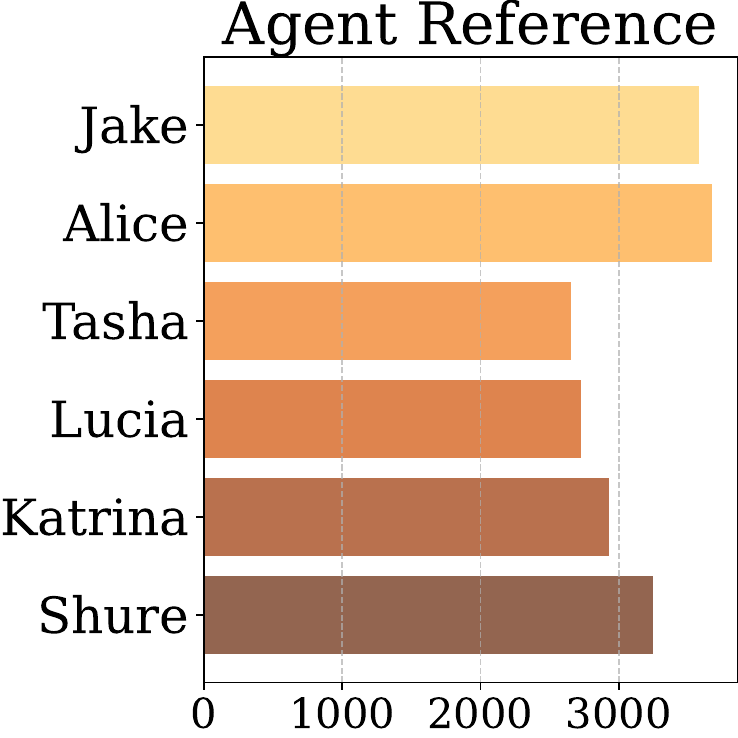}
    \end{minipage}
    \hfill
    \begin{minipage}[t]{0.24\textwidth}
    \centering
    \includegraphics[width=\linewidth]{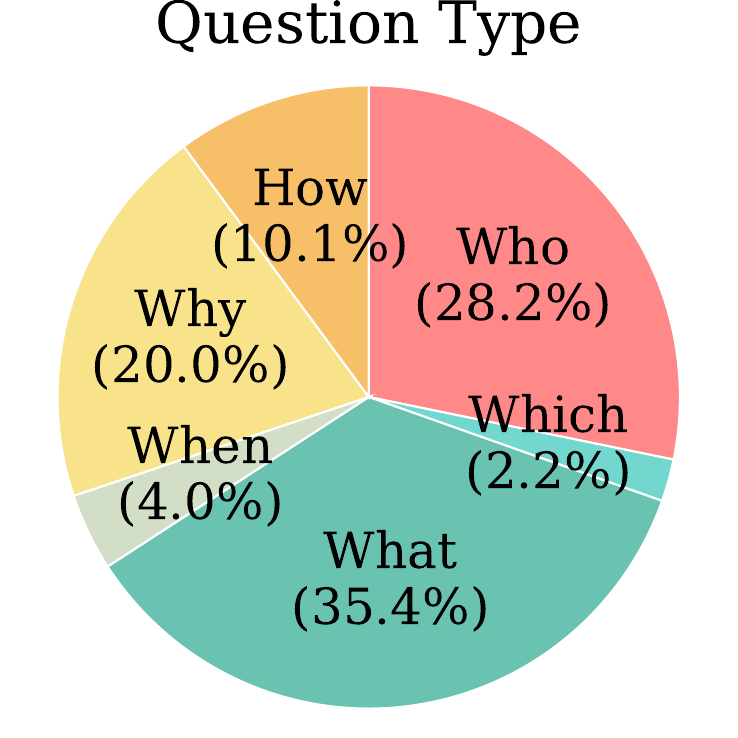}
    \end{minipage}
    \vspace{-0.1in}
    \caption{Statistics for MA-EgoQA; \textit{(Left)}~number of samples per category; \textit{(Center)}~day/agent reference counts of all categories; \textit{(Right)}~question type distribution. }
    \vspace{-0.2in}
    \label{fig:stats}
\end{figure}

%% file: sec/4_benchmark_construction.tex
\section{Benchmark Construction}
\subsection{Data Generation}
\label{sec:data_generation}
We construct candidate question–answer pairs using GPT-based generation. Since the nature of questions differs across categories, we adopt two complementary strategies. For SI, TC, and ToM categories, where question formats are diverse and open-ended, we generate a large pool of samples and subsequently filter low-quality pairs using the LLM-based pipeline described in \cref{sec:llm_filtering}. In contrast, for TR and EI, which typically involve more structured queries, we design predefined templates and generate QA pairs by instantiating them with contextual information, followed by filtering and verification. For convenient evaluation, MA-EgoQA is designed to be multiple-choice questions with five options. We describe the candidate generation process in detail below. The prompts and templates used in each generation stage are shown in Suppl. \ref{appendix:prompts} and \ref{appendix:templeates}. 
\input{fig/generation}

\paragraph{Single-span Question Generation}
Each long egocentric video from an agent is divided into fixed-length segments of 5 minutes. For each 5-minute window, we collect dense captions and transcripts from all available agents during that period and provide them to GPT-4o, together with a system prompt and a category-specific prompt for SI, TC, and ToM categories. The prompts instruct the model to understand the events within the 5-minute context and to generate a question, an answer, a rationale, the names and timestamps of the referenced agents, and four wrong options. Each generated question must be grounded by more than two agents and relevant to its assigned category.
In total, we generate 33.4k, 31.6k, and 34.1k QA samples for the SI, TC, and ToM categories, respectively.

\paragraph{Multi-span Question Generation}
Multi-span questions are constructed by grouping semantically similar single-span questions, ensuring that each multi-span question addresses the same topic across different timestamps. This design ensures that the resulting questions are coherent and contextually plausible. To be specific, we concatenate each single-span question $q$ and its paired answer $a$, and obtain a sentence embedding $e$ using a text encoder $E(\cdot)$, which is represented as $e = E\big([q; a]\big).$
We then define pairwise cosine similarity between two samples as $s_{ij} = \langle e_i, e_j \rangle/\|e_i\|\|e_j\|$.
A similarity graph is constructed where each vertex corresponds to a QA pair and there is an edge $(i,j)$ if $s_{ij} \geq \delta$, with $\delta$ a predefined threshold. Connected components are extracted from this graph as groups, where each group contains at least two semantically related QA pairs. For each group, all single span questions and their correct answers in the group are given to GPT-5 with the system prompt and category-wise prompt, and the model synthesizes multi-span question and answer only, not false options at this stage. To ensure difficulty and quality, false options are created separately, rather than being inherited from the single-span samples. We generated 15.9k samples for SI multi-span questions and 16.3k samples for TC multi-span questions.

\paragraph{Template-based Question Generation}
To improve the efficiency of QA generation, we employ predefined template-based methods for the TR and EI categories, as questions in these categories share common objectives within each category, such as ordering event timestamps or counting object usage. To generate TR questions, we first create captions for every agent's videos using temporal windows of 30 seconds, 10 minutes, and 1 hour to encourage coverage across multiple temporal scales. We then define two subcategories within TR: the \textit{comparison} type, which focuses on ordering events across multiple agents, and the \textit{concurrency} type, which requires temporal alignment among agents. 
After constructing all captions, we generate each QA instance by selecting several captions from multiple agents at the same temporal scale and providing them to GPT-5 along with a prompt. For the \textit{comparison} type, we randomly sample different timestamps for each agent. For the \textit{concurrency} type, we randomly select a single timestamp and gather captions from all agents at that time. GPT-5 then generates a question with five answer options, including one correct answer. 

For the EI category, we first manually list the objects with which agents interact in the shared environment. For each object and for each six hour or one day interval, we collect all dense captions from all agents that include the object. The collected captions are then paired with predefined QA templates and provided to GPT-5, which analyzes interactions between the agents and the object within the interval and generates a question with five answer options.

\subsection{LLM Filtering and Human Verification}
\label{sec:llm_filtering}
We employ an automatic LLM-based filtering prior to the human verification stage. Each generated question–answer pair is verified by prompting an LLM with a restricted input context and evaluating whether the model can correctly infer the correct answer. If the model predicts the answer under these constraints, the sample is deemed insufficiently challenging and excluded from the benchmark. All filtering steps are conducted using GPT-5 in the following order, which serves as the backbone for quality control in our data generation pipeline.

\paragraph{Zero-shot Filtering}
In this stage, each question is evaluated without providing any context, thereby testing whether the correct answer can be trivially inferred. If the model consistently predicts the correct option under such a zero-context setting, the sample is considered unchallenging and excluded. Concretely, we query the model three times per question, and if it produces the correct answer in more than two trials, the corresponding sample is discarded from the dataset.

\paragraph{Single Agent Filtering}
Since MA-EgoQA is designed to evaluate reasoning in a multi-agent context, we remove any sample that can be answered using the memory of a single individual. To enforce this criterion, we extract all human names appearing in the question and the correct answer, and perform inference using each corresponding individual’s memory. If no human name is present, we randomly select one from the available agents and conduct inference with their memory. If the model produces the correct answer under any single-agent memory condition, the sample is eliminated from the benchmark.

\paragraph{Cross-model Validation}
All preceding filtering stages rely on GPT-5, which may introduce model-specific biases. To address this, we include an additional verification using two external models, Gemini-2.5-Flash and Claude-Sonnet-4. Each sample is re-examined under the same contextual information available at generation time. 
This step not only checks the correctness of the answer, but also validates the question and the false options, ensuring that the query is non-trivial.
If either external model flags a sample as invalid, it is removed.

\paragraph{Human Verification}
\label{sec:human_verification}
Following all automatic filtering steps, a final human verification stage is conducted to ensure the quality and validity of the remaining candidates. Human verifiers are granted access to the complete set of information, including captions, transcripts, and videos from all agents.
Each verifier evaluates the samples according to the same guidelines used during data generation, ensuring consistency with the original design criteria. In total, four human verifiers reviewed 3,436 candidates, and 1,741 high-quality samples were selected.

%% file: fig/generation.tex
\begin{figure*}[t!]
    \centering
    \includegraphics[width=\textwidth]{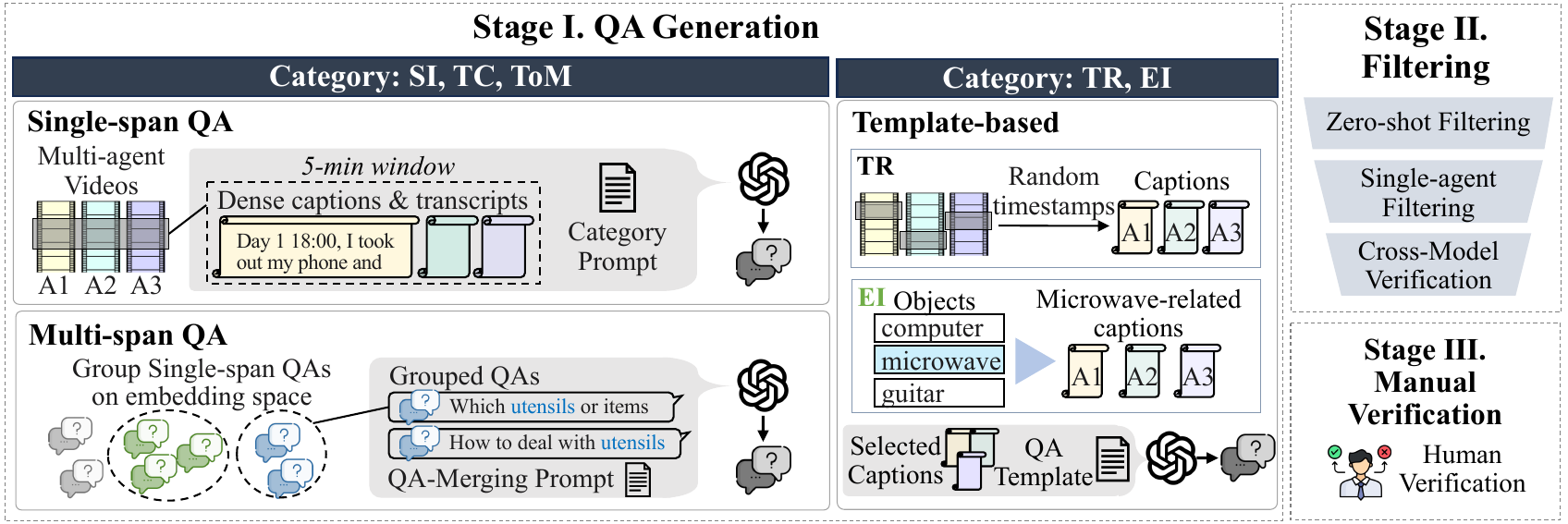}
    \vspace{-0.25in}
    \caption{Benchmark construction pipeline. QA pairs are generated for each category through its dedicated process, and refined through LLM filtering and manual check.}
    \label{fig:generation}
    \vspace{-0.2in}
\end{figure*}

%% file: sec/5_egomas.tex
\input{fig/egomas}

\section{Egocentric Video Reasoning in Multi-Agent System}
\label{sec:egomas}
Beyond introducing the new benchmark, we also propose a simple yet effective training-free baseline, named EgoMAS (Egocentric Multi-Agent System), to stimulate future research on MA-EgoQA. EgoMAS is a centralized MAS that is specifically designed to tackle the unique challenges of multi-agent egocentric reasoning. First, it leverages an event-based shared memory that enables a system-level global understanding by integrating fragmented events across agents. Second, EgoMAS dynamically selects which agent’s memory to reference and adapts the search query for each agent, allowing fine-grained reasoning across multiple perspectives. Prompts for each step are provided in Suppl. \ref{appendix:egomas_prompts}.

\subsection{Event-based Shared Memory}
At every 10-minute interval, each embodied agent provides a caption summarizing its own observations during the preceding time span. A centralized manager then integrates these individual captions into a system-level summary. Rather than producing a flat textual condensation, the manager first identifies key events across agents and, for each event, explicitly records the corresponding 4W1H fields: \texttt{When}, \texttt{What}, \texttt{Where}, \texttt{Who}, and \texttt{How}, producing a coherent global memory that aligns agent perspectives while preserving critical details for reasoning.

\subsection{Agent-wise Dynamic Retrieval}
Given a query $q$, EgoMAS first retrieves top-$n$ system-level memories from the shared memory $\mathcal{M}_{\text{shared}}$ using BM25 ranking 
$$\mathcal{R}_{\text{sys}}(q) = \text{Top-}n \; \{ (m, s(m,q)) \mid m \in \mathcal{M}_{\text{shared}} \},$$
where $s(m,q)$ denotes the BM25 score between memory $m$ and query $q$.

\noindent From the retrieved system-level context, EgoMAS generates a set of agent-specific retrieval requests $\mathcal{Q}_{\text{agent}} = \{ (a_j, q_j) \}_{j=1}^J,$
where each request consists of an agent identifier $a_j$ and a sub-query $q_j$. For each $(a_j, q_j)$, EgoMAS performs agent-level retrieval from the agent’s memory $\mathcal{M}_{a_j}$:
$$
\mathcal{R}_{a_j}(q_j) = \text{Top-}k \; \{ (m, s(m,q_j)) \mid m \in \mathcal{M}_{a_j} \}.
$$
To ensure relevance, we filter out memories with scores below a threshold $\tau$:
$$
\widetilde{\mathcal{R}}_{a_j}(q_j) = \{ (m, s(m,q_j)) \in \mathcal{R}_{a_j}(q_j) \mid s(m,q_j) \geq \tau \}.
$$
Finally, the system generates the final response by conditioning on both the retrieved system-level context $\mathcal{R}_{\text{sys}}(q)$ and the aggregated agent-level results $\widetilde{\mathcal{R}} = \bigcup_{j=1}^J \widetilde{\mathcal{R}}_{a_j}(q_j)$:
$$
\hat{y} = F\big(q, \mathcal{R}_{\text{sys}}(q), \widetilde{\mathcal{R}}\big),
$$
where $\hat{y}$ and $F$ denote the response and response generation function.

%% file: fig/egomas.tex
\begin{figure*}[t!]
    \centering
    \includegraphics[width=0.95\textwidth]{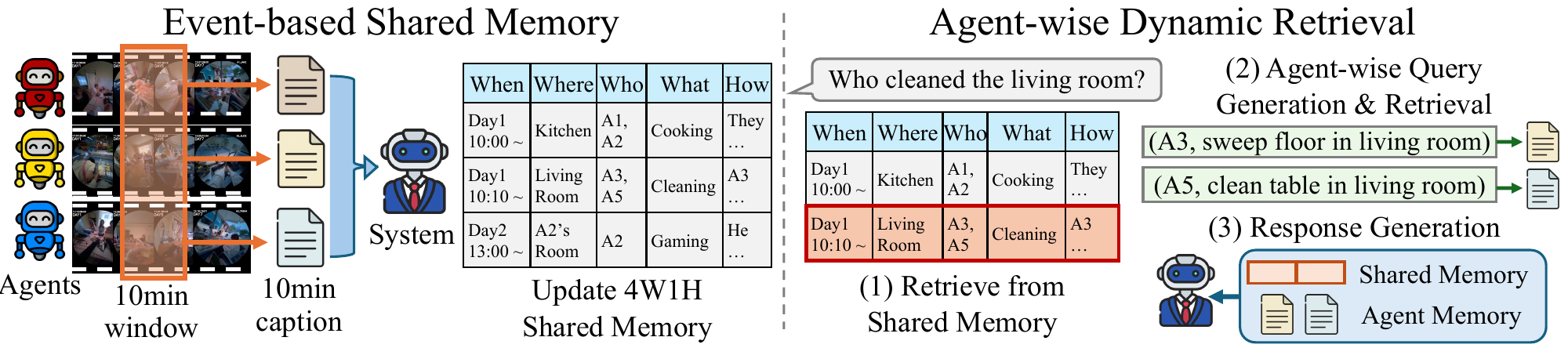}
    \caption{Overview of EgoMAS. EgoMAS builds an event-based shared memory from multi-agent videos and dynamically retrieve relevant information from it.}
    \label{fig:egomas}
    \vspace{-0.2in}
\end{figure*}

%% file: sec/6_experiment.tex
\section{Experimental Setup and Results}
\subsection{Experimental Setup}
\label{sec:experimental_setups}
We evaluate a range of competitive open-source and proprietary LLMs and video LLMs on MA-EgoQA. Proprietary LLMs include Gemini-2.5-Flash~\cite{comanici2025gemini25} and GPT-5~\cite{OpenAI_GPT5_2025}. Open-source LLMs include Llama-3.1~\cite{grattafiori2024llama}, Qwen2.5-1M~\cite{yang2025qwen251m}, Qwen3~\cite{yang2025qwen3}, and gpt-oss~\cite{agarwal2025gptoss}. Video LLMs include VideoChat-Flash~\cite{li2024videochatflash}, Video-XL-2~\cite{qin2025videoxl2}, and Qwen2.5-VL-7B~\cite{bai2025qwen25vl}. For text-only LLMs, we concatenate captions from all agents in chronological order and directly pass them to the LLM along with the question.
For video LLMs, we concatenate all video frames from all agents in chronological order and uniformly sample the number of frames that each model can receive without exceeding the model's input limits. 

\input{tab/main_results}

We also evaluate several RAG-based baselines, including simple text-based retrieval methods BM25~\cite{robertson2009bm25} and DPR~\cite{karpukhin2020dpr} with Qwen3-VL-8B-Instruct~\cite{bai2025qwen3vl} and VideoRAG~\cite{jeong2025videorag} with InternVideo2~\cite{wang2024internvideo2} encoder. There are three baselines that work on the EgoLife dataset, but in a single-agent setting: EgoRAG~\cite{egolife}, Ego-R1-Agent-3B~\cite{tian2025egor1}, and WorldMM-8B~\cite{yeo2025worldmm}. Since they do not support multi-agent scenarios, we only use videos from a single person in EgoRAG, 
and use an LLM router to select which individual's memory to refer to in other models. RAG-based baselines leverage both frames and captions as memories.

For EgoMAS, we evaluate the system with four different backbone models: Gemini-2.5-Flash~\cite{comanici2025gemini25}, Qwen3VL-8B-Thinking~\cite{bai2025qwen3vl}, Qwen3VL-8B-Instruct~\cite{bai2025qwen3vl}, and Qwen2.5VL-7B-Instruct~\cite{bai2025qwen25vl}. The hyperparameters are set to $n=20$, $k=5$, $\tau=10$. Lastly, we also evaluate an Oracle setting, which generates answers using the same context used to generate the questions. More details including frame sampling and memory retrieval in baselines are described in Suppl. \ref{appendix:experimentalsetups}.

\subsection{Evaluation Results}
We present the MA-EgoQA evaluation results in \cref{tab:main_results}. Overall, the results show that MA-EgoQA is a highly challenging benchmark for current models. Even the strongest model, Gemini-2.5-Flash, achieves an average accuracy of only 36.93\%, and many baselines, including gpt-oss models and VideoXL-2, perform only marginally above random chance, highlighting the difficulty of the task. Specifically, in \texttt{All Caption Concat Baselines}, only proprietary models achieve accuracies above 30\%, while smaller open-source models show below 27\% despite handling extremely large input context (128k to 1M tokens). These results indicate that providing all captions without retrieval tends to distract the model with irrelevant information while requiring substantial computation costs. Meanwhile, \texttt{All Frame Concat Baselines} achieve the lowest average performance among all baseline categories, likely due to the absence of transcript information and the presence of many irrelevant frames.

In contrast, \texttt{RAG Baselines}, including BM25 and WorldMM-8B, outperform other non-retrieval open-source models while using significantly shorter input contexts. This demonstrates that retrieval is crucial for efficiently understanding multiple long egocentric videos. 
Finally, EgoMAS significantly outperforms all baselines. EgoMAS (Gemini-2.5-Flash) achieves 41.41\% accuracy, which is 4.48\% higher than the Gemini-2.5-Flash baseline. Notably, even Qwen3VL-8B-based EgoMAS models surpass Gemini-2.5-Flash and GPT-5, highlighting the effectiveness of the EgoMAS framework. We also observe that EgoMAS performance scales with the capability of the backbone model, improving from Qwen2.5VL-7B-Instruct to Gemini-2.5-Flash backbone, suggesting that stronger reasoning ability enables more effective retrieval strategies and better integration of information retrieved from multiple agents. 
However, all models exhibit a substantial gap compared to the oracle setting, indicating significant room for improvement.



%% file: tab/main_results.tex
\begin{table*}[t]
\centering
\caption{Evaluation results of 16 baselines and EgoMAS on MA-EgoQA.}
\label{tab:main_results}
\vspace{-0.1in}
\renewcommand{\arraystretch}{0.95}
\setlength{\tabcolsep}{0.05in}
\scriptsize
\begin{adjustbox}{width=0.95\textwidth}
\begin{tabular}{p{4.5cm} c c c c c c c}
\toprule
\textbf{Models} & \begin{tabular}[c]{@{}c@{}}\textbf{Input}\\ \textbf{context}\end{tabular} 
& \textbf{SI} & \textbf{TC} & \textbf{ToM} & \textbf{TR} & \textbf{EI} & \textbf{Avg.} \\
\midrule
\midrule
Random & - & 20.00 & 20.00 & 20.00 & 20.00 & 20.00 & 20.00 \\
\midrule
\multicolumn{8}{l}{\cellcolor{gray!20}\textbf{All Caption Concat Baselines}}\\
Gemini-2.5-Flash~\cite{comanici2025gemini25} & 1M & 41.22 & 36.36 & 24.26 & \underline{46.59} & 33.98 & 36.93 \\
Llama-3.1-Nemotron-8B~\cite{bercovich2025llama} & 1M & 20.48 & 22.93 & 20.43 & 22.65 & 21.17 & 21.65 \\
Qwen2.5-7B-Instruct-1M~\cite{yang2025qwen251m} & 500k & 27.13 & 28.93 & 20.85 & 24.04 & 26.18 & 26.08 \\
GPT-5~\cite{OpenAI_GPT5_2025} & 272k & 36.17 & 33.88 & 22.55 & 39.71 & 38.72 & 34.81 \\
Qwen3-30b-a3b-instruct~\cite{grattafiori2024llama} & 240k & 29.52 & 27.69 & 16.60 & 23.00 & 26.46 & 25.56 \\
gpt-oss-120b~\cite{agarwal2025gptoss} & 128k & 28.72 & 27.48 & 22.55 & 28.22 & 25.63 & 26.82 \\
gpt-oss-20b~\cite{agarwal2025gptoss} & 128k & 19.15 & 28.10 & 23.83 & 26.83 & 25.35 & 24.81 \\
\midrule
\multicolumn{8}{l}{\cellcolor{gray!20}\textbf{All Frame Concat Baselines}}\\
VideoChat-Flash~\cite{li2024videochatflash} & 32k & 23.67 & 25.00 & 20.00 & 24.39 & 20.73 & 23.06 \\
VideoXL-2~\cite{qin2025videoxl2} & 32k & 20.21 & 19.83 & 19.15 & 20.91 & 21.73 & 20.39 \\
Qwen2.5-VL-7B~\cite{bai2025qwen25vl} & 128k & 26.33 & 26.45 & 21.28 & 22.22 & 25.07 & 25.22 \\
\midrule
\multicolumn{8}{l}{\cellcolor{gray!20}\textbf{RAG Baselines}}\\
VideoRAG~\cite{jeong2025videorag} & 8.0k & 27.13 & 26.03 & 20.85 & 25.09 & 25.63 & 25.33 \\
EgoRAG~\cite{egolife} & - & 22.34 & 21.07 & 18.30 & 15.33 & 18.87 & 19.57 \\
Ego-R1-Agent-3B~\cite{tian2025egor1} & - & 22.34 & 25.62 & 18.30 & 26.83 & 28.41 & 24.70 \\
WorldMM-8B~\cite{yeo2025worldmm} & 4.1k & 29.26 & 34.50 & 21.70 & 25.09 & 22.56 & 27.63 \\
BM25~\cite{robertson2009bm25} & 8.1k & \textbf{44.68} & 37.60 & \underline{30.21} & 33.45 & 30.64 & 36.01 \\
DPR~\cite{karpukhin2020dpr} & 7.8k &  31.12 & 28.31 & 18.72 & 26.48 & 22.84 & 26.19 \\
\midrule
\multicolumn{8}{l}{\cellcolor{gray!20}\textbf{Ours}} \\
EgoMAS \scriptsize{(Gemini-2.5-Flash)} & 4.6k & 41.49 & \textbf{41.32} & \textbf{33.62} & 39.37 & \textbf{48.19} & \textbf{41.41} \\
EgoMAS \scriptsize{(Qwen3VL-8B-Thinking)} & 5.4k & 38.03 & \underline{39.88} & 28.94 & \textbf{47.39} & \underline{44.85} & \underline{40.26} \\
EgoMAS \scriptsize{(Qwen3VL-8B-Instruct)} & 7.4k & \underline{43.09} & 39.05 & 24.26 & 35.54 & 40.67 & 37.68 \\
EgoMAS \scriptsize{(Qwen2.5VL-7B-Instruct)} & 5.4k & 37.77 & 38.64 & 25.96 & 34.15 & 36.49 & 35.55 \\
\midrule
Oracle \scriptsize{(Gemini-2.5-Flash)} & - &  86.44 & 97.60 & 70.21 & 88.15 & 81.34 & 83.80 \\
Oracle \scriptsize{(Qwen3VL-8B-Instruct)} & - & 76.86 & 80.99 & 57.02 & 77.35 & 69.92 & 73.98 \\
\bottomrule
\end{tabular}
\end{adjustbox}
\vspace{-0.1in}
\end{table*}

%% file: sec/7_analysis.tex
\input{tab/maegoqa_analysis}

\section{Analysis}
\subsection{Analysis on the MA-EgoQA Benchmark}

\paragraph{Does MA-EgoQA require multi-agent memory?} To verify that MA-EgoQA requires access to multiple agents' memories, we evaluate EgoMAS using only a single agent's memory and compare the results with the original performance. As shown in \cref{fig:single_agent}, restricting EgoMAS to a single agent leads to a substantial performance drop, demonstrating that models should utilize memories from multiple agents to correctly answer MA-EgoQA queries.

\paragraph{Multi-span and multi-agent reasoning as key challenges} We analyze which aspects of MA-EgoQA make the task challenging from two perspectives: long-horizon reasoning and multi-agent knowledge fusion. First, we compare model performance on queries that require context from a single timestamp versus multiple timestamps across SI, TC, and TR categories. As shown in \cref{tab:multi-span}, models achieve substantially lower accuracy on queries grounded in multiple spans (Multi-span in SI and TC, and Comparison in TR). This suggests that identifying and relating multiple events across seven days of video requires stronger retrieval capabilities. 
Second, we analyze performance with respect to the number of agents required to answer each query in \cref{fig:num_required_agents}. Performance consistently decreases as the number of required agents increases, indicating that current approaches to multi-agent knowledge fusion remain limited. This highlights the need for future research on constructing a coherent global understanding beyond simple memory sharing among agents. As a result, both aspects significantly contribute to increasing the task difficulty.  

\input{fig/ablation_efficiency}

\paragraph{ToM is the hardest category in MA-EgoQA}
As shown in \cref{tab:main_results}, ToM consistently exhibits the lowest accuracies across most models. Although EgoMAS substantially improves ToM performance, it still lags behind other categories. This is because ToM requires inferring latent mental states, such as goals and beliefs, rather than extracting explicitly observable visual or textual cues. This observation corroborates prior work identifying ToM as exceptionally challenging due to its inherent ambiguity, the need for recursive mental-state reasoning~\cite{muma-tom}, and complex goal–belief–action inference~\cite{li2025egotom}.

\subsection{Analysis on EgoMAS}
\paragraph{Efficiency of EgoMAS} 
\cref{fig:latency} shows the average latency measured over 100 randomly selected MA-EgoQA samples and overall accuracy of baseline models and EgoMAS (Qwen2.5VL-7B-Instruct). Non-retrieval-based models show substantially higher computational overhead, whereas retrieval-based models, including EgoMAS, exhibit notably lower latency. In particular, EgoMAS achieves the highest accuracy among retrieval-based models while requiring only 1.3 seconds per query. These results highlight that EgoMAS is a practical approach for building a multi-agent egocentric video QA system in real-world settings.

\paragraph{Efficacy of Two Methods in EgoMAS} To validate the efficacy of each method in EgoMAS, we perform an ablation study using EgoMAS (Qwen2.5VL-7B-Instruct), as shown in \cref{tab:ablation_method}. For the model without shared memory, we use 10-minute captions for all agents as a retrieval source. Meanwhile, the models without agent-wise dynamic retrieval leverage generate the response only with shared memories. Using both methods together achieves the highest accuracy, highlighting that aggregating memories from all agents into shared memory is effective, and a dynamic retrieval method helps the model to find the most informative and related knowledge from the corpus. Ablation studies on the hyperparameter for each module is presented in Suppl. \ref{appendix:hyperparamter}.

\paragraph{Analysis on Sub-modules of EgoMAS} 
We validate the contribution of each sub-module with ablation studies in \cref{tab:abl_memory}.
In \cref{subtable:abl_shared_memory}, we evaluate multiple memory construction strategies introduced in prior works~\cite{zeng2024structural, gutierrez2025from} and 4W1H memory. The event-based 4W1H memory structure clearly outperforms alternative methods, demonstrating its strong ability to abstract and fuse the events across agents.
\cref{subtable:abl_memory_retriever} presents a comparison between our choice of memory retriever and other dense retrievers~\cite{zhang2025qwen3, lee2024nv}. Although NV-Embed-v2 achieves the highest accuracy, it contains 7B parameters and consequently incurs substantial computational overhead. In contrast, BM25 employs lightweight keyword-based retrieval and delivers competitive performance, highlighting its practicality.

\input{tab/ablations}
\input{fig/case_study_tab}

\paragraph{Case Study} We further analyze the baselines and EgoMAS through the qualitative example in \cref{fig:case_study}. As discussed earlier, Gemini-2.5-Flash and VideoChat-Flash receive excessive information, which prevents them from focusing on the relevant events and leads to incorrect predictions. WorldMM attempts to retrieve memories, but since memories from agents are not aggregated, it retrieves each agent's memory iteratively and ultimately fails to answer the query. In contrast, EgoMAS successfully locates target events using shared memory, retrieves additional details through agent-wise dynamic retrieval, and responses correctly.  

%% file: tab/maegoqa_analysis.tex
\begin{figure*}[t!]
\centering

\begin{minipage}[c]{0.6\textwidth}
\input{tab/multi_span}
\end{minipage}
\hfill
\begin{minipage}[c]{0.38\textwidth}
    \includegraphics[width=\linewidth]{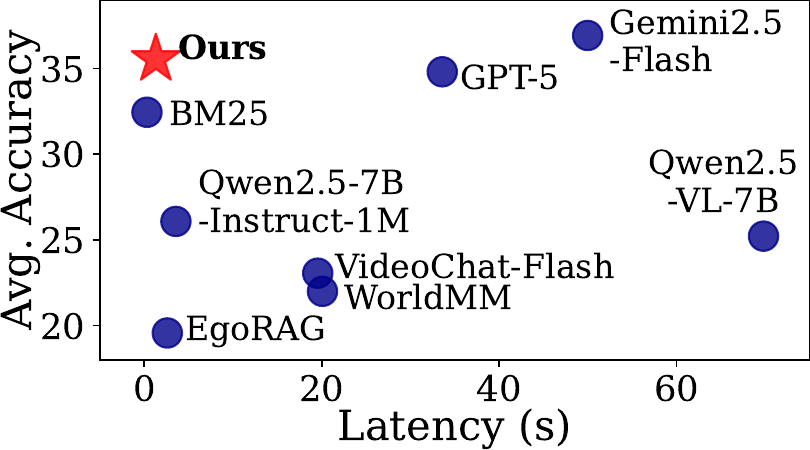}
    \vspace{-0.27in}
    \captionof{figure}{Comparison of inference latency and accuracy.}
    \label{fig:latency}
\end{minipage}


\end{figure*}

%% file: tab/multi_span.tex
\centering
\captionsetup{type=table, skip=2pt}
\captionof{table}{Performance comparison on sub-categories in SI, TC, and TR.}
\label{tab:multi-span}
\renewcommand{\arraystretch}{0.9}
\setlength{\tabcolsep}{2.5pt}
\begin{adjustbox}{width=\textwidth}
\begin{tabular}{@{}l cc cc cc@{}}
\toprule
 & \multicolumn{2}{c}{\textbf{SI}} 
& \multicolumn{2}{c}{\textbf{TC}}
& \multicolumn{2}{c}{\textbf{TR}} \\
\cmidrule(lr){2-3}\cmidrule(lr){4-5}\cmidrule(lr){6-7}
\textbf{Models} & Single & Multi & Single & Multi & Concur. & Compare \\
\midrule
Qwen2.5-7B-Instruct-1M & 27.78 & 23.08 & 28.23 & 31.43 & 28.00 & 20.99 \\
GPT-5 & 37.65 & 26.92 & 35.62 & 27.62 & 42.40 & 37.65 \\
gpt-oss-120b & 30.86 & 15.38 & 29.55 & 20.00 & 25.60 & 30.25 \\
VideoChat-Flash & 24.07 & 21.15 & 26.65 & 19.05 & 27.20 & 22.22 \\
VideoRAG & 29.94 & 9.62 & 27.97 & 19.05 & 28.00 & 22.84 \\
BM25 & 41.05 & 23.08 & 39.84 & 18.10 & 38.40 & 22.84 \\
EgoMAS \scriptsize{(Qwen3-VL-8B-Inst)} & 46.30 & 23.08 & 43.27 & 23.81 & 36.00 & 35.19 \\
\bottomrule
\end{tabular}
\end{adjustbox}

%% file: fig/ablation_efficiency.tex
\begin{figure}[!t]
    \centering
    \vspace{-0.1in}
    \begin{minipage}[c]{0.48\linewidth}
        \centering
        \includegraphics[width=\linewidth]{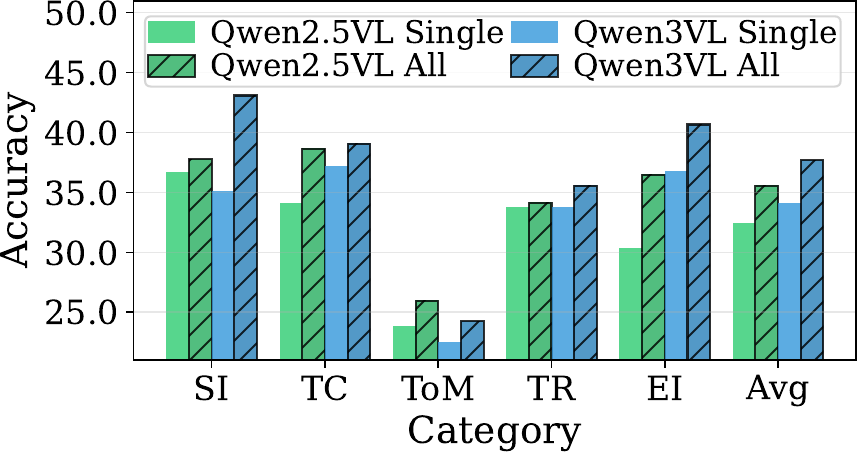}
        \vspace{-0.25in}
        \captionof{figure}{Evaluation with single-agent memory and all agents memory.}
        \label{fig:single_agent}
    \end{minipage}
    \hfill
    \begin{minipage}[c]{0.48\linewidth}
        \centering
        \includegraphics[width=\linewidth]{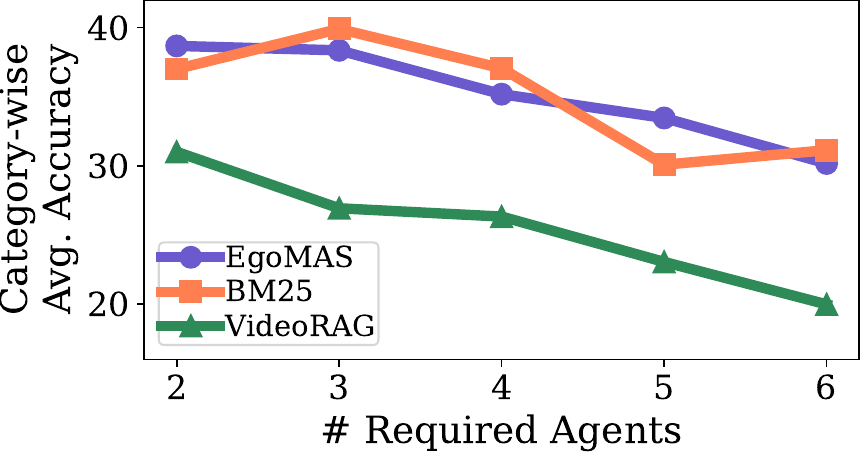}
        \vspace{-0.25in}
        \captionof{figure}{MA-EgoQA performance with varying numbers of required agents.}
        \label{fig:num_required_agents}
    \end{minipage}
    %
    \vspace{-0.2in}
\end{figure}

%% file: tab/ablations.tex
\begin{figure*}[t!]
    \vspace{-0.1in}
    \centering
    \begin{minipage}[c]{0.3\linewidth}
        \centering
        \captionof{table}{Ablation study on EgoMAS structure.}
        \renewcommand{\tabcolsep}{1mm}
        \resizebox{\linewidth}{!}{
            \begin{tabular}{ccc}
                    \toprule
                    \begin{tabular}[c]{@{}c@{}}\textbf{Shared}\\\textbf{ Memory}\end{tabular}  & \begin{tabular}[c]{@{}c@{}}\textbf{Dynamic}\\\textbf{Retrieval}\end{tabular} & \textbf{Acc.}\\
                    \midrule
                    \xmark & \xmark & 27.80 \\
                    \xmark & \cmark & 28.20 \\
                    \cmark & \xmark & 30.04 \\
                    \cmark & \cmark & \textbf{35.55} \\
                    \bottomrule
                \end{tabular}
        }
        \label{tab:ablation_method}
    \end{minipage}
    \hfill
    \begin{minipage}[c]{0.68\linewidth}
        \captionof{table}{Ablation studies on sub-modules of EgoMAS.}
        \vspace{-0.1in}
        \label{tab:abl_memory}
        \centering
        \begin{subtable}[t]{0.53\linewidth}
            \centering
            \caption{Shared Memory Structure}
            \label{subtable:abl_shared_memory}
            \resizebox{\linewidth}{!}{
                \begin{tabular}{lc}
                    \toprule
                    \textbf{Shared Memory Structure} & \textbf{Acc.} \\
                    \midrule
                    Summary & 30.67 \\
                    Triplet & 30.44 \\
                    Chunk  & 25.96 \\
                    Graph & \underline{31.99} \\
                    Ours (4W1H) & \textbf{35.55} \\
                    \bottomrule
                \end{tabular}
            }
        \end{subtable}
        \hfill
        \begin{subtable}[t]{0.44\linewidth}
            \centering
            \caption{Memory Retriever}
            \label{subtable:abl_memory_retriever}
            \resizebox{\linewidth}{!}{
                \renewcommand{\arraystretch}{1.13}
                \begin{tabular}{lc}
                    \toprule
                    \textbf{Memory Retriever} & \textbf{Acc.} \\
                    \midrule
                    DPR & 28.67 \\
                    Qwen3-Embed-0.6B & 33.03 \\
                    NV-Embed-v2 7B  & \textbf{37.91} \\
                    Ours (BM25) & \underline{35.55} \\
                    \bottomrule
                \end{tabular}
            }
        \end{subtable}
    \end{minipage}

    \vspace{-0.1in}
\end{figure*}

%% file: fig/case_study_tab.tex
\begin{figure*}[t!]
    \centering
    \includegraphics[width=0.95\textwidth]{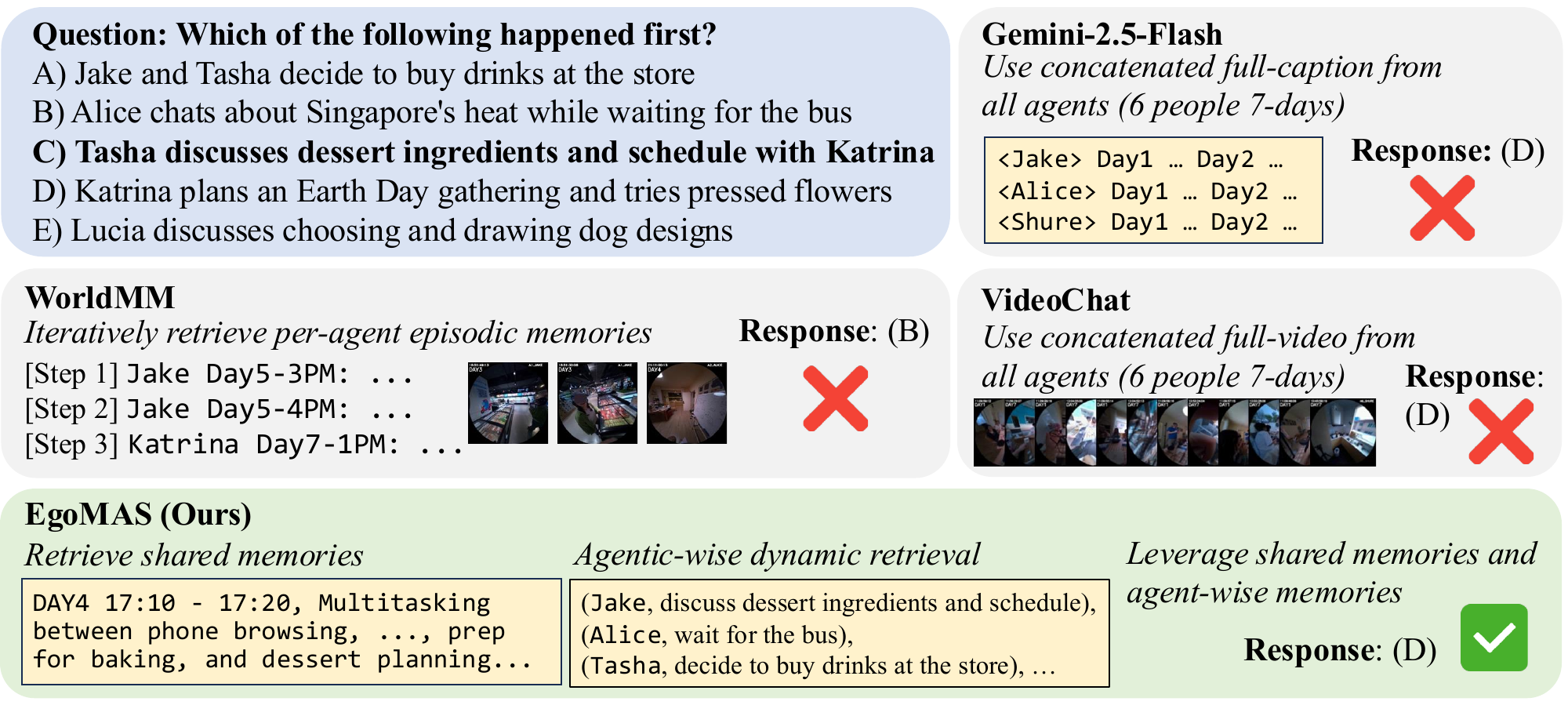}
    \vspace{-0.1in}
    \caption{Case Study: Comparative analysis across models on MA-EgoQA}
    \label{fig:case_study}
    \vspace{-0.2in}
\end{figure*}

%% file: sec/8_conclusion.tex
\section{Conclusion}
In this work, we present MA-EgoQA, a novel benchmark for question answering over multiple long-horizon egocentric video streams from embodied agents, spanning five categories that capture the core challenges of multi-agent reasoning. Alongside, we propose EgoMAS, a simple training-free baseline that combines shared memory with agent-wise dynamic retrieval, achieving superior performance to recent frontier models with significantly long context despite using a much smaller model. Our evaluations show that current LLMs and video LLMs struggle with a multi-agent egocentric setting, particularly in the theory of mind and multi-agent reasoning, underscoring the complexity of the scenario and highlighting MA-EgoQA as a promising future direction for multiple agents.

%% file: sec/X_appendix.tex
\newpage
\appendix
\setcounter{figure}{0}
\setcounter{table}{0}
\renewcommand{\thefigure}{S\arabic{figure}}
\renewcommand{\thetable}{S\arabic{table}}
\renewcommand{\theHsection}{appendix.\Alph{section}}

\section*{Supplementary Materials} 
This supplementary information includes additional analyses (Section~\ref{appendix:analysisegomas}), more details on the experimental setups (Section~\ref{appendix:experimentalsetups}), and a discussion of limitations and future work (Section~\ref{appendix:limitation}). We further provide benchmark details including category examples (Section~\ref{appendix:samples}), prompts used for benchmark generation (Section~\ref{appendix:prompts}), and question templates for the TR and EI categories (Section~\ref{appendix:templeates}). Lastly, we present additional analysis on the performance with different input context modalities on MA-EgoQA (Section~\ref{appendix:visual}).

\section{Additional Analysis on EgoMAS}
\label{appendix:analysisegomas}
\subsection{Ablation on Hyperparameters}
\label{appendix:hyperparamter}
\input{fig/hyperparameter}
There are three hyperparameters in EgoMAS: shared memory retrieval size ($n$), agent-wise dynamic retrieval size ($k$), and score threshold ($\tau$). In \cref{fig:hyperparameters}, we report the performance of EgoMAS with different values of the hyperparameter. Shared memory retrieval size is optimal around 20 to 60, since if it is too small or big, then models suffer from finding meaningful information from the given context. However, the model shows robust performance over $k$, while high threshold $\tau$ decreases the accuracy by filtering most of the retrieved data.

\begin{wrapfigure}{r}{0.3\textwidth}
    \vspace{-0.4in}
    \centering
    \includegraphics[width=\linewidth]{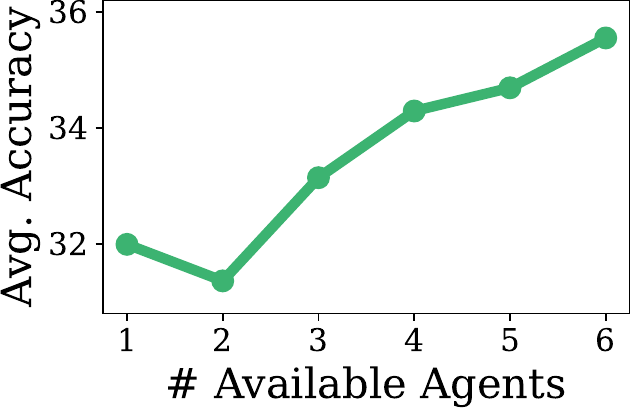}
    \vspace{-0.2in}
    \captionof{figure}{Accuracy of EgoMAS with limited agents.}
    \label{fig:num_agents}
    \vspace{-0.3in}
\end{wrapfigure}

\subsection{Sensitivity Analysis to the Number of Available Agents} 
\label{appendix:available_agents}
We investigate the sensitivity of MA-EgoQA to the number of agents by limiting the available agent views based on EgoMAS performance. As shown in \cref{fig:num_agents}, the results show a clear trend that using more agents leads to better accuracy from 31.99\% to 35.55\%. These results highlight that MA-EgoQA indeed requires models to incorporate information from multiple agents. 


\section{Additional Details on Experimental Setups}
\label{appendix:experimentalsetups}
For \texttt{All Caption Concat Baselines} in \cref{tab:main_results}, we concatenate all captions from all agents and use them as context during response generation. For Gemini-2.5-Flash, Llama-3.1-Nemotron-8B, and Qwen2.5-7B-Instruct-1M, we concatenate captions from a 10-minute window for each agent, while for the other models we concatenate captions from a 1-hour window, according to the maximum context length of each model. If the resulting context exceeds the maximum context length, we truncate each caption until the length falls within the limit.

For \texttt{All Frame Concat Baselines}, we concatenate all frames captured by the agents over 7 days and uniformly sample frames until each model can process them without errors. As a result, 10k frames are sampled for VideoChat-Flash, 4k frames for VideoXL-2, and 1.9k frames for Qwen2.5VL-7B. For the VideoRAG model, we employ the InternVideo2 multimodal encoder to extract embeddings for each 30-second clip and its corresponding caption. We then retrieve five clips based on the cosine similarity between the query and each clip. From each retrieved clip, we uniformly sample 4 frames and provide them along with the caption to the response generation model. For BM25 and DPR, we select five 30-second clips using each retriever, and four frames uniformly sampled from each clip, together with their captions, are provided to the responder. For all \texttt{RAG Baselines} except Ego-R1, we use Qwen3-VL-8B-Inst as the response generator.

\section{Limitation and Future Work}
\label{appendix:limitation}
This work has limitations as follows. First, while EgoLife provides 266 hours of egocentric videos from six people that cover a wide range of daily activities and environments, MA-EgoQA is based only on EgoLife and does not include other scenarios in different environments. This is because EgoLife is currently the only publicly available video dataset providing long-term, egocentric videos captured simultaneously from multiple agents. We hope more video datasets proposed in this direction and evaluate the model in various types of videos to ensure the generalization. 
Regarding EgoMAS, its performance still falls significantly behind the oracle performance (with a gap of around 42.4\%), indicating that further research is required to effectively address the multi-agent egocentric video QA task. One promising direction is to employ more effective retrieval methods. As shown in \cref{subtable:abl_memory_retriever}, stronger retrievers lead to better performance. Future work may explore more advanced retrieval strategies, for example hybrid retrieval methods that combine text embeddings with lexical matching.





\section{Benchmark QA Samples}
\label{appendix:samples}

We present example questions and options for each category following tables with the correct answers shown in bold.
\input{tab/qa_samples}

\clearpage
\section{Benchmark Generation Prompts}
\label{appendix:prompts}
\input{fig/prompts}

\newpage
\section{Question Templates}
\label{appendix:templeates}
\vspace{-0.2in}
\begin{table}[!htbp]
    \centering
    \caption{Question templates for the TR category.}
    \label{tab:placeholder}
    \vspace{-0.1in}
    \resizebox{0.9\linewidth}{!}{
    \begin{tabular}{p{\linewidth}}
    \toprule
    - When [Person A] was [doing X], and [Person B] was [doing Y], what was [Person C] doing? \\
    - Which of the following happened while [Person A] was doing [Action A]? \\
    - Which pair of events happened at around the same time? \\
    - Which is the correct sequence of events?\\A) Person A did X, B) Person B did Y, C) Person C did Z \\
    - Which of the following happened first?\\A) Person A did X, B) Person B did Y, C) Person C did Z, D) Person D did W, E) Person E did V \\
    - Which of the following happened last?\\A) Person A did X, B) Person B did Y, C) Person C did Z, D) Person D did W, E) Person E did V \\
    - What happened between when [Person A] was [doing X] and when [Person B] was [doing Y]? \\
    \bottomrule
    \end{tabular}
    }
\end{table}
\vspace{-0.3in}
\begin{table}[!htbp]
    \centering
    \caption{Question templates for the EI category.}
    \label{tab:placeholder}
    \vspace{-0.1in}
    \renewcommand{\arraystretch}{1.2}
    \begin{tabular}{l}
    \toprule
    - When was the first time ([object] was used/[action] done) on DAY[day]? \\
    - Who (used [object]/did [action]) the most on DAY[day]? \\
    - How many people (used [object]/did [action]) on DAY[day]? \\
    - Who was the last person to (use [object]/do [action]) on DAY[day]? \\
    \bottomrule
    \end{tabular}
\end{table}

\vspace{-0.1in}
\section{Prompts for EgoMAS}
\label{appendix:egomas_prompts}
\vspace{-0.1in}
\begin{figure}[!htbp]
    \centering
    \includegraphics[width=0.98\linewidth]{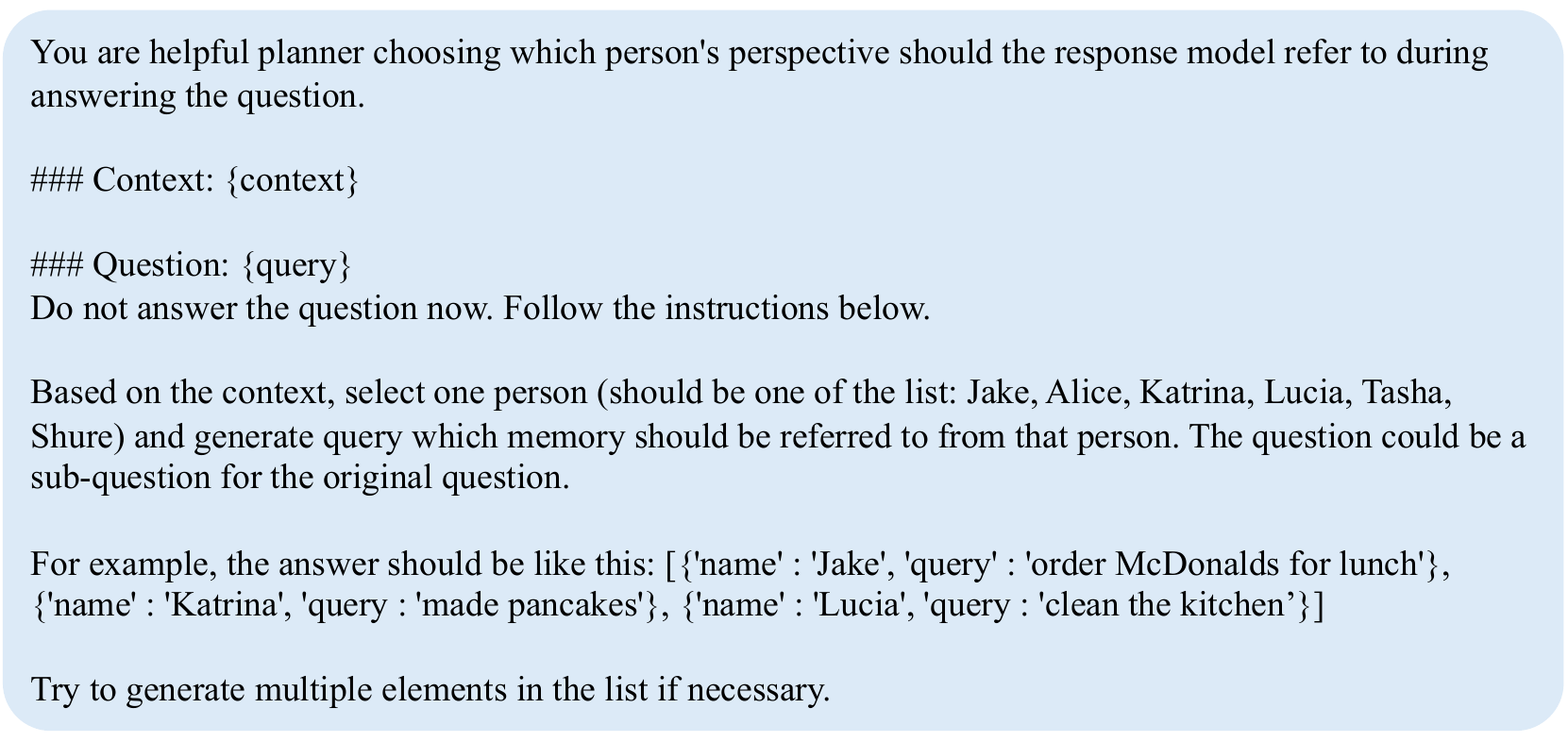}
    \vspace{-0.1in}
    \caption{Prompt for agentic-wise dynamic retrieval in EgoMAS.}
\end{figure}

\begin{figure}[!htbp]
    \centering
    \includegraphics[width=0.98\linewidth]{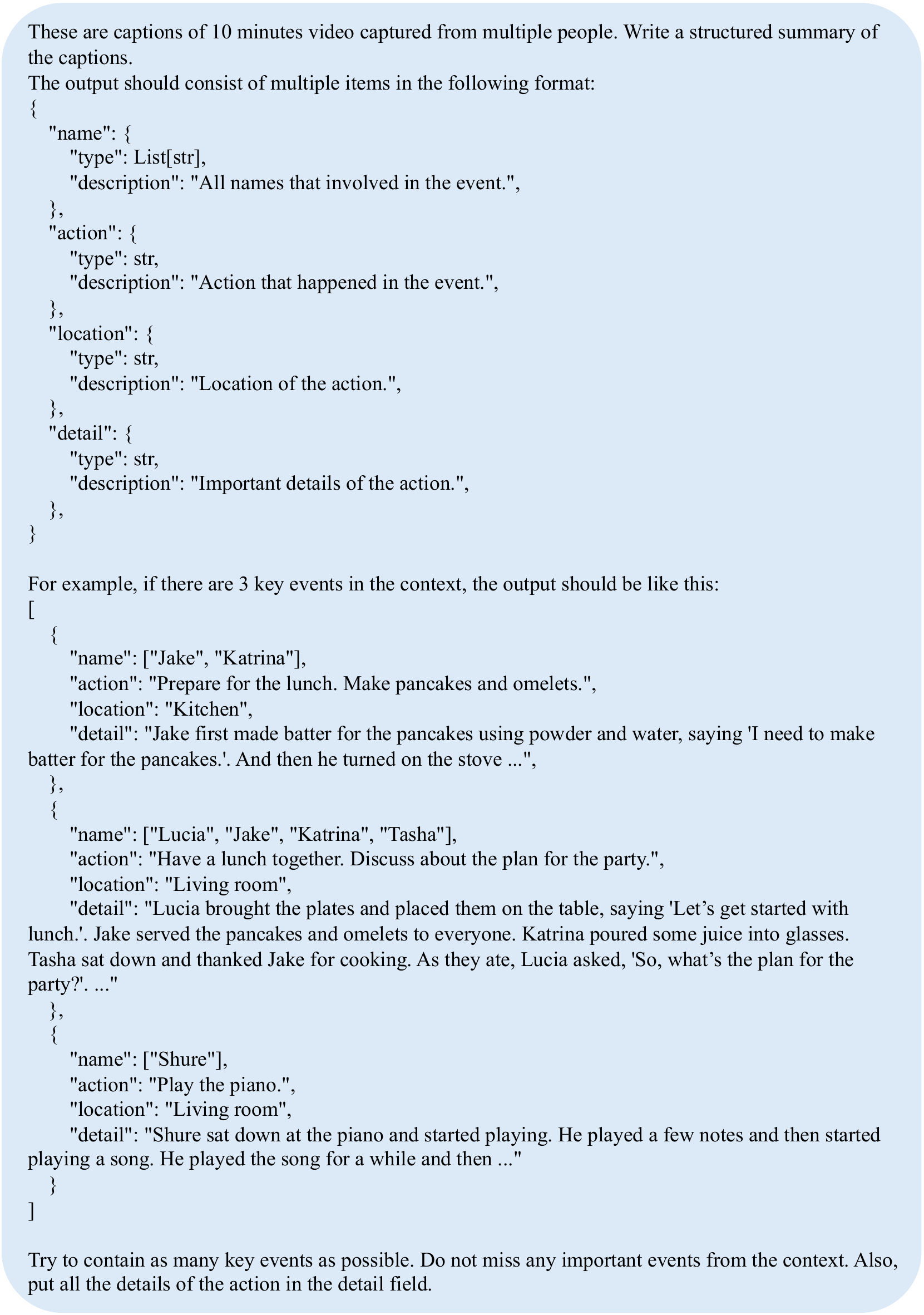}
    \vspace{-0.1in}
    \caption{Prompt for shared memory construction in EgoMAS.}
\end{figure}

\newpage
\section{Analysis on Input Context Modality}
\label{appendix:visual}
To investigate whether MA-EgoQA can benefit from information across different modalities, we compare the performance of the WorldMM-8B and EgoMAS models when using only text captions versus using both text captions and video frames as context during the response generation stage, as shown in \cref{fig:text_video}. 
For WorldMM-8B, the performance consistently improves when video frames are provided together with text captions, indicating that the MA-EgoQA benchmark requires visual knowledge contained in the frames. This observation is further supported by the QA examples in \cref{fig:visual_example}. In these examples, actions such as placing items randomly, grabbing tissues, and switching to a thinner brush, as well as states such as being in another room upstairs, can be effectively identified from the video rather than from the transcripts. These results demonstrate that visual information is essential for answering certain questions in MA-EgoQA.

On the other hand, for the EgoMAS models, providing video frames improves performance in the SI, TR, and EI categories, but leads to performance drops in the TC and ToM categories. These results suggest that including video frames is not always beneficial and can even distract the model when unnecessary frames are provided, as also observed in the \texttt{All Frame Concat Baselines} in \cref{tab:main_results}. Therefore, future research should focus on adaptively leveraging the necessary modalities and developing effective methods for frame selection.

\begin{table}
    \centering
    \caption{Analysis of input context modality on the MA-EgoQA benchmark.}
    \label{fig:text_video}
    \vspace{-0.1in}
    \setlength{\tabcolsep}{0.05in}
    \resizebox{0.75\linewidth}{!}{
        \begin{tabular}{l cc ccccc}
            \toprule
            Models & Text & Video & SI & TC & ToM & TR & EI \\
            \midrule
            \multirow{2}{*}{WorldMM-8B} & \cmark & \xmark & 27.13 & 30.37 & 17.87 & 24.04 & 19.78\\
            & \cmark & \cmark & 29.26 & 34.50 & 21.70 & 25.09 & 22.56 \\
            \rowcolor{gray!20} $\Delta$ & & & \textcolor{ForestGreen}{+2.13} & \textcolor{ForestGreen}{+4.13} & \textcolor{ForestGreen}{+3.83} & \textcolor{ForestGreen}{+1.05} & \textcolor{ForestGreen}{+2.78} \\
            \noalign{\vskip 0.25ex}\cdashline{1-8}\noalign{\vskip 0.75ex}
            \multirow{2}{1.3in}{EgoMAS \scriptsize{(Qwen2.5-VL-7B-Inst)}} & \cmark & \xmark & 37.77 & 38.64 & 25.96 & 34.15 & 36.49 \\
            & \cmark & \cmark & 39.10 & 38.02 & 24.68 & 35.19 & 37.88  \\
            \rowcolor{gray!20} $\Delta$ & & & \textcolor{ForestGreen}{+1.33} & \textcolor{Red}{-0.62} & \textcolor{Red}{-1.28} & \textcolor{ForestGreen}{+1.04} & \textcolor{ForestGreen}{+1.39} \\
            \noalign{\vskip 0.25ex}\cdashline{1-8}\noalign{\vskip 0.75ex}
            \multirow{2}{1.3in}{EgoMAS \scriptsize{(Gemini-3-preview)}} & \cmark & \xmark & 50.53 & 47.52 & 41.70 & 51.92 & 44.29  \\
            & \cmark & \cmark & 50.80 & 46.90 & 40.85 & 54.70 & 47.08 \\
            \rowcolor{gray!20} $\Delta$ & & & \textcolor{ForestGreen}{+0.27} & \textcolor{Red}{-0.62} & \textcolor{Red}{-0.85} & \textcolor{ForestGreen}{+2.78} & \textcolor{ForestGreen}{+2.79} \\
            \bottomrule
        \end{tabular}
    }
\end{table}

\vspace{-0.2in}
\begin{table}[!htpb]
    \centering
    \caption{QA samples of MA-EgoQA that require visual information.}
    \label{fig:visual_example}
    \vspace{-0.1in}
    \resizebox{0.75\linewidth}{!}{
    \begin{tabular}{p{4in}}
    \toprule
    Q. Why didn't Katrina know where to put the big kitchen tools? \\
    A. Shure had placed things randomly, making it hard to locate spaces. \\
    \\
    Q. Which pair of events happened at around the same time? \\
    A. Tasha was grabbing some tissues from the table and Alice was switching to a thinner brush to start her eye makeup. \\
    \\
    Q. Why didn’t Katrina notice the glasses charging issue earlier? \\
    A. She was upstairs in another room at the time. \\
    \bottomrule
    \end{tabular}
    }
\end{table}

%% file: fig/hyperparameter.tex
\vspace{-0.1in}
\begin{figure}[h]
    \centering
    \includegraphics[width=0.9\linewidth]{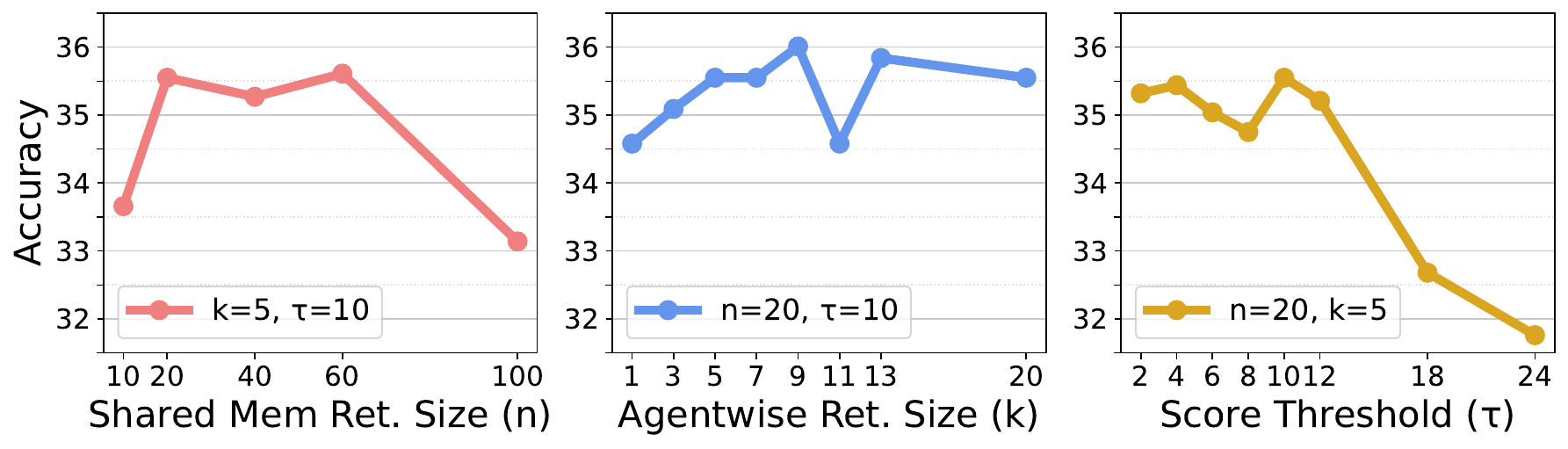}
    \caption{Ablation studies on shared memory retrieval size ($n$), agent-wise dynamic retrieval size ($k$), and score threshold ($\tau$).}
    \label{fig:hyperparameters}
\end{figure}

%% file: tab/qa_samples.tex
\begin{table}[h]
    \centering
    \caption{Social Interaction (SI) - Single Span category samples}
    \label{tab:qasample_si_ss}
    \scriptsize
    \begin{tabular}{@{}p{0.11\linewidth} p{0.88\linewidth}@{}}
        \toprule
        \rowcolor{gray!20} \textbf{Category} & \textbf{Social Interaction, Single Span} \\
        \midrule
        \textbf{Question} & \textbf{Who helped each other find scissors, and what were the scissors used for?} \\
        ~ & (A) Jake lent scissors to Shure, who used them for prepping coffee beans. \\
        ~ & \textbf{(B) Jake handed scissors to Katrina for her package and flower crafts.} \\
        ~ & (C) Tasha and Lucia found scissors to cut plastic wrappings for food. \\
        ~ & (D) Alice shared scissors with Katrina for cutting papers. \\
        ~ & (E) Lucia requested scissors from Tasha for assembling wooden shelves \\
        \\
        \textbf{Evidence} & 
        \makecell[l]{
        Source: Jake (Day4 12PM), Katrina (Day4 12 PM)\\
        \includegraphics[width=0.18\linewidth]{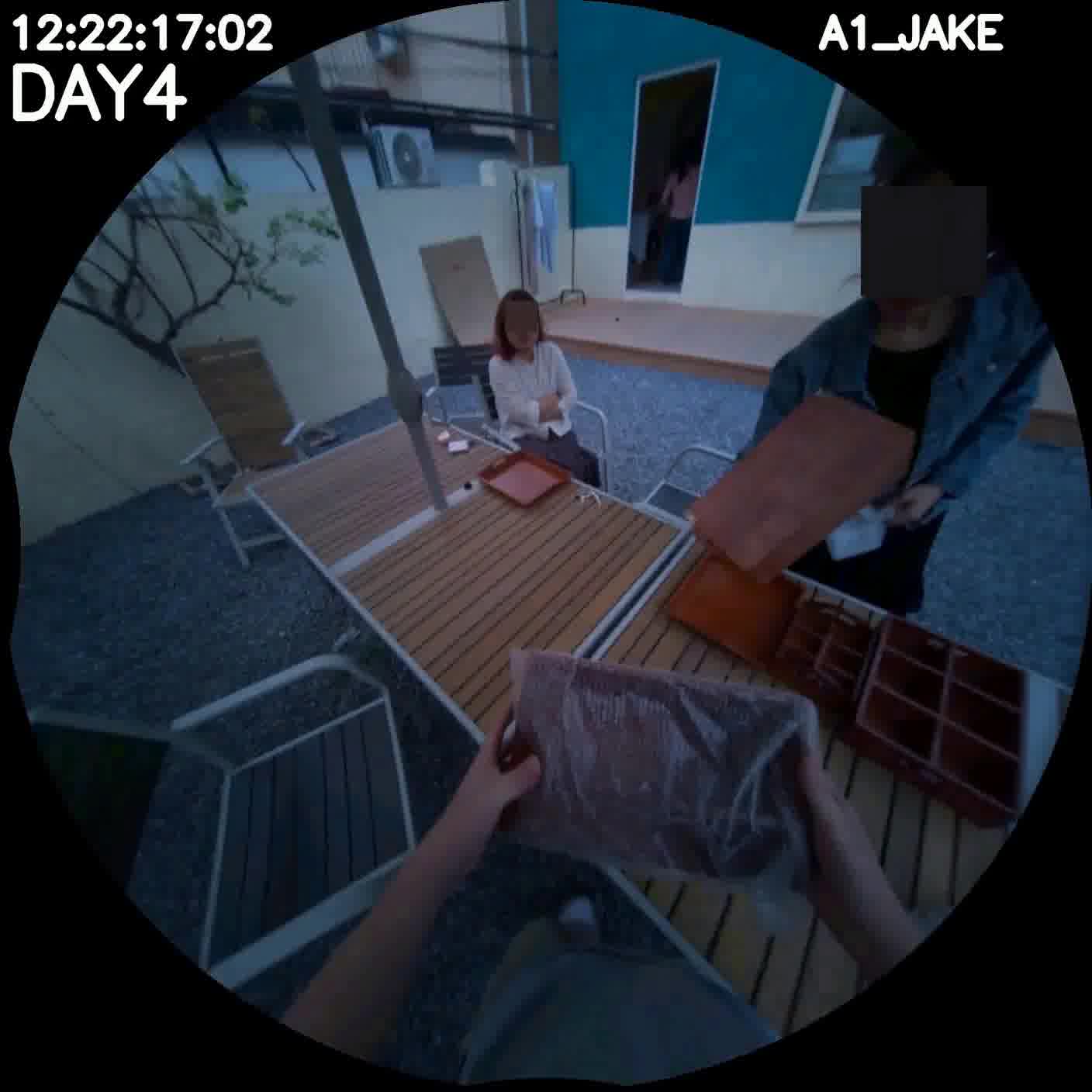} \hspace{1pt}
        \includegraphics[width=0.18\linewidth]{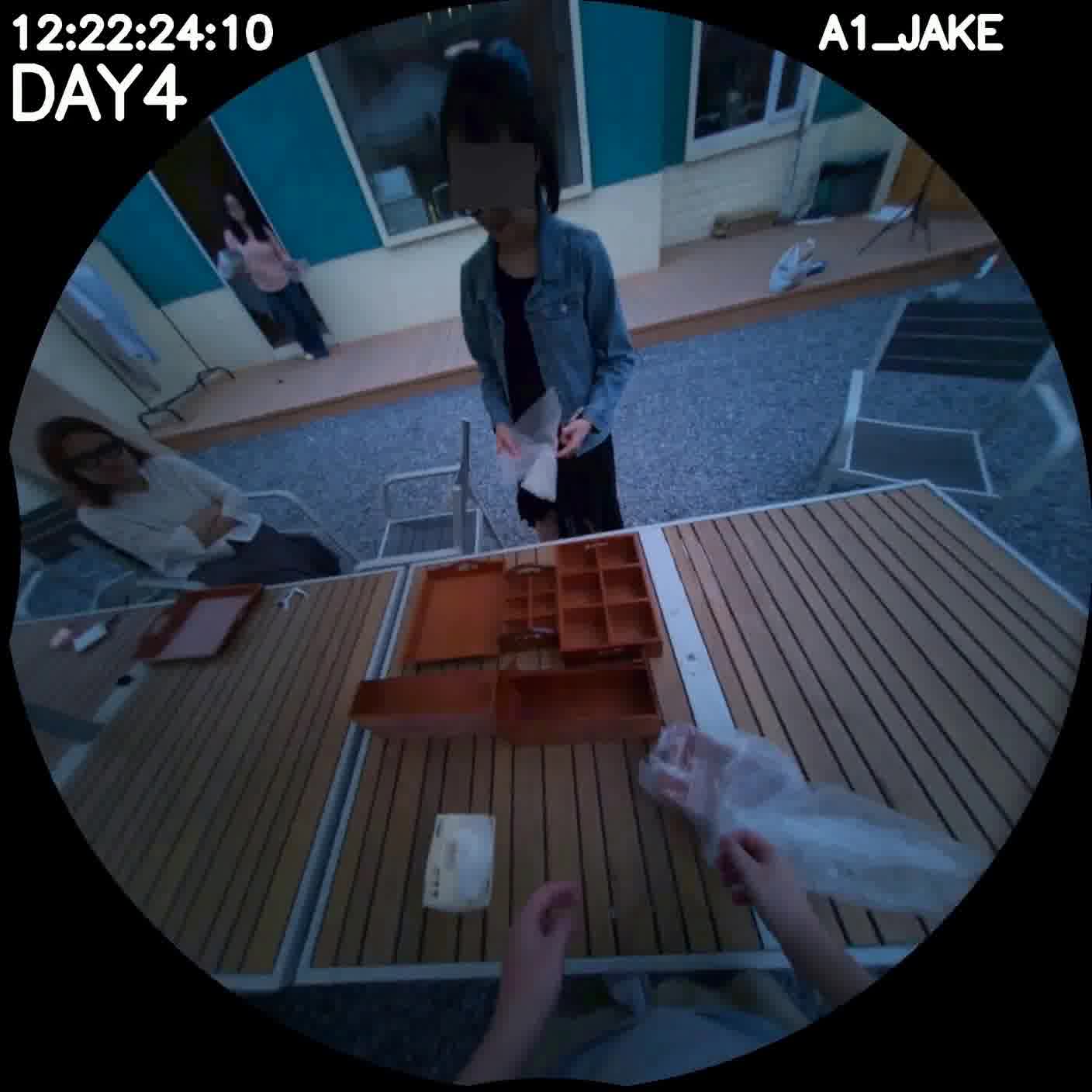} \hspace{1pt}
        \includegraphics[width=0.18\linewidth]{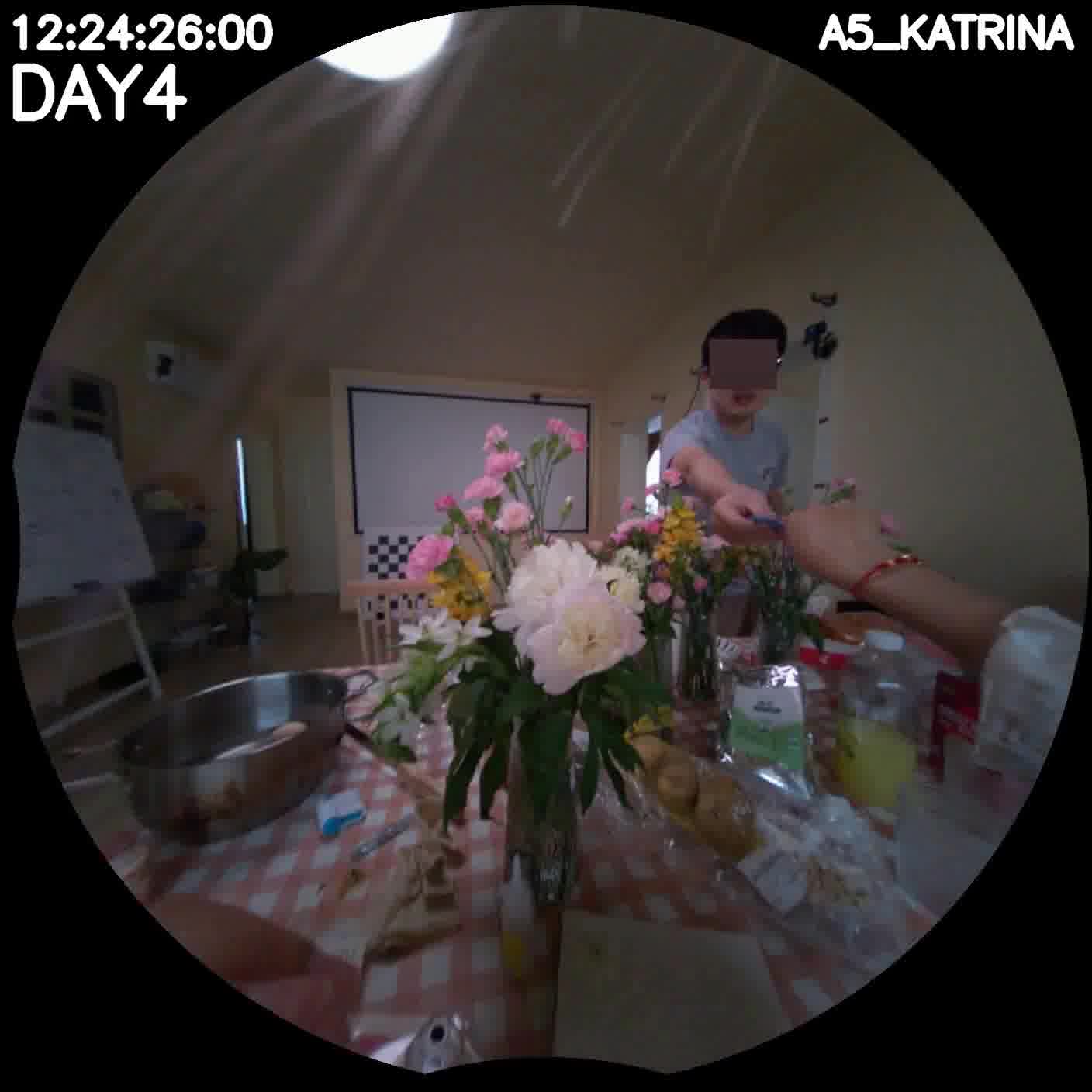} \hspace{1pt}
        \includegraphics[width=0.18\linewidth]{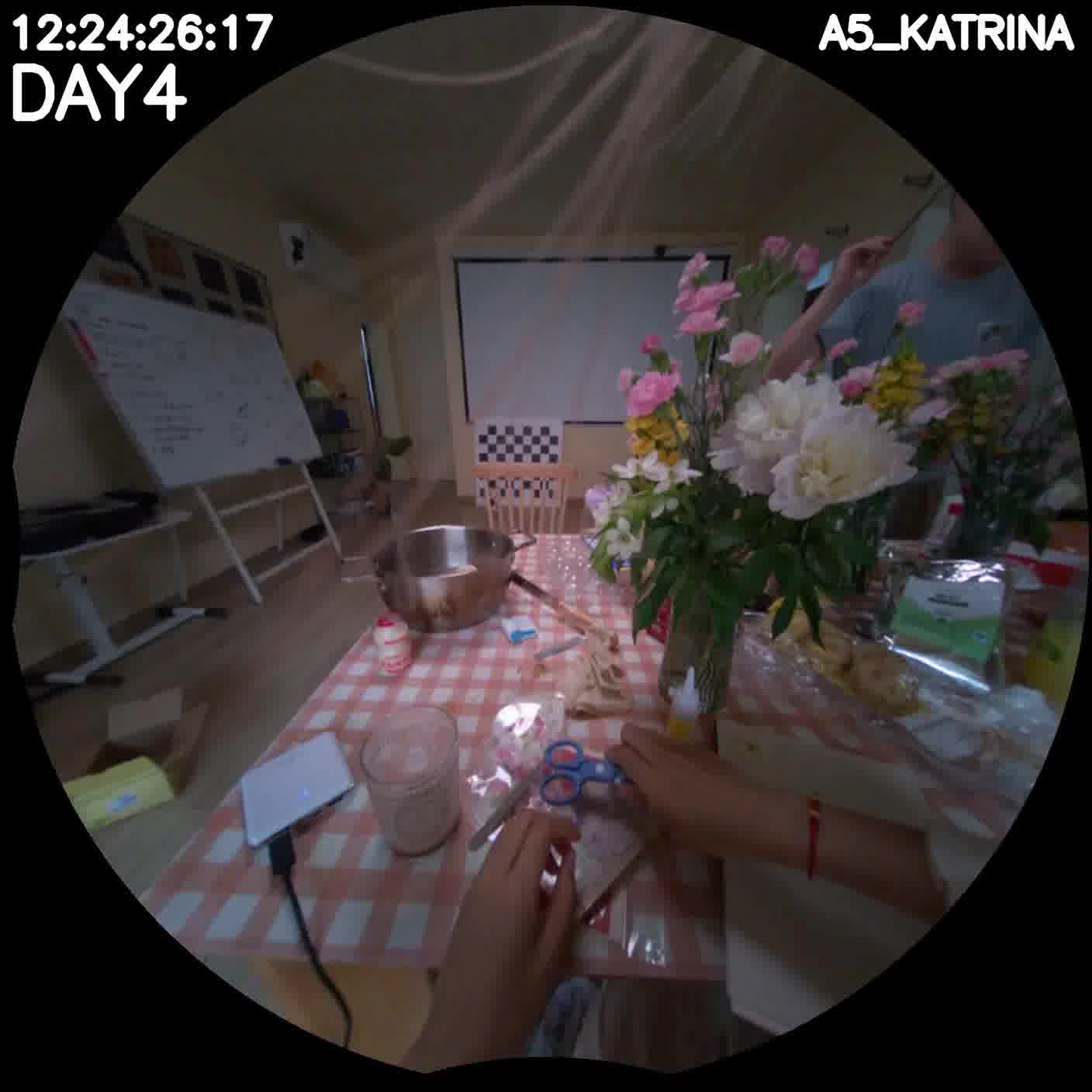} \hspace{1pt}
        \includegraphics[width=0.18\linewidth]{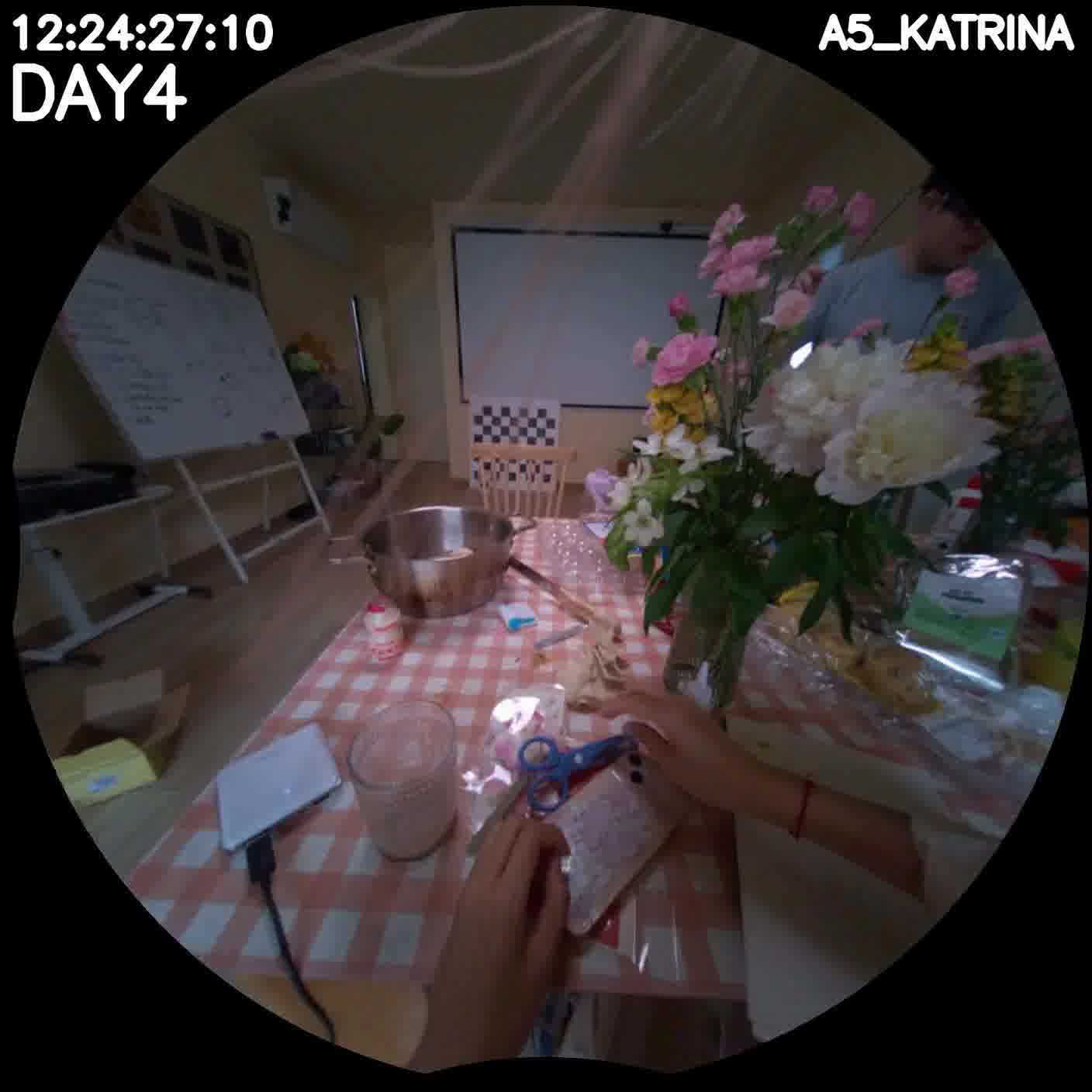}
        }
        \\
        \midrule
        \textbf{Question} & \textbf{Who asked for a power bank and how did Jake respond to their request?} \\
        ~ &  (A) Lucia borrowed a power bank, but Jake didn't have any left. \\
       ~ & (B) Nicous asked for a power bank, but Jake refused to let him unplug anything. \\
       ~ & (C) Shure asked for a power bank, and Jake offered heavy assistance. \\
       ~ & \textbf{(D) Tasha borrowed a power bank, and Jake offered to find another one.} \\
       ~ & (E) Alice needed a power bank, and Jake handed his own immediately.  \\
       \\
        \textbf{Evidence} & 
        \makecell[l]{
        Source: Jake (Day6 12PM)\\
        \includegraphics[width=0.18\linewidth]{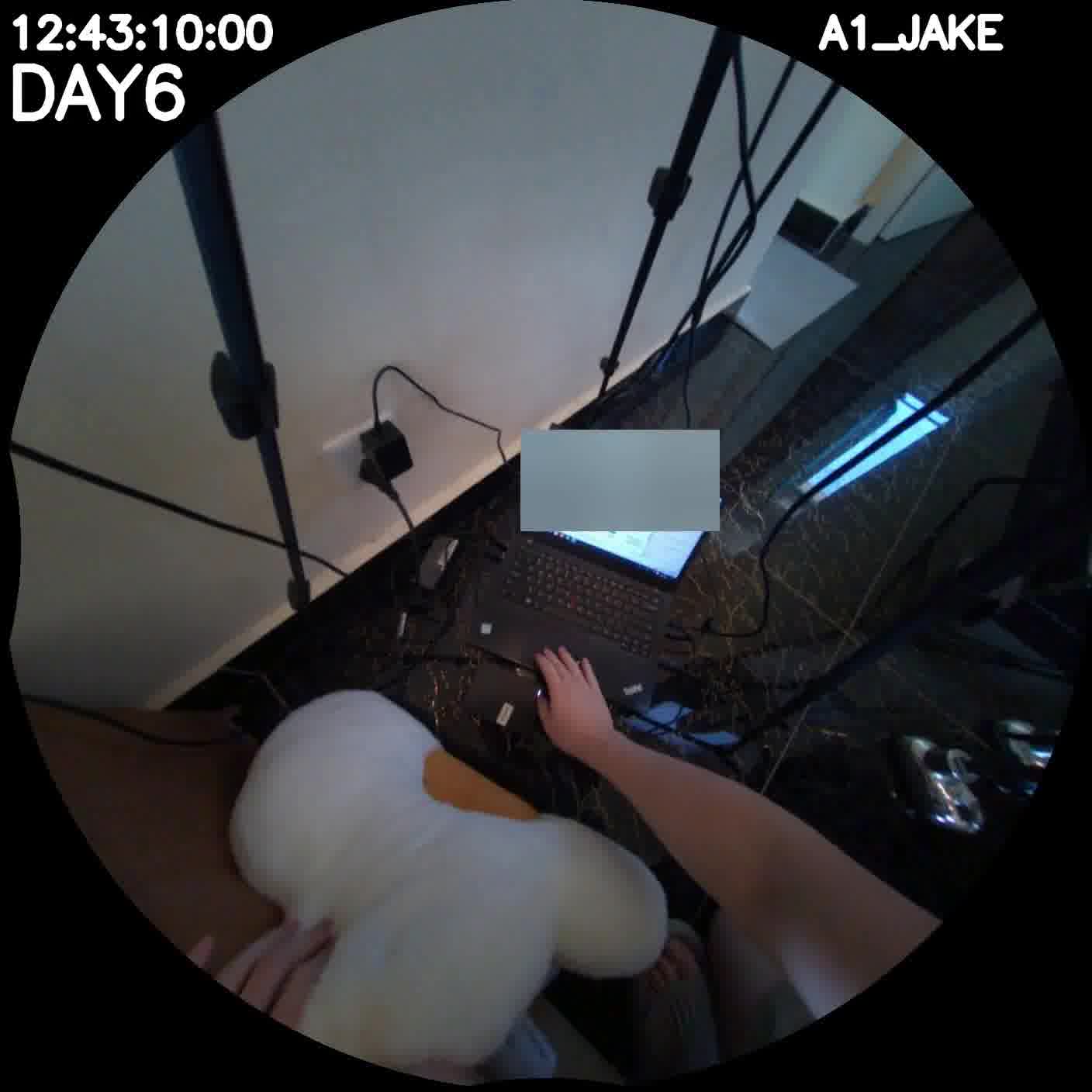} \hspace{1pt}
        \includegraphics[width=0.18\linewidth]{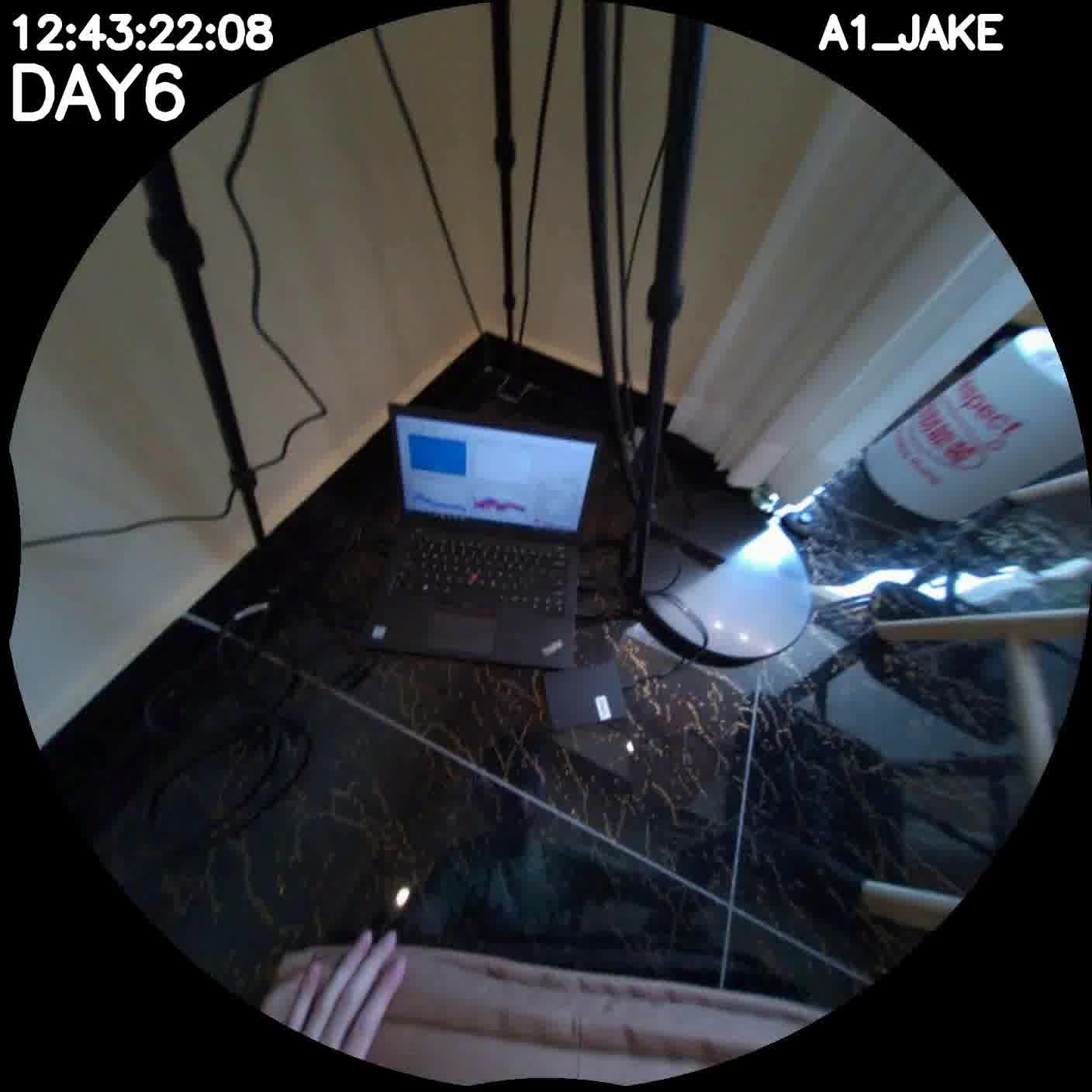} \hspace{1pt}
        \includegraphics[width=0.18\linewidth]{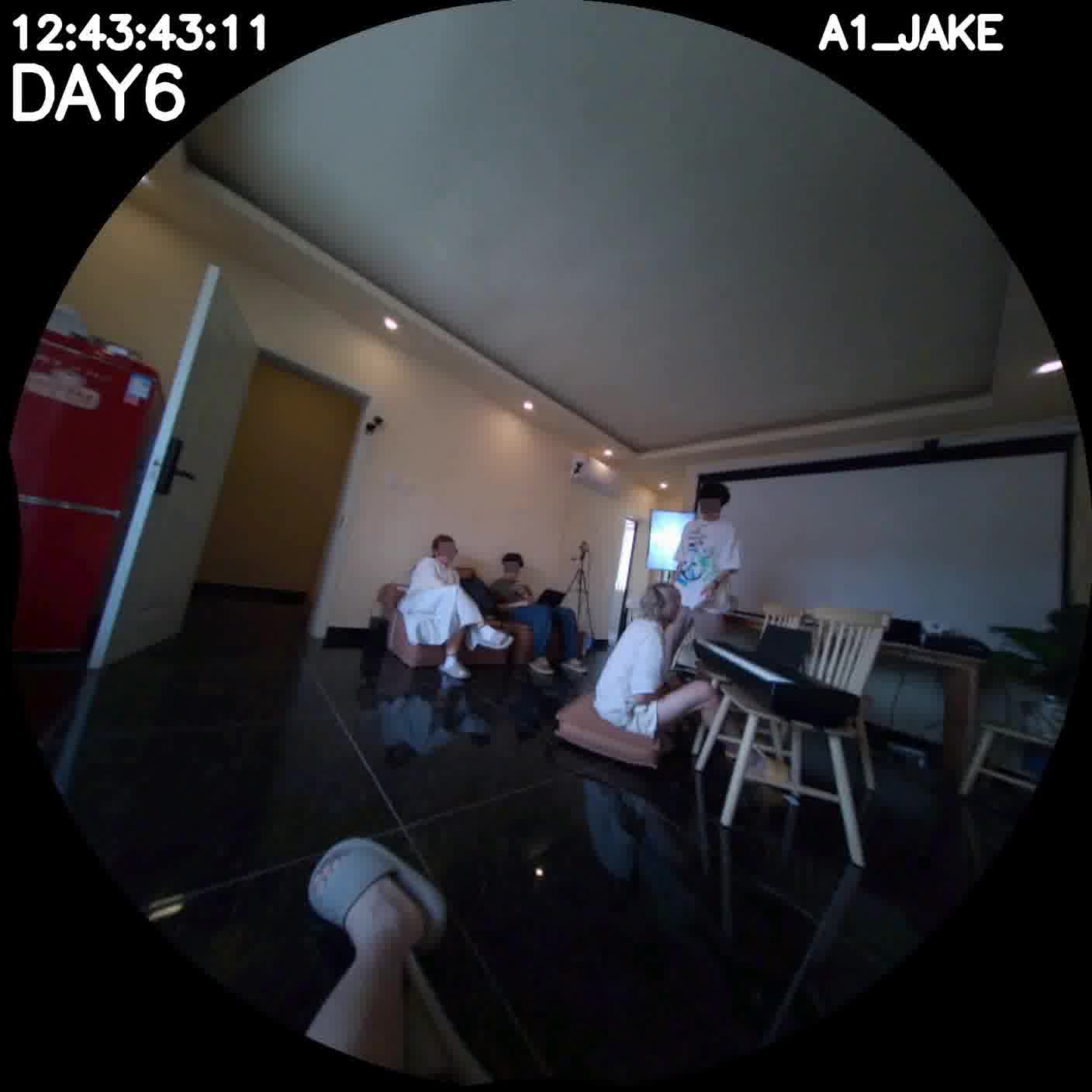}
        }
        \\
        \bottomrule
    \end{tabular}
\end{table}

\begin{table}[h]
    \centering
    \caption{Social Interaction (SI) - Multi Span category samples}
    \label{tab:qasample_si_ms}
    \scriptsize
    \adjustbox{width=\linewidth}{
    \begin{tabular}{@{}p{0.11\linewidth} p{0.88\linewidth}@{}}
        \toprule
        \rowcolor{gray!20}\textbf{Category} & \textbf{Social Interaction, Multi Span} \\
        \midrule
        \textbf{Question} & \textbf{How did egg tart impressions shift between the sight and bite?} \\
        ~ & (A) Alice mistook the egg tart for a muffin, then Shure found the egg tart tasty despite looks. \\
        ~ & \textbf{(B) Jake mistook the egg tart for a pancake, then Shure found the egg tart tasty despite looks.} \\
        ~ & (C) Alice mistook the egg tart for a muffin, then Shure found the egg tart tasty despite looks. \\
        ~ & (D) Jake mistook the egg tart for a pancake, then Tasha found the egg tart bland despite looks. \\
        ~ & (E) Alice mistook the egg tart for a muffin, then Tasha found the egg tart bland despite looks. \\
        \\
        \textbf{Evidence} & 
        \makecell[l]{
        Source: Jake (Day5 8PM), Alice (Day5 8PM), Lucia (Day5 8PM), Shure (Day6 5PM) \\
        \includegraphics[width=0.18\linewidth]{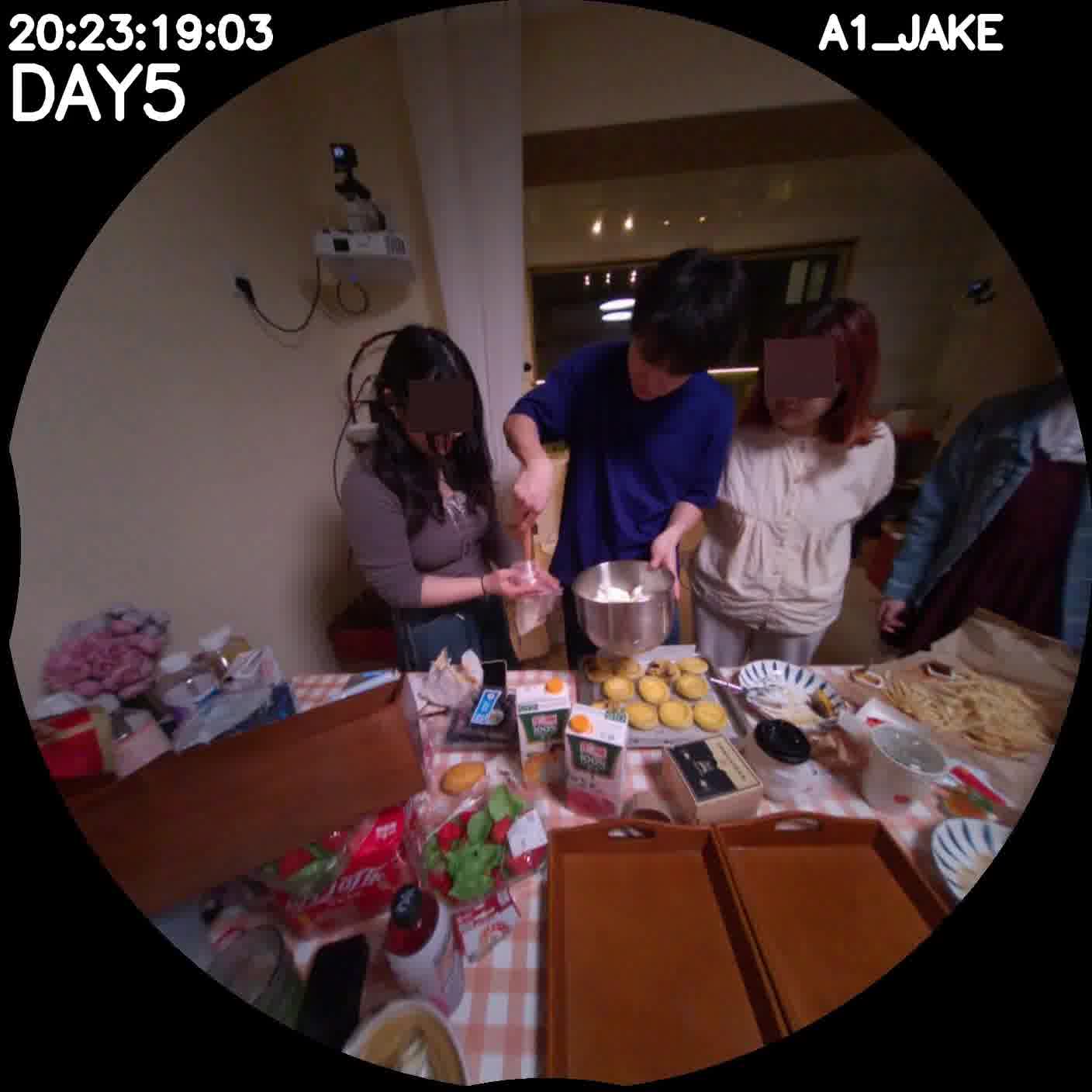} \hspace{1pt}
        \includegraphics[width=0.18\linewidth]{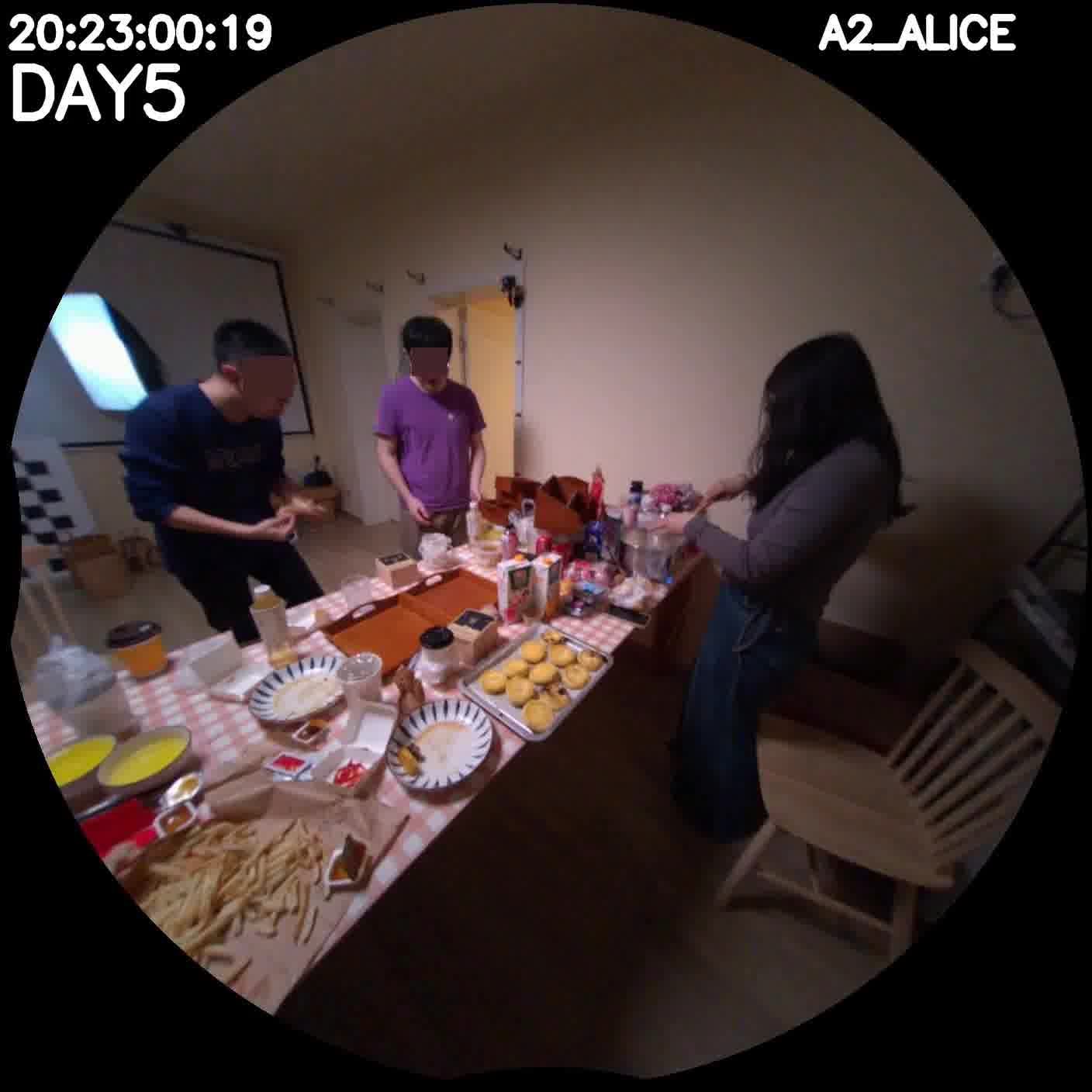} \hspace{1pt}
        \includegraphics[width=0.18\linewidth]{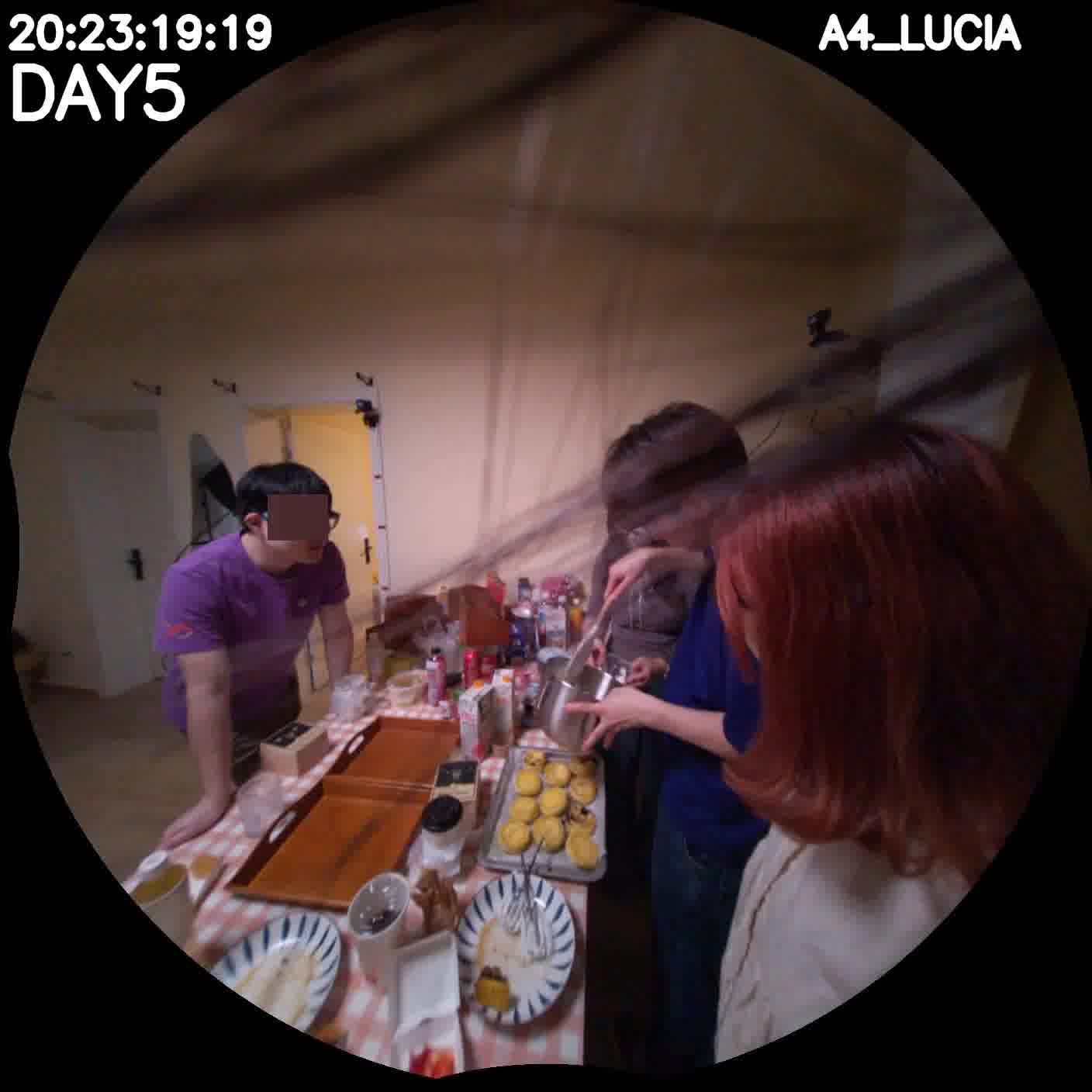} \hspace{1pt}
        \includegraphics[width=0.18\linewidth]{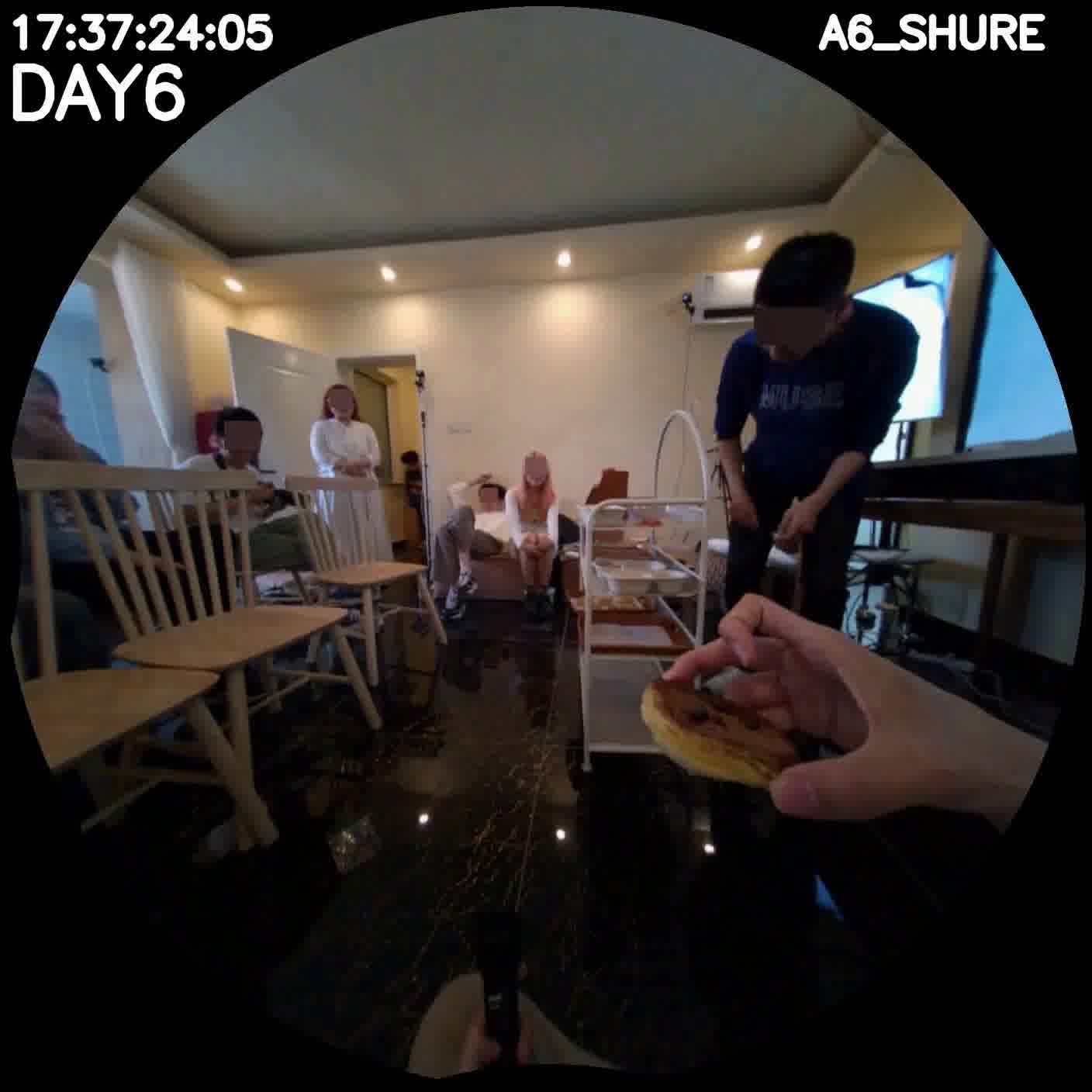} \hspace{1pt}
        \includegraphics[width=0.18\linewidth]{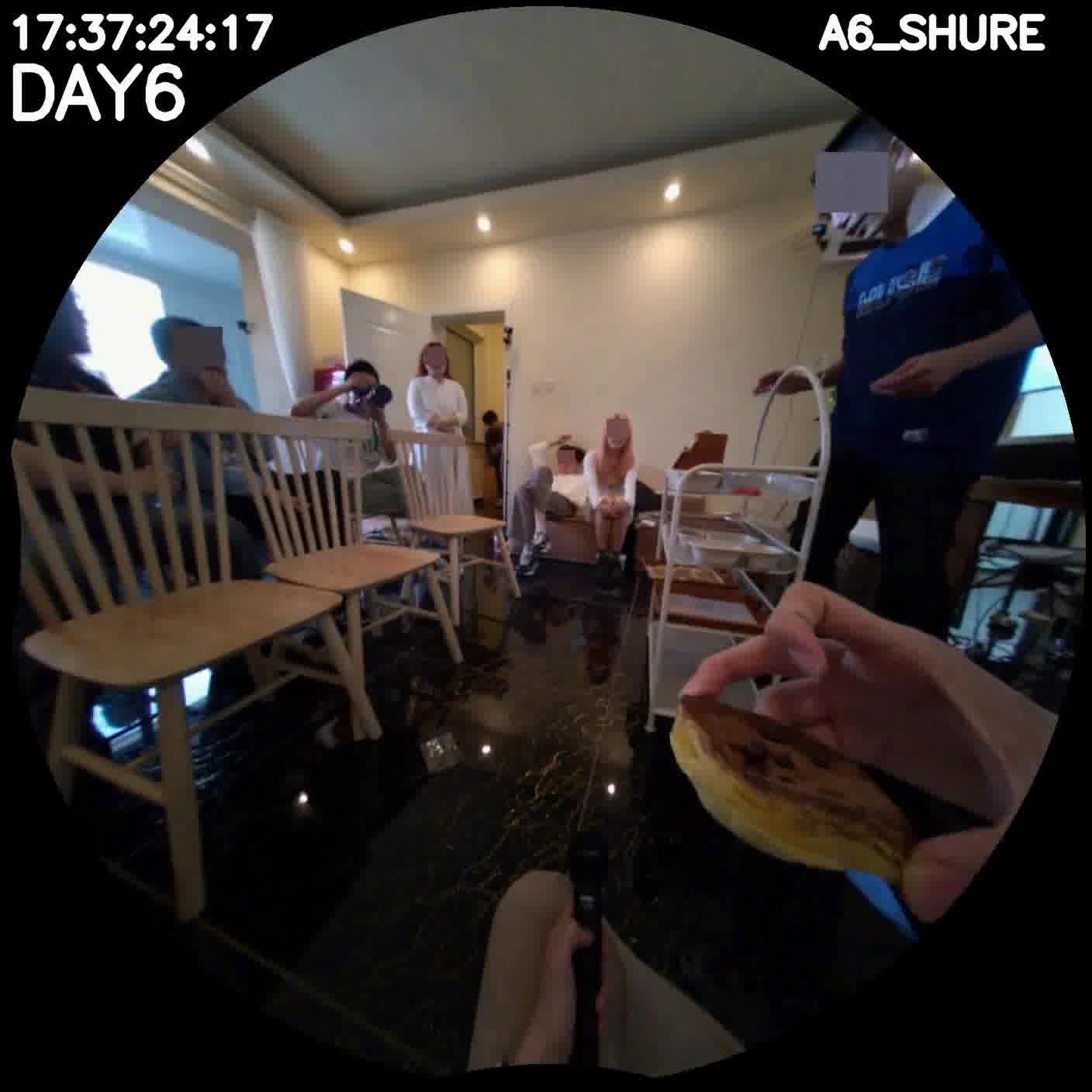}}
        \\
        \midrule
        \textbf{Question} & \textbf{What coffee discussions happened in both situations?} \\
        ~ & (A) Jake proposed restocking coffee beans and equipment and later Lucia offered packing help after Lucia mentioned coffee. \\ 
        ~ & \textbf{(B) Shure proposed restocking coffee beans and equipment, and later Alice offered packing help after Katrina mentioned coffee.} \\
        ~ & (C) Jake proposed restocking coffee beans and equipment, and later Lucia offered packing help after Lucia mentioned coffee. \\
        ~ & (D) Jake proposed restocking coffee beans and equipment, and later Alice offered packing help after Katrina mentioned coffee. \\
        ~ & (E) Shure proposed restocking coffee beans and equipment, and later Lucia offered packing help after Lucia mentioned coffee. \\
        \\
        \textbf{Evidence} & 
        \makecell[l]{
        Source: Jake (Day1 1PM), Shure (Day 1PM), Alice (Day7 6PM), Katrina (Day7 6PM)\\
        \includegraphics[width=0.18\linewidth]{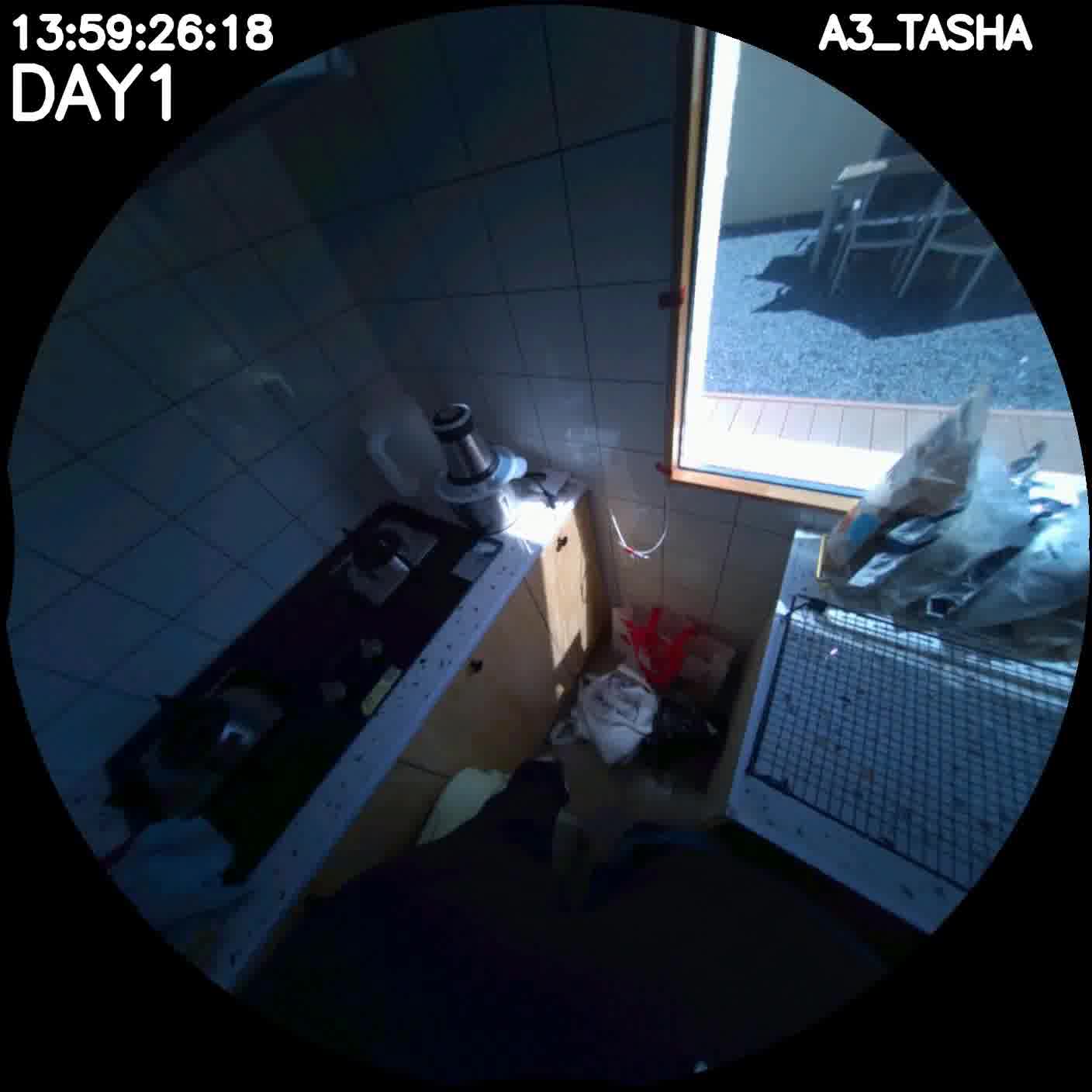} \hspace{1pt}
        \includegraphics[width=0.18\linewidth]{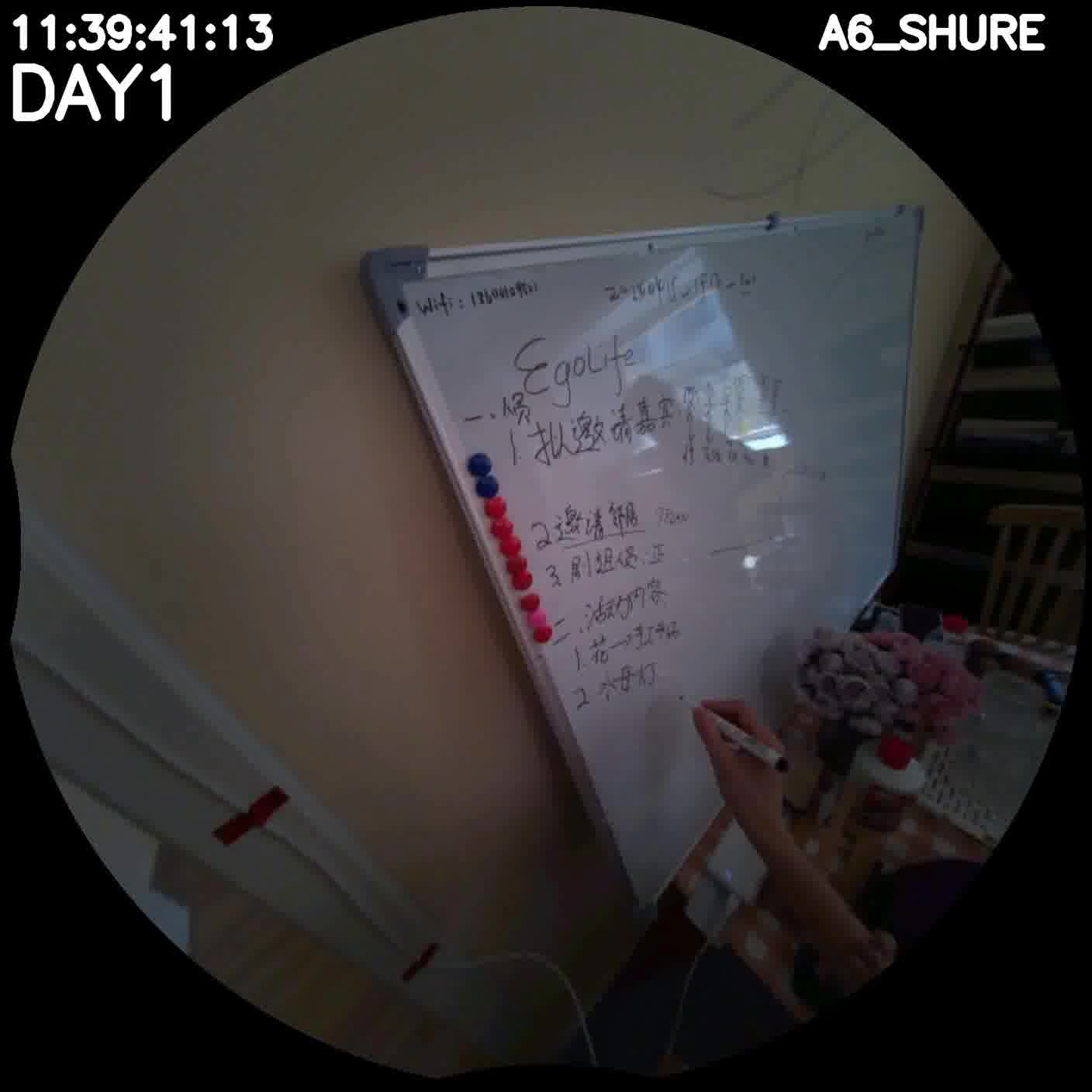} \hspace{1pt}
        \includegraphics[width=0.18\linewidth]{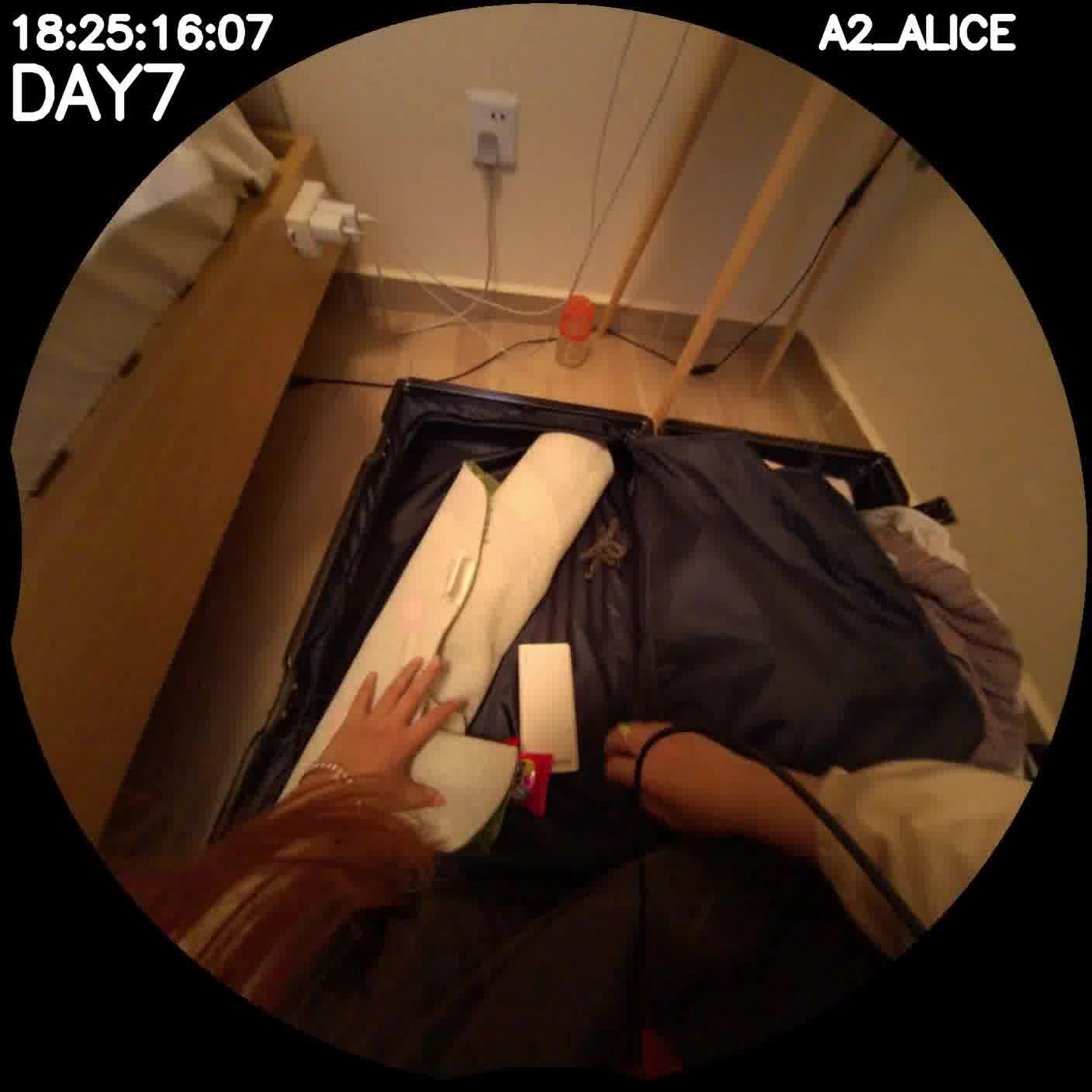} \hspace{1pt}
        \includegraphics[width=0.18\linewidth]{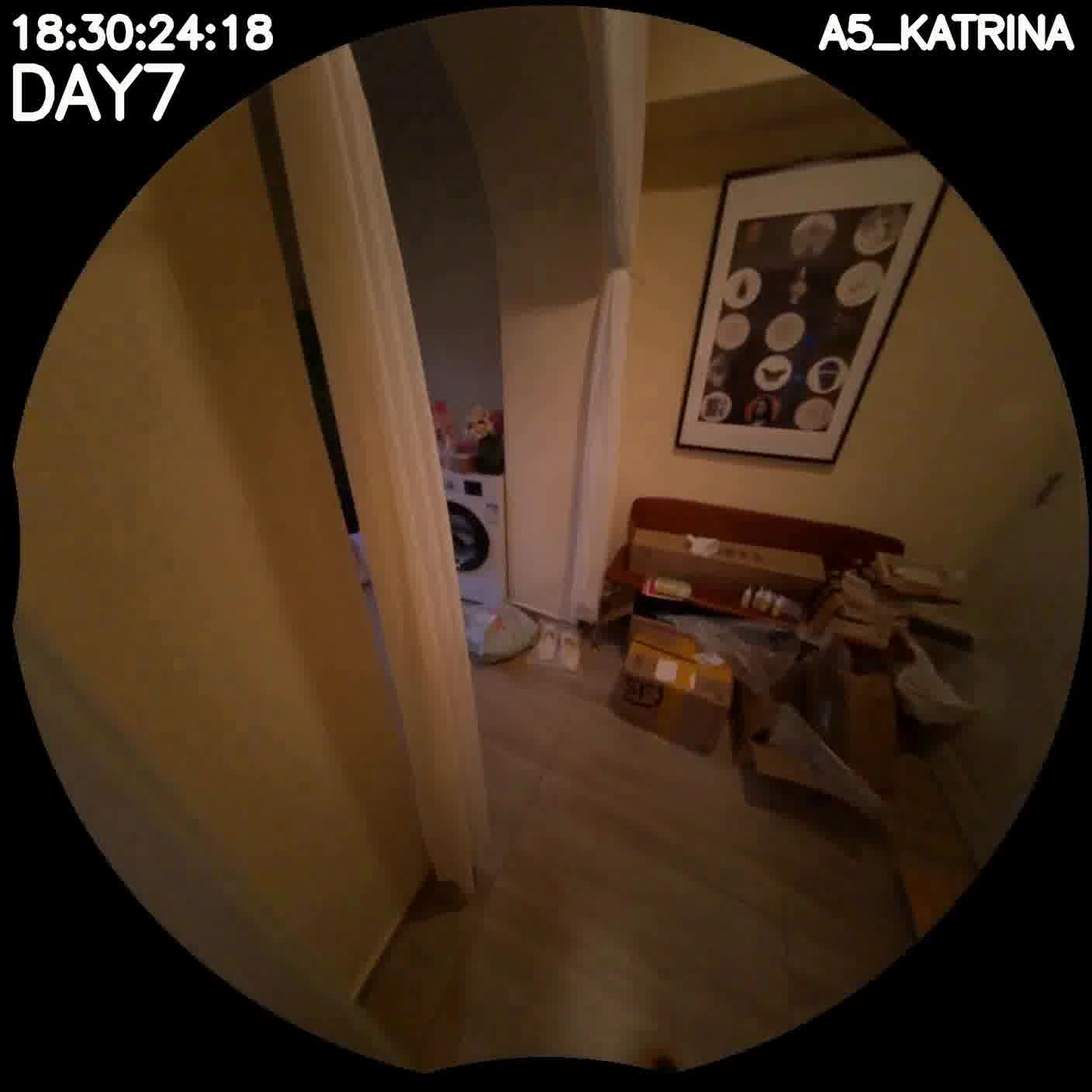} \hspace{1pt}
        \includegraphics[width=0.18\linewidth]{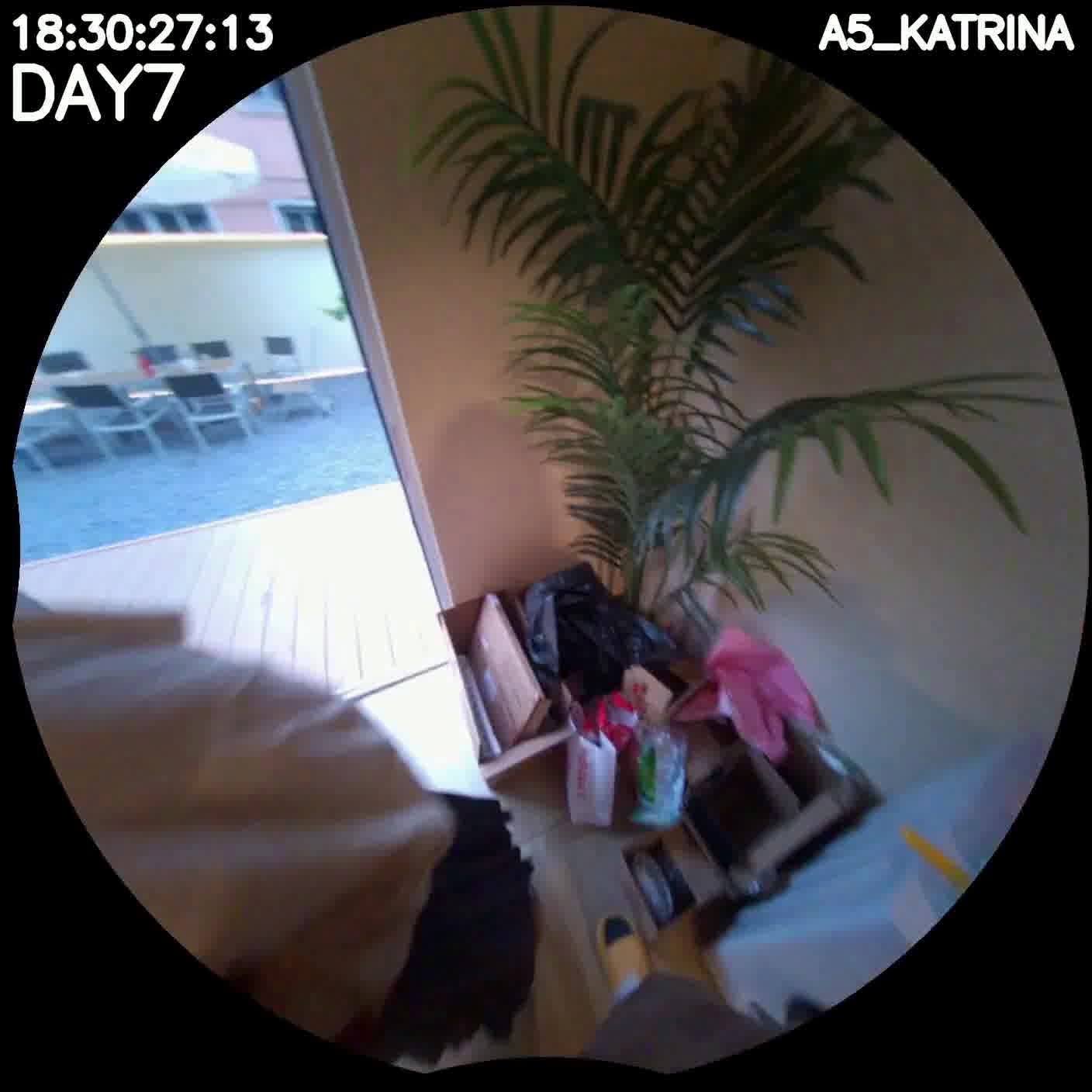} 
        }
        \\
        \bottomrule
    \end{tabular}
    }
\end{table}

\begin{table}[h]
    \centering
    \caption{Task Coordination (TC) - Single span category samples}
    \label{tab:qasample_tc_ss}
    \scriptsize
    \begin{tabular}{@{}p{0.11\linewidth} p{0.88\linewidth}@{}}
    \toprule
        \rowcolor{gray!20} \textbf{Category} & \textbf{Task Coordination, Single Span} \\
        \midrule
        \textbf{Question} & \textbf{How did they resolve concerns about the safety of the power strip?} \\
        ~ & (A) Lucia conducted a test and decided it was fine for usage. \\
       ~ & (B) Shure insisted safety wasn't a big issue and moved forward without discussion. \\
       ~ & (C) Shure reassured everyone it was brand-new and safe to use. \\
       ~ & (D) Katrina brought over a different power strip for better safety. \\
       ~ & \textbf{(E) Jake and Alice discussed using a reputable brand and reassured others.} \\
       \\
        \textbf{Evidence} & 
        \makecell[l]{
        Source: Jake (Day2 6PM), Alice (Day 6PM) \\
        \includegraphics[width=0.18\linewidth]{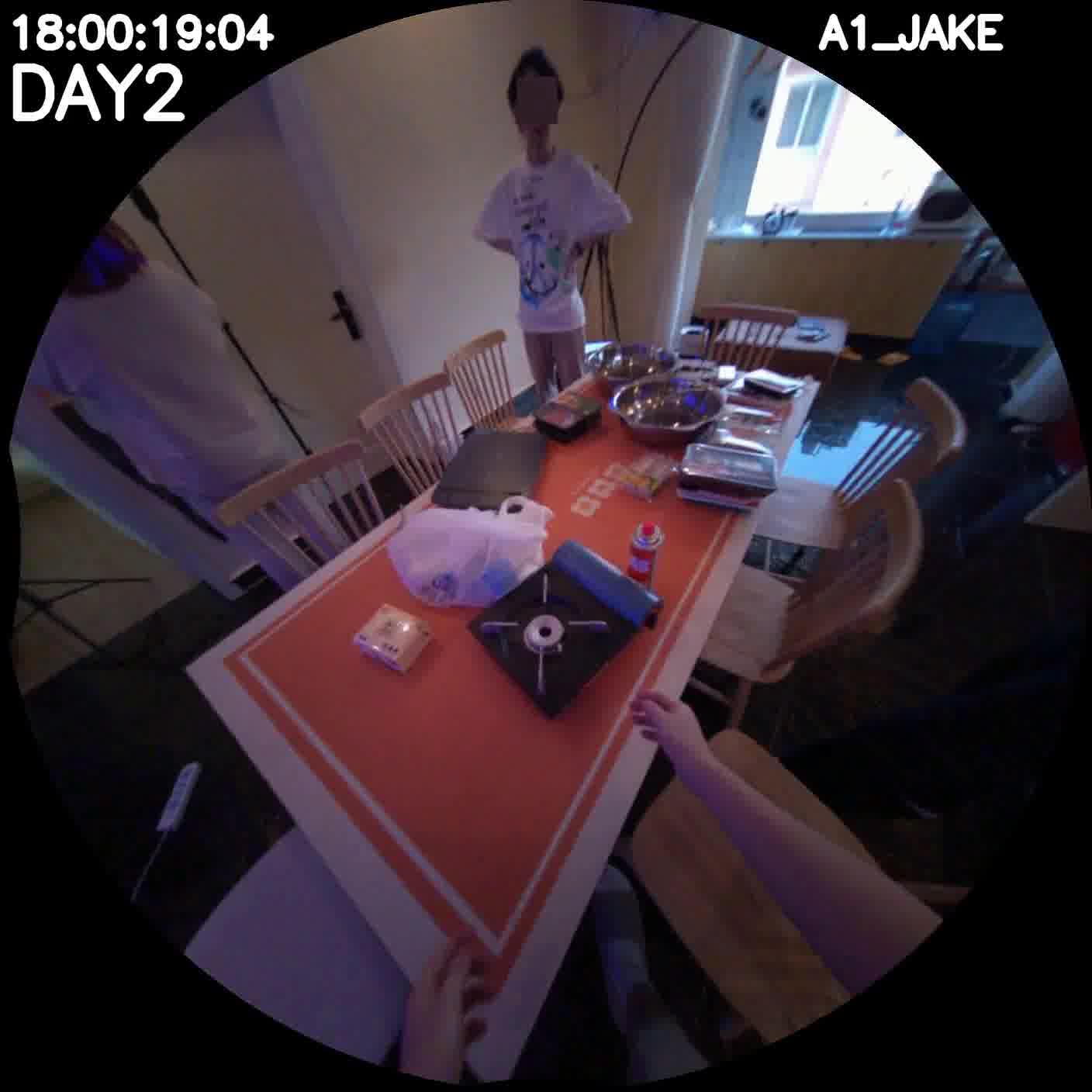} \hspace{1pt}
        \includegraphics[width=0.18\linewidth]{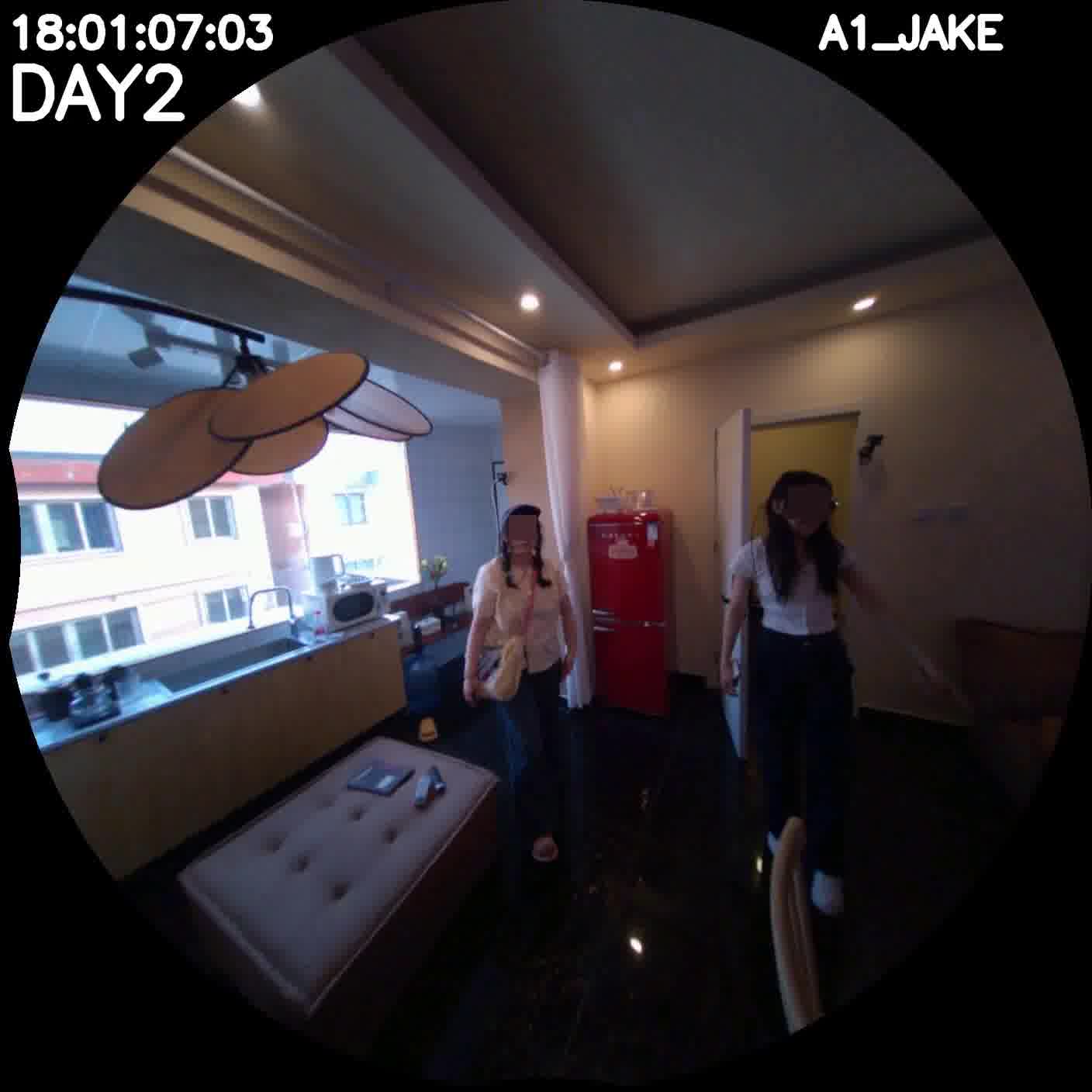} \hspace{1pt}
        \includegraphics[width=0.18\linewidth]{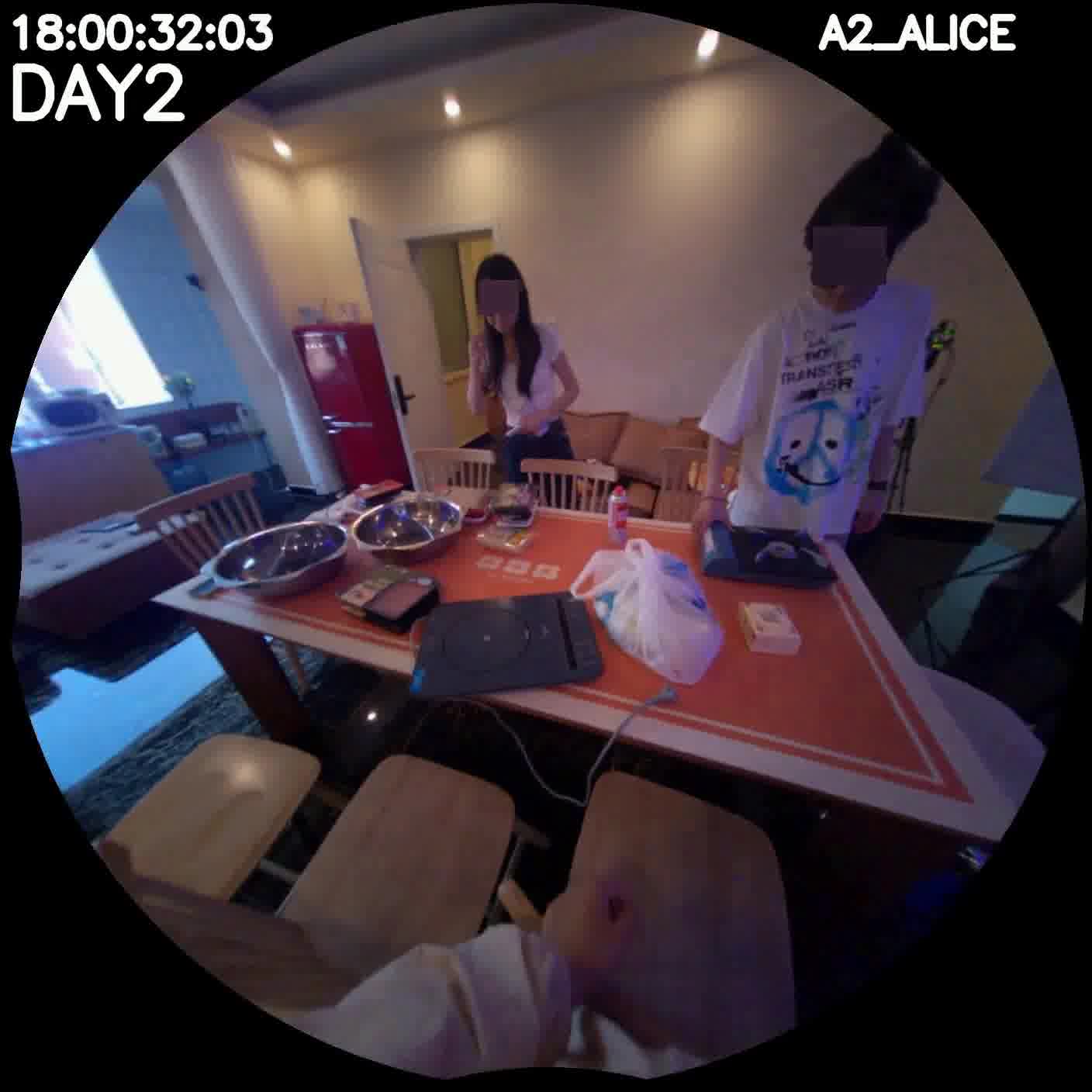} \hspace{1pt}
        \includegraphics[width=0.18\linewidth]{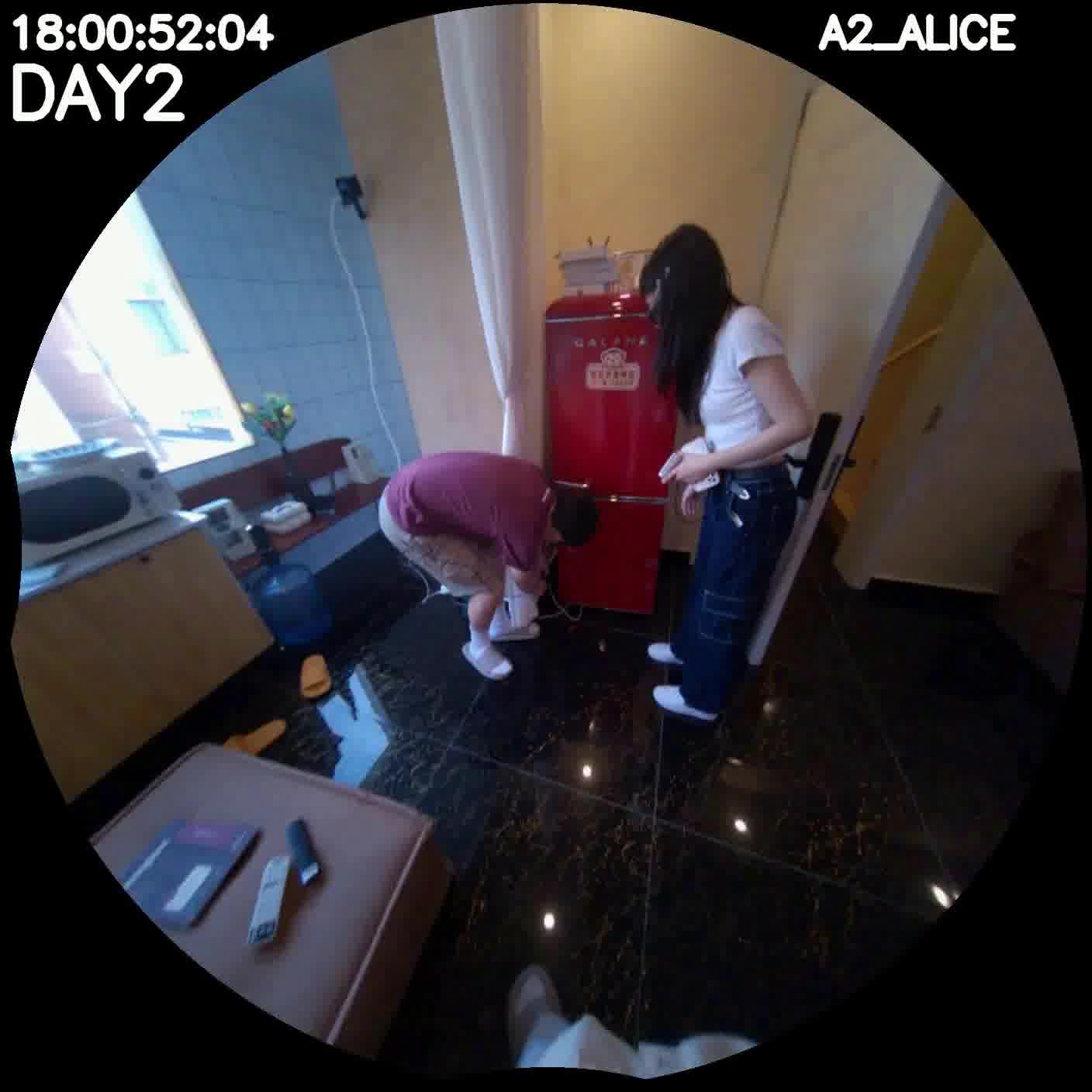}
        }
        \\
        \midrule
        \textbf{Question} & \textbf{When they were trying to light the charcoal, how did the group coordinate and divide tasks to make it work?} \\
        ~ & \textbf{(A) Jake shielded the fire with a cardboard while Shure carefully transported it outside.} \\
       ~ & (B) Alice ignited the fire using dry leaves while Katrina monitored the process. \\
       ~ & (C) Lucia set up a fan to make the fire stronger while Tasha prepared water for safety. \\
       ~ & (D) Jake and Lucia used chemicals to ignite the charcoal while Alice held a bucket nearby. \\
       ~ & (E) Tasha managed the cooking station while Shure brought wood for the fire. \\
               \\
        \textbf{Evidence} & 
        \makecell[l]{
        Source: Jake (Day3 8PM) \\
        \includegraphics[width=0.18\linewidth]{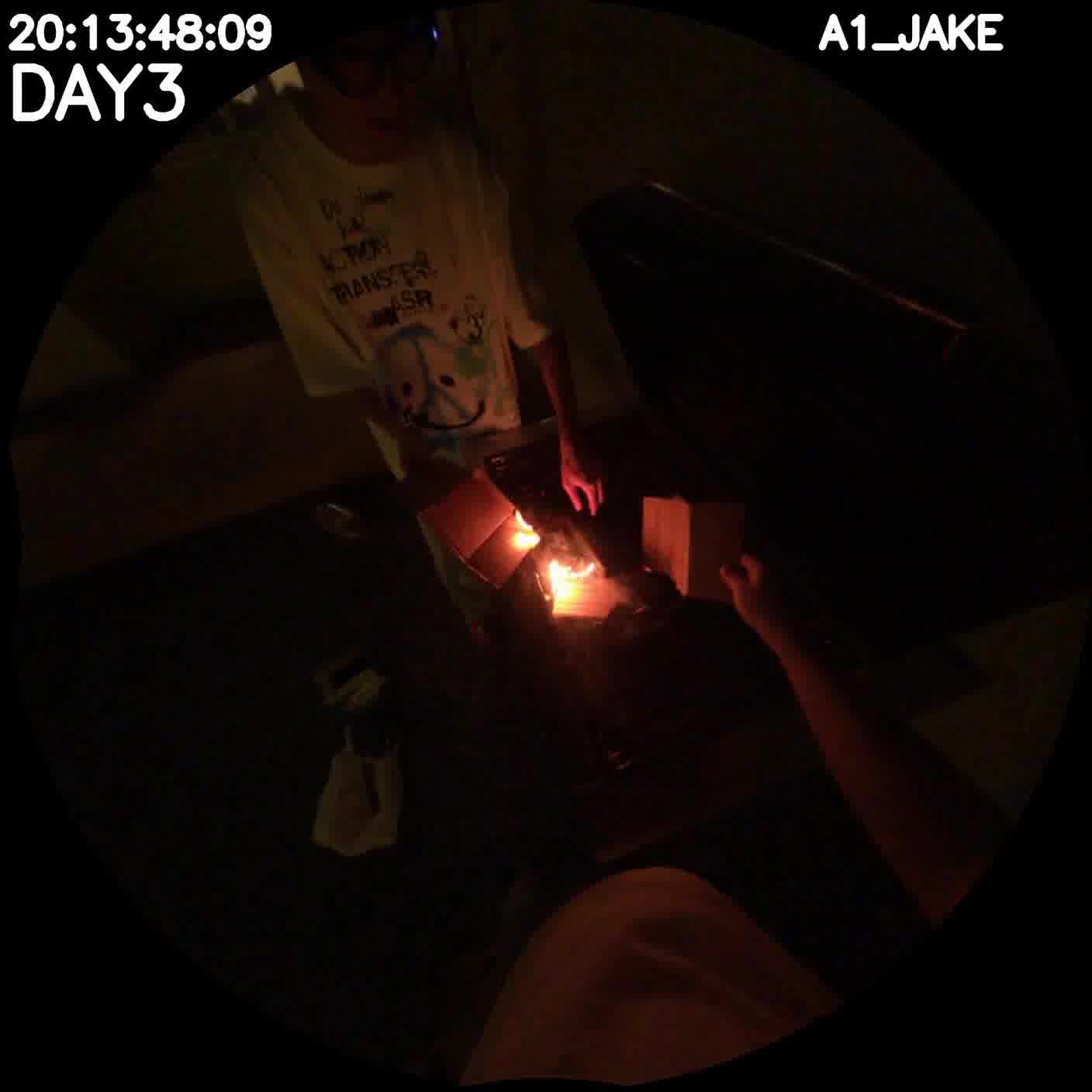} \hspace{1pt}
        \includegraphics[width=0.18\linewidth]{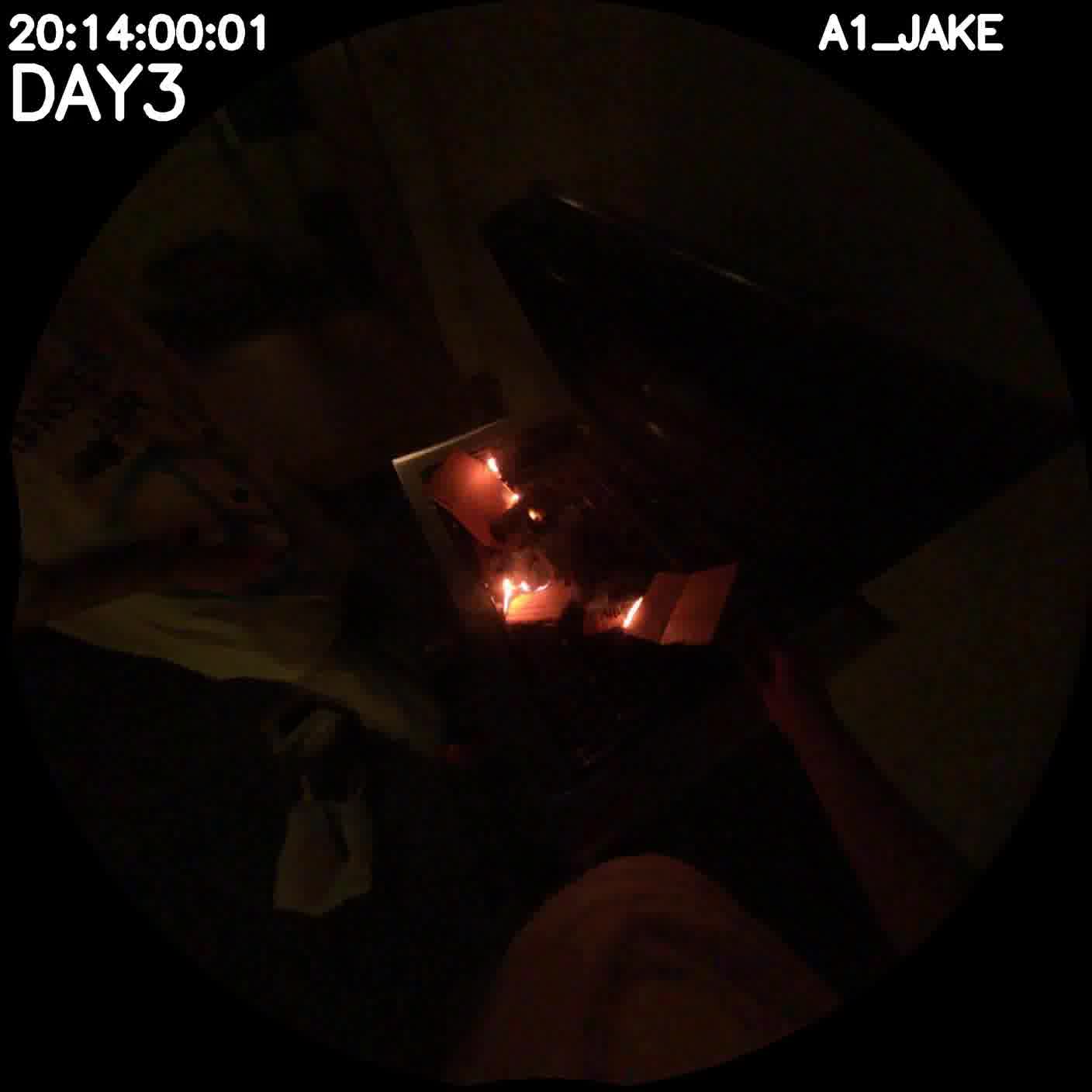} \hspace{1pt}
        \includegraphics[width=0.18\linewidth]{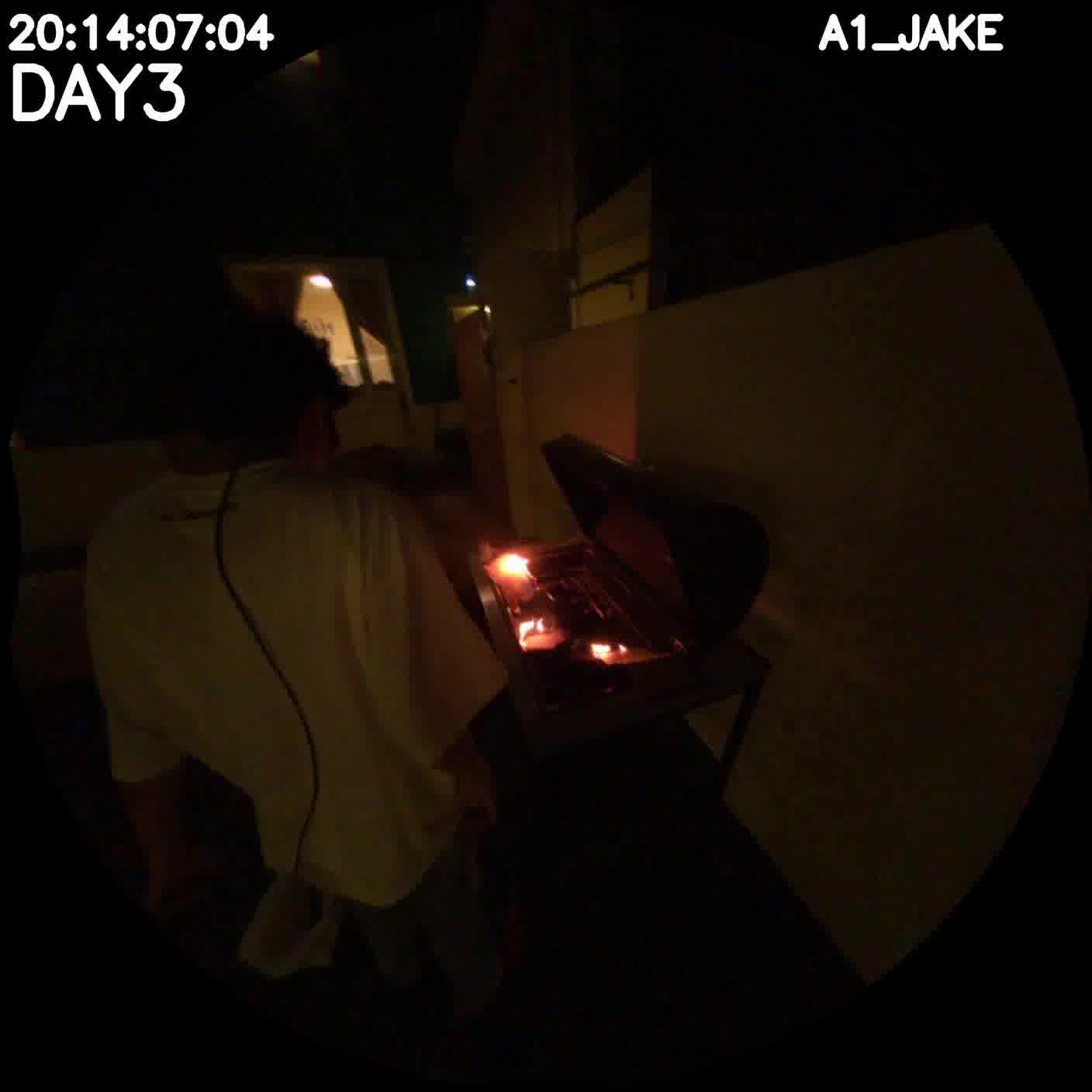} \hspace{1pt}
        \includegraphics[width=0.18\linewidth]{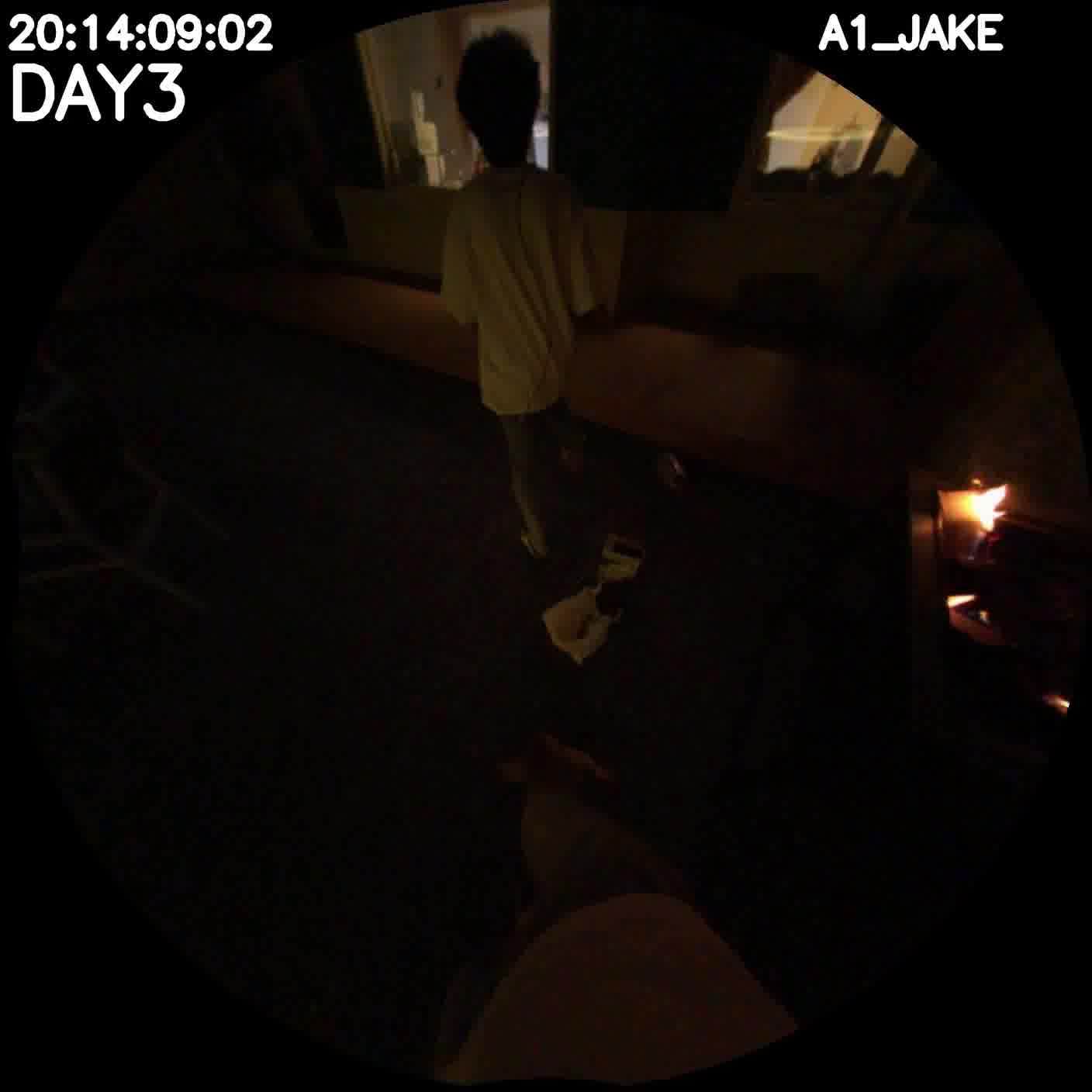}
        }
        \\
        \bottomrule
    \end{tabular}
\end{table}

\begin{table}[h]
    \centering
    \caption{Task Coordination (TC) - Multi span category samples}
    \label{tab:qasample_tc_ms}
    \scriptsize
    \begin{tabular}{@{}p{0.11\linewidth} p{0.88\linewidth}@{}}
        \toprule
        \rowcolor{gray!20} \textbf{Category} & \textbf{Task Coordination, Multi Span} \\
        \midrule
        \textbf{Question} & \textbf{In the flower tasks, what roles did participants take in different events?} \\
        ~ &  (A) Jake observed and asked clarifying questions, later arranged the bouquets, while Katrina handled hands-on trimming and shaping. \\
       ~ & (B) Alice observed and asked clarifying questions, later asked about flower types, while Katrina handled hands-on trimming and shaping. \\
       ~ & \textbf{(C) Jake observed and asked clarifying questions, later asked about flower types, while Shure handled hands-on trimming and shaping.} \\
       ~ & (D) Alice observed and asked clarifying questions, later asked about flower types, while Shure handled hands-on trimming and shaping. \\
       ~ & (E) Jake observed and asked clarifying questions, later asked about flower types, while Katrina handled hands-on trimming and shaping. \\
       \\
        \textbf{Evidence} & 
        \makecell[l]{
        Source: Jake (Day1 11AM), JAKE (Day2 4PM), Shure (Day4 4PM)\\
        \includegraphics[width=0.18\linewidth]{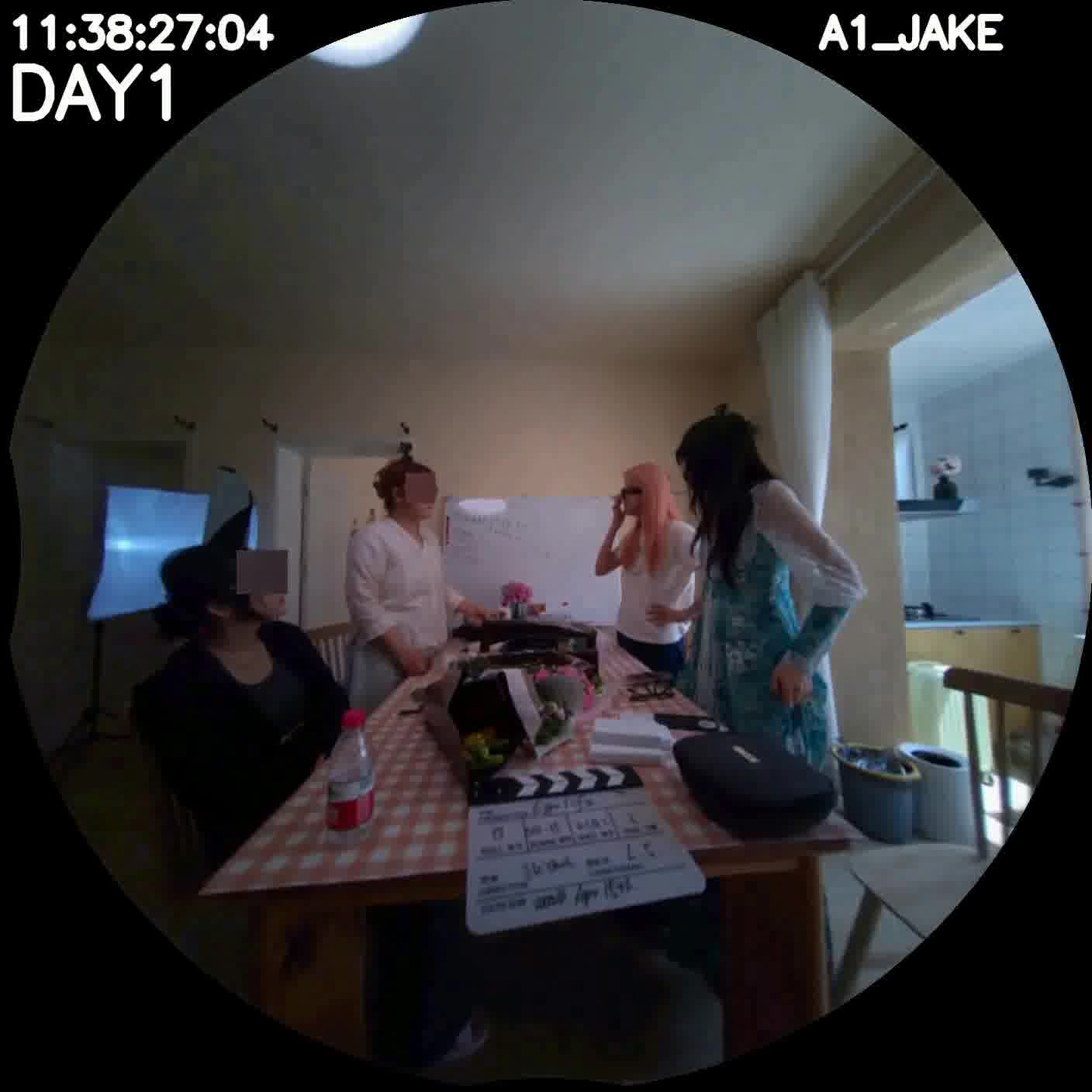} \hspace{1pt}
        \includegraphics[width=0.18\linewidth]{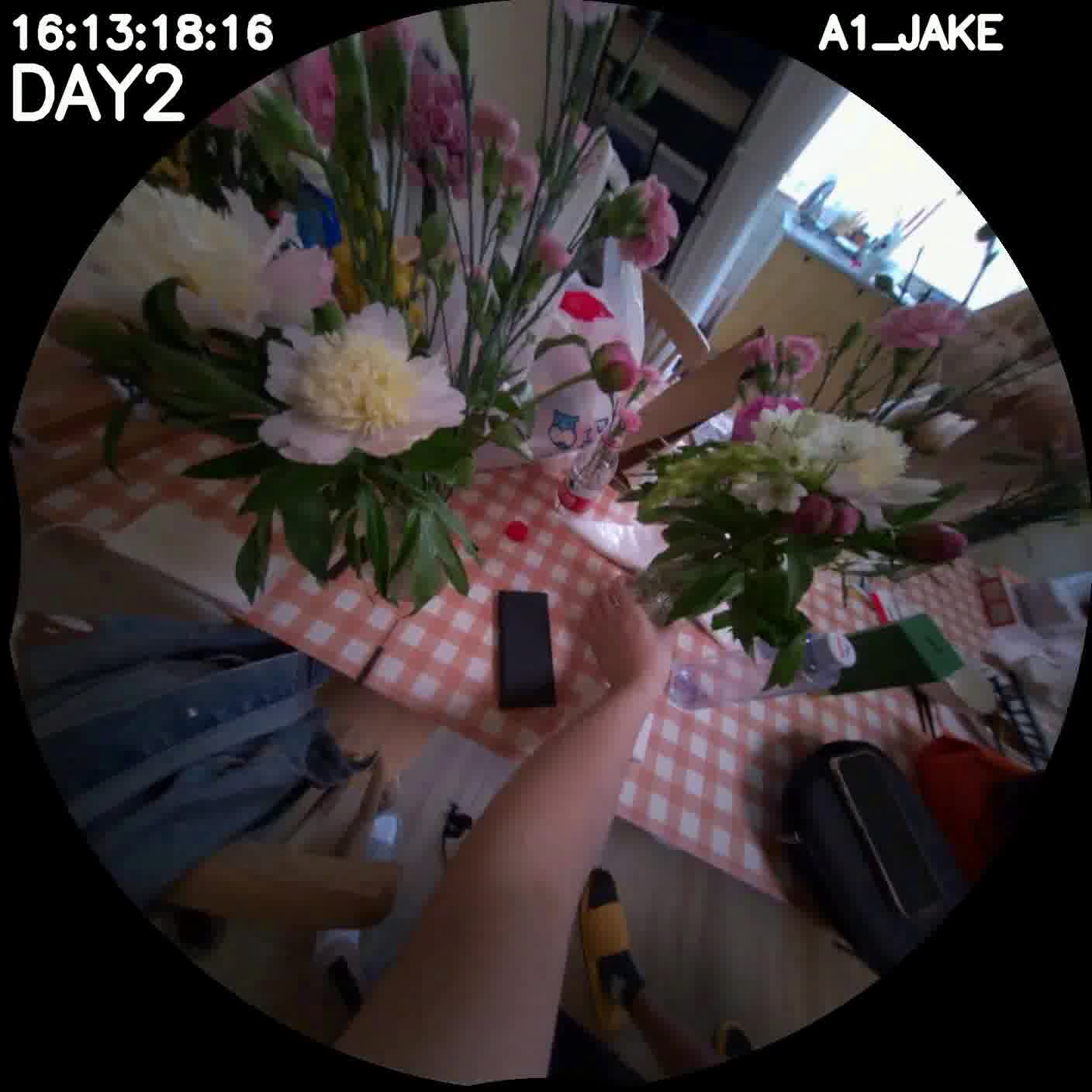} \hspace{1pt}
        \includegraphics[width=0.18\linewidth]{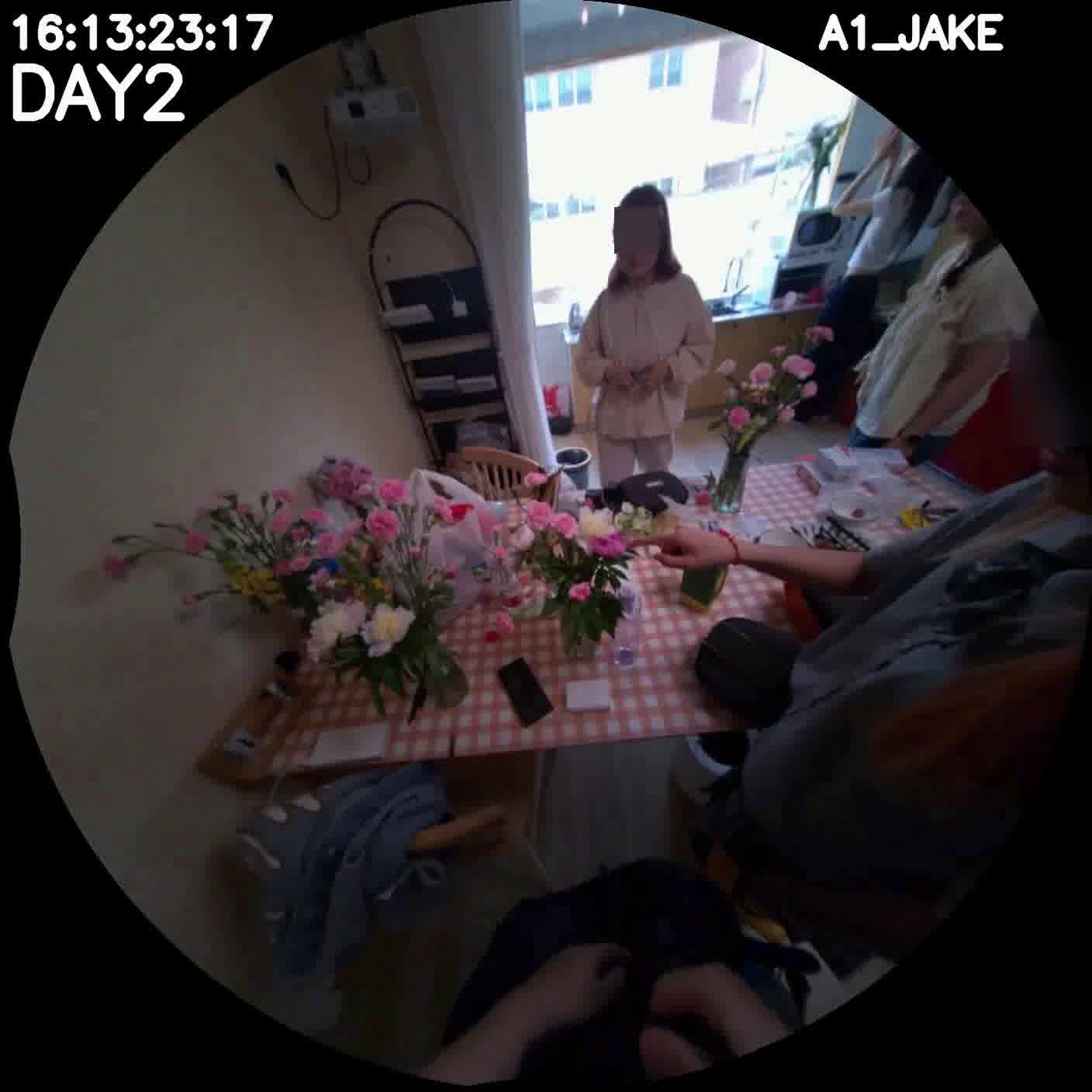} \hspace{1pt}
        \includegraphics[width=0.18\linewidth]{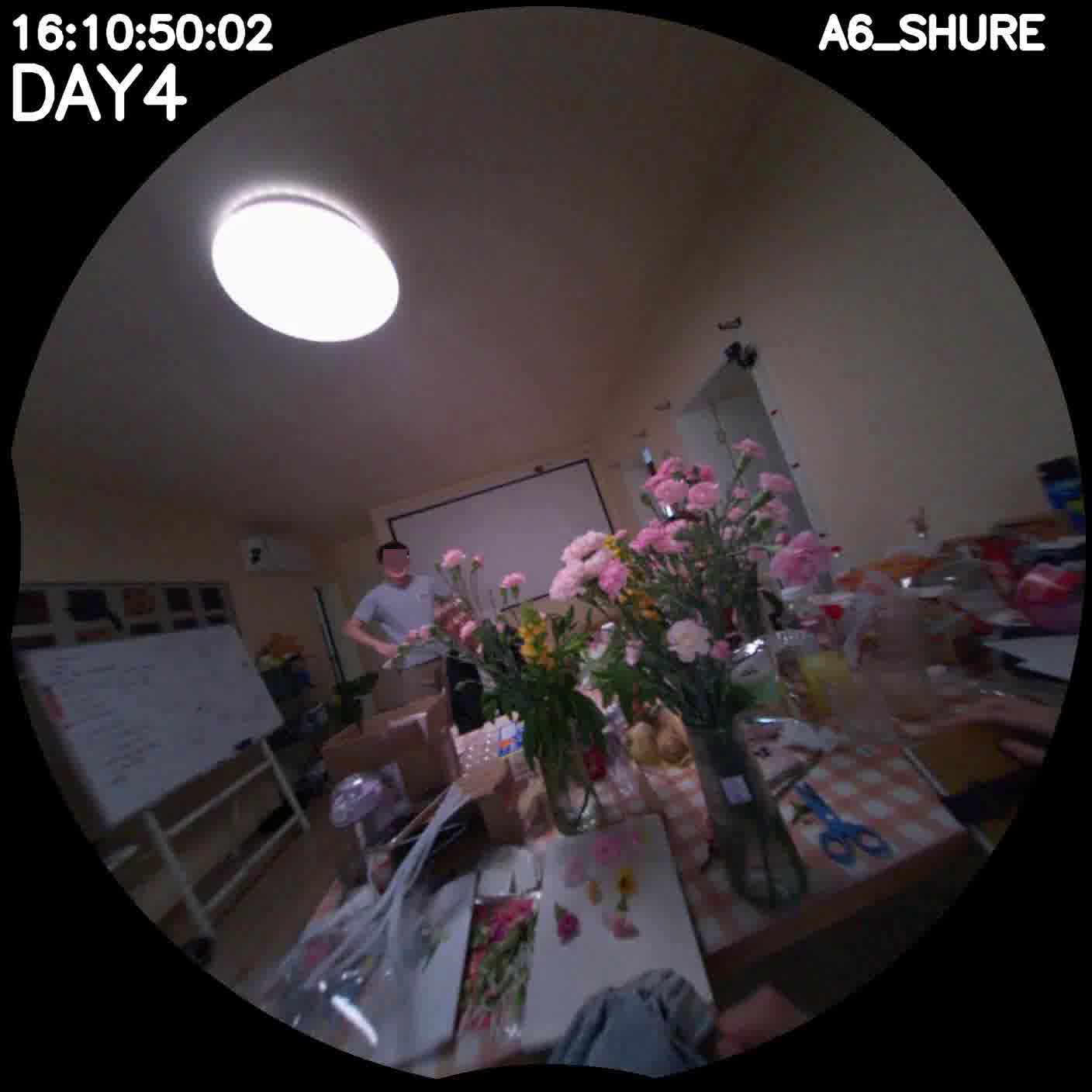} \hspace{1pt}
        \includegraphics[width=0.18\linewidth]{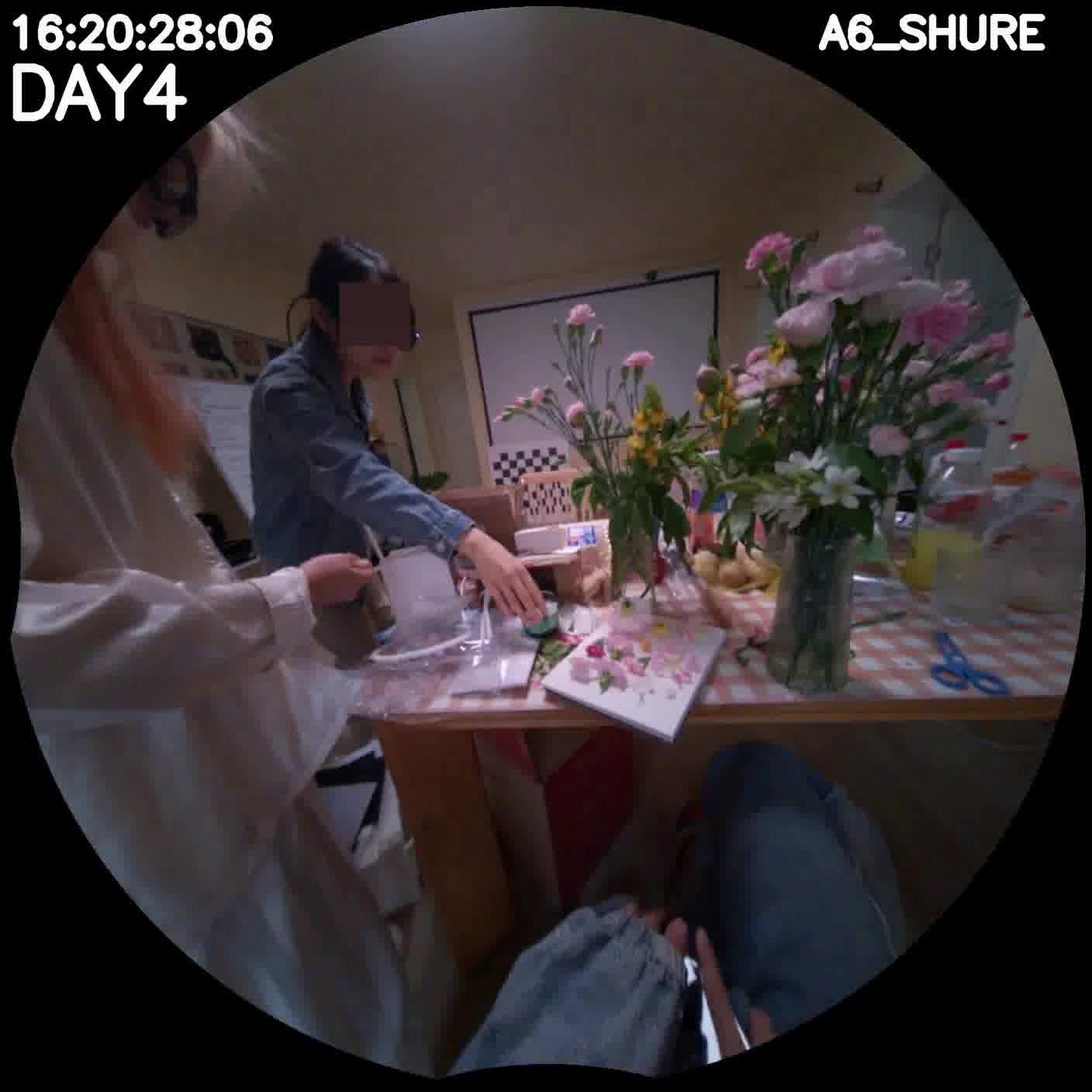}
        }
        \\
        \midrule
        \textbf{Question} & \textbf{In the delivery task, who coordinated unpacking and organizing, and who later managed packing in a different event?} \\
        ~ & (A) Alice coordinated unpacking and organizing, and later Lucia managed packing the delivery. \\
       ~ & \textbf{(B) Lucia and Katrina coordinated unpacking and organizing, and later on Jake managed packing the delivery.} \\
       ~ & (C) Alice coordinated unpacking and organizing, and later Jake managed packing the delivery. \\
       ~ & (D) Lucia and Katrina coordinated unpacking and organizing, and later Jake managed packing the delivery. \\
       ~ & (E) Lucia and Katrina coordinated unpacking and organizing, and later Lucia managed packing the delivery." \\
       \\
        \textbf{Evidence} & 
        \makecell[l]{
        Source: Lucia (Day4 4PM), Katrina (Day4 4PM)\\
        \includegraphics[width=0.18\linewidth]{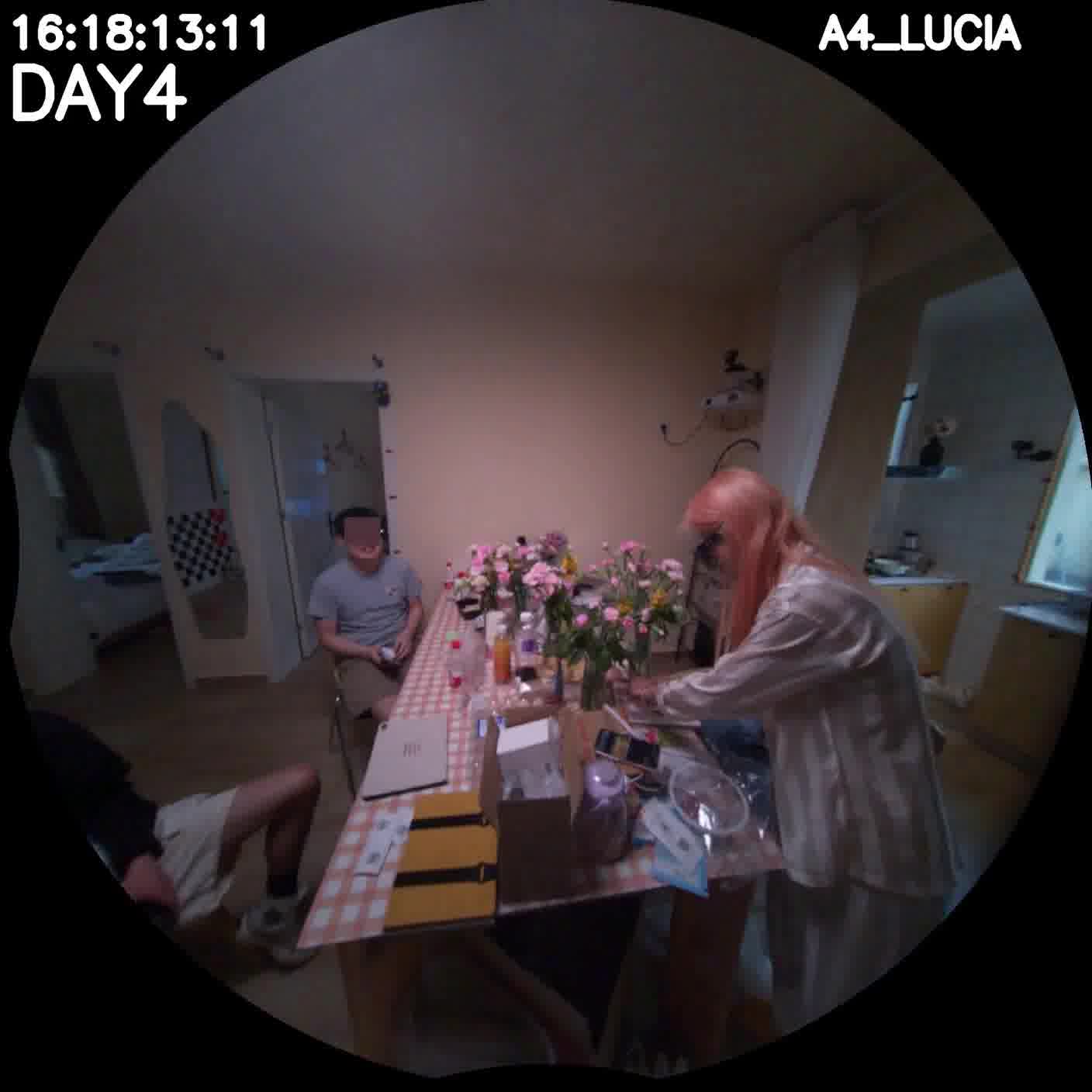} \hspace{1pt}
        \includegraphics[width=0.18\linewidth]{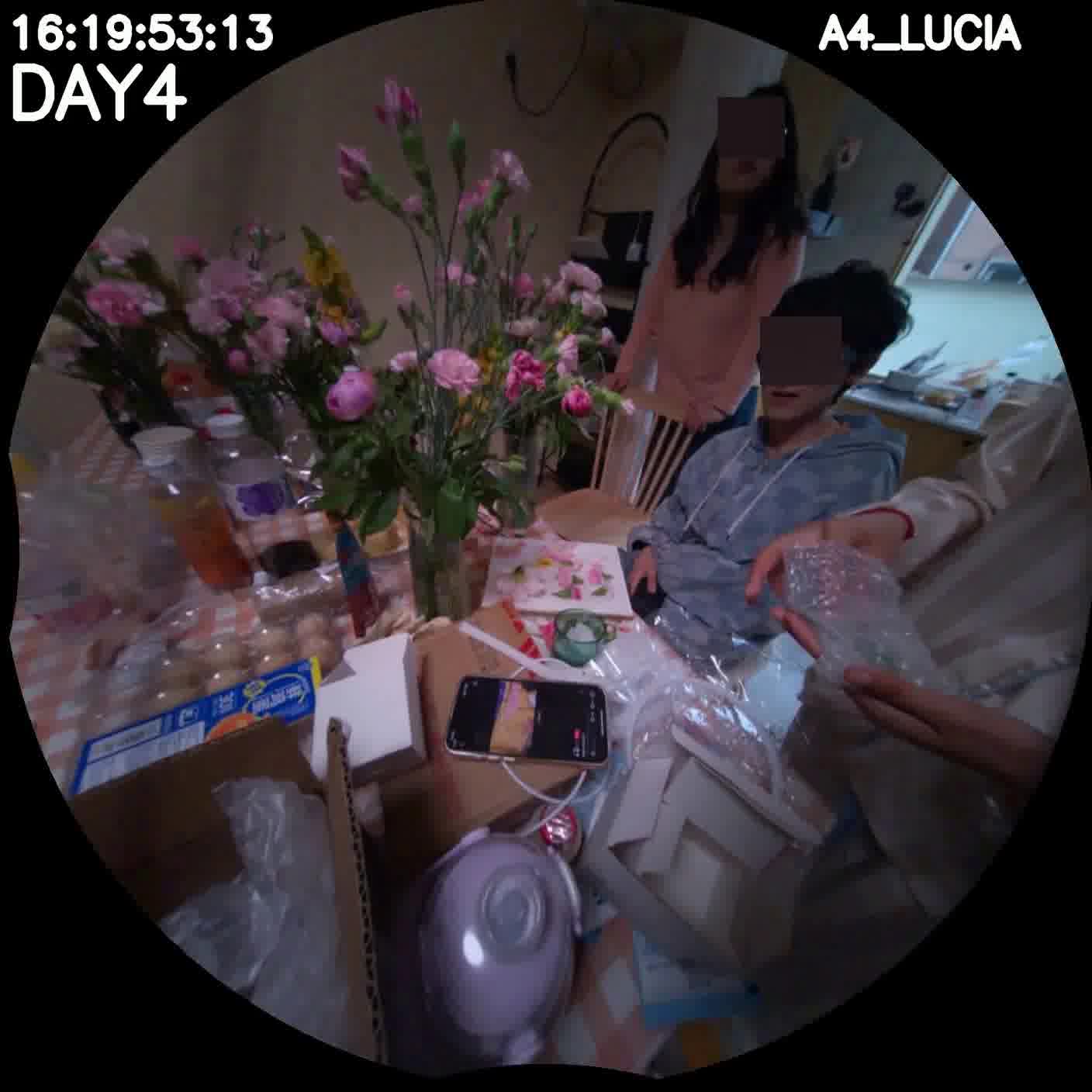} \hspace{1pt}
        \includegraphics[width=0.18\linewidth]{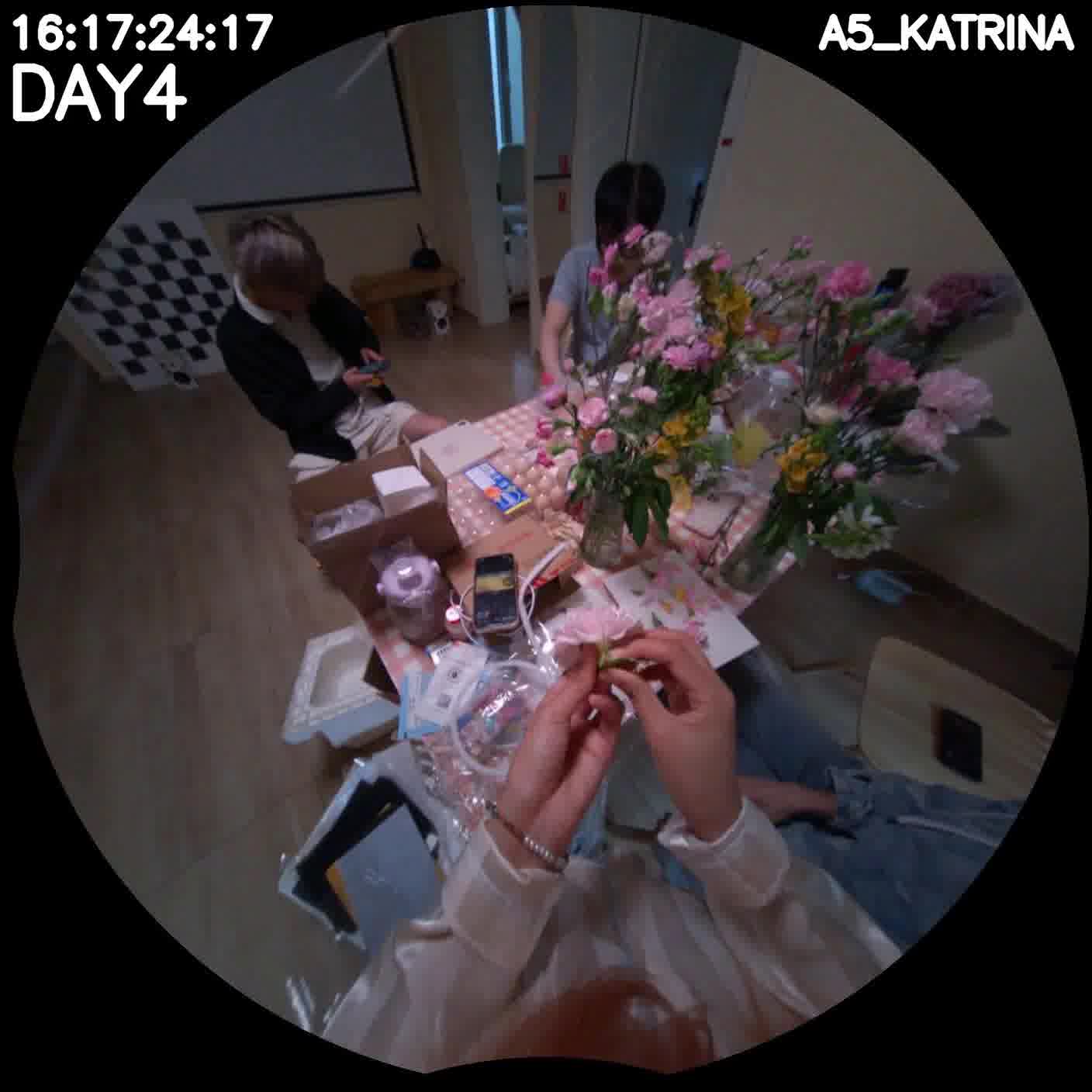} \hspace{1pt}
        \includegraphics[width=0.18\linewidth]{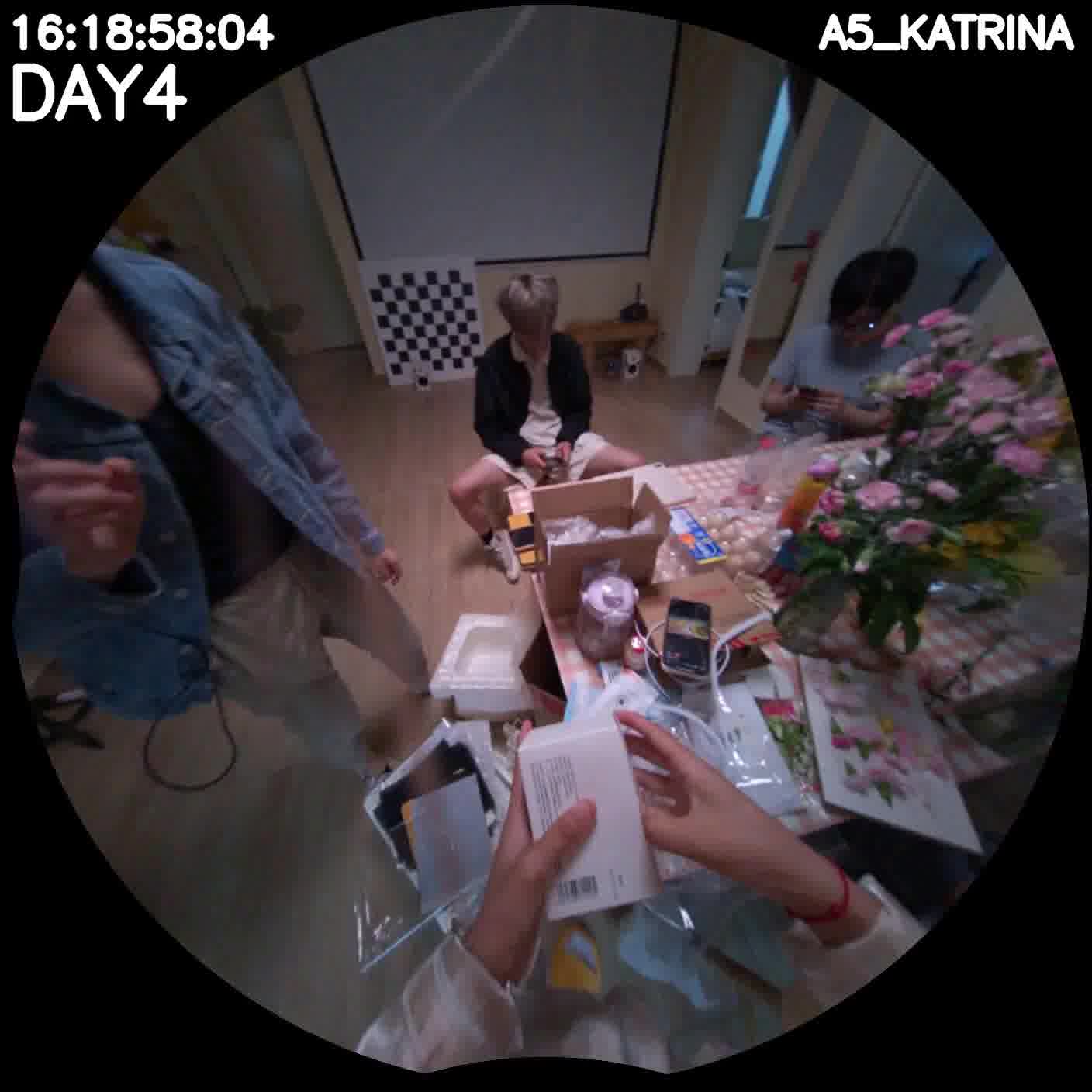} \hspace{1pt}
        \includegraphics[width=0.18\linewidth]{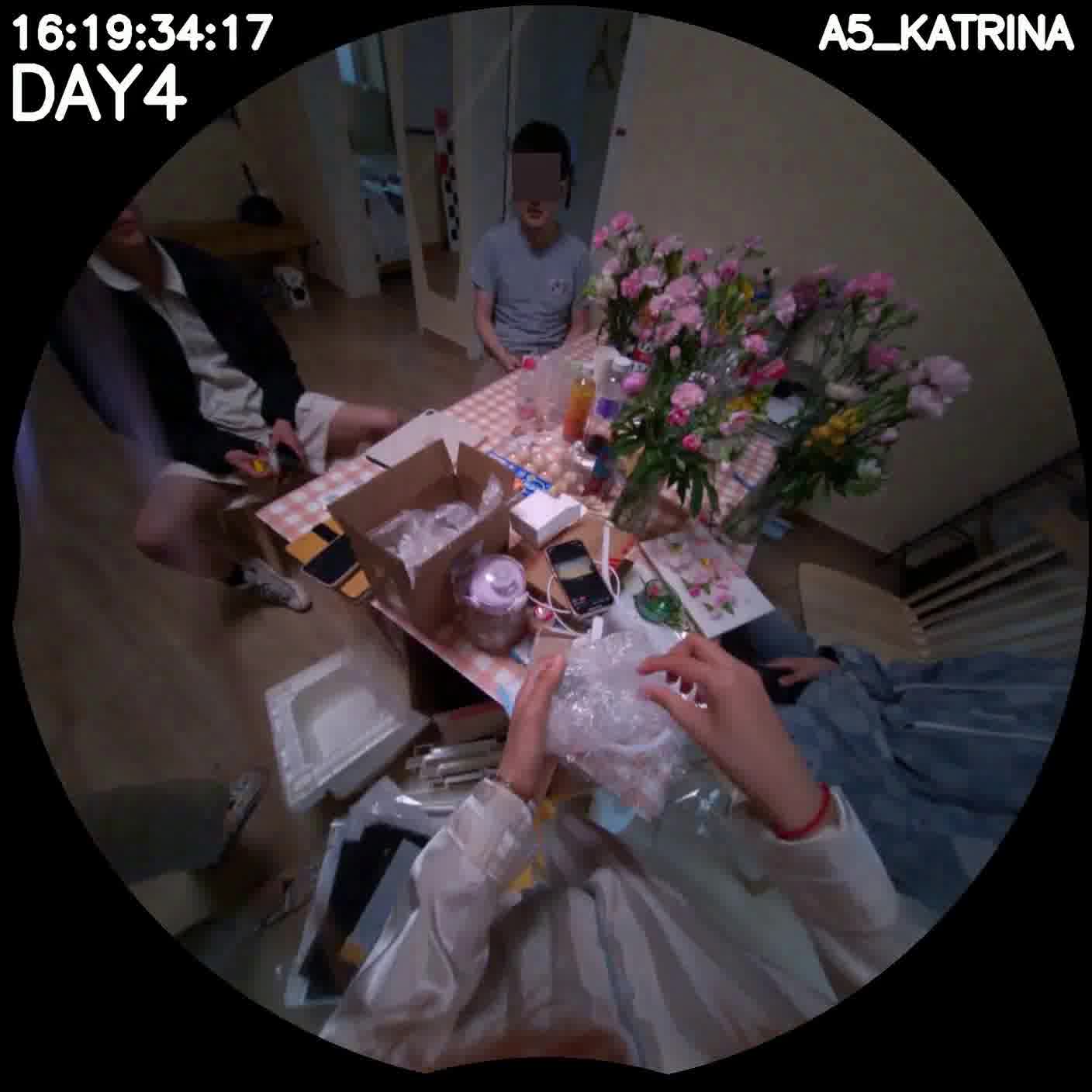}
        }
        \\
        \bottomrule
    \end{tabular}
\end{table}

\newpage
\begin{table}[h]
    \centering
    \caption{Theory of Mind (ToM) category samples}
    \label{tab:qasample_tom}
    \scriptsize
    \begin{tabular}{@{}p{0.11\linewidth} p{0.88\linewidth}@{}}
        \toprule
        \rowcolor{gray!20} \textbf{Category} & \textbf{Theory of Mind} \\
        \midrule
        \textbf{Question} & \textbf{What did Katrina wrongly assume while looking at Nicous and Violet filming?} \\
        ~ & \textbf{(A) That they were wrapping up their session immediately.} \\
        ~ & (B) That Jake wasn't involved in their filming plans. \\
        ~ & (C) That they were photographing desserts. \\
        ~ & (D) That filming had been done in the second-floor living room. \\
        ~ & (E) That Violet and Nicous would soon ask for her help. \\
        \\
        \textbf{Evidence} & 
        \makecell[l]{
        Source: Jake (Day6 10PM), Katrina (Day6 10PM) \\
        \includegraphics[width=0.18\linewidth]{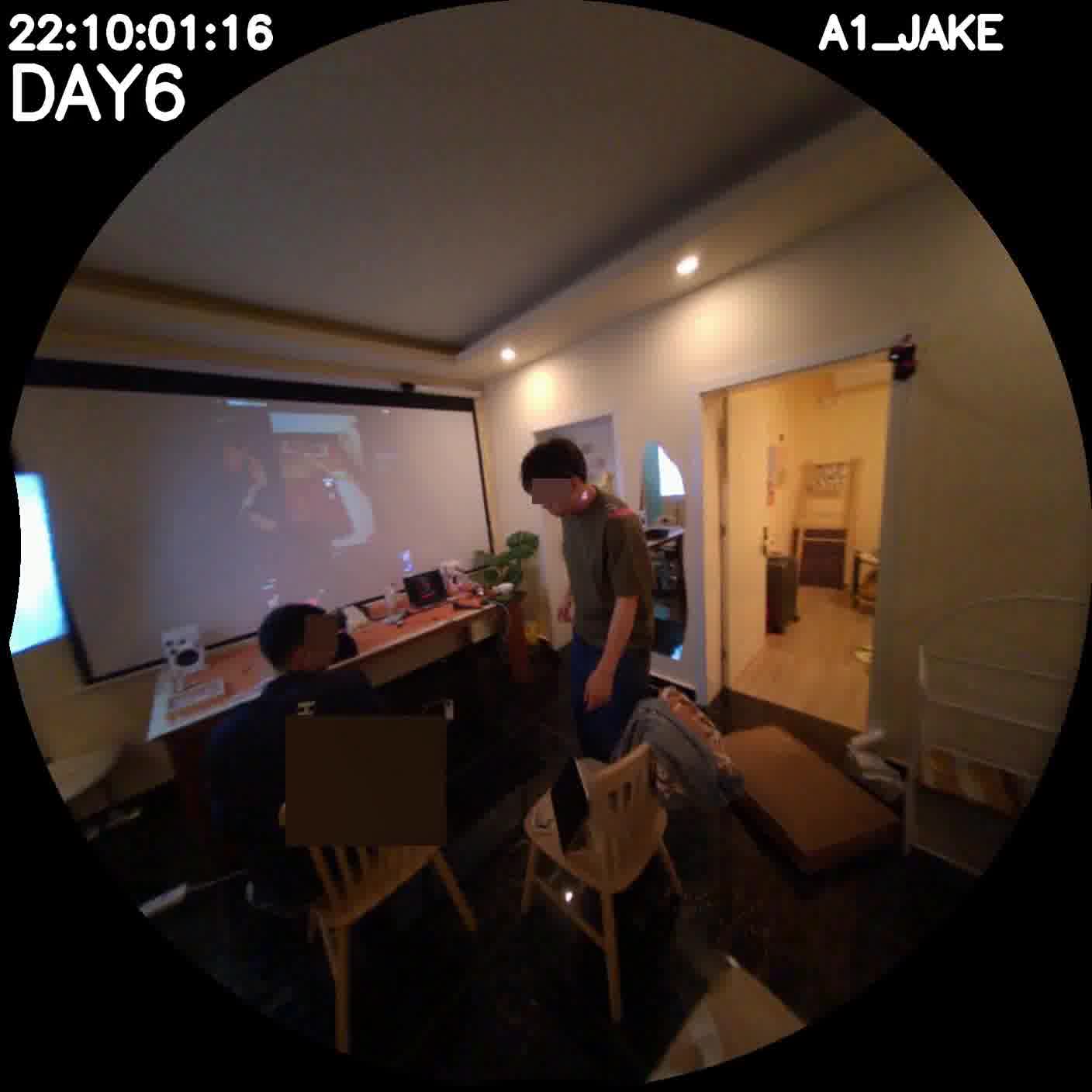} \hspace{1pt}
        \includegraphics[width=0.18\linewidth]{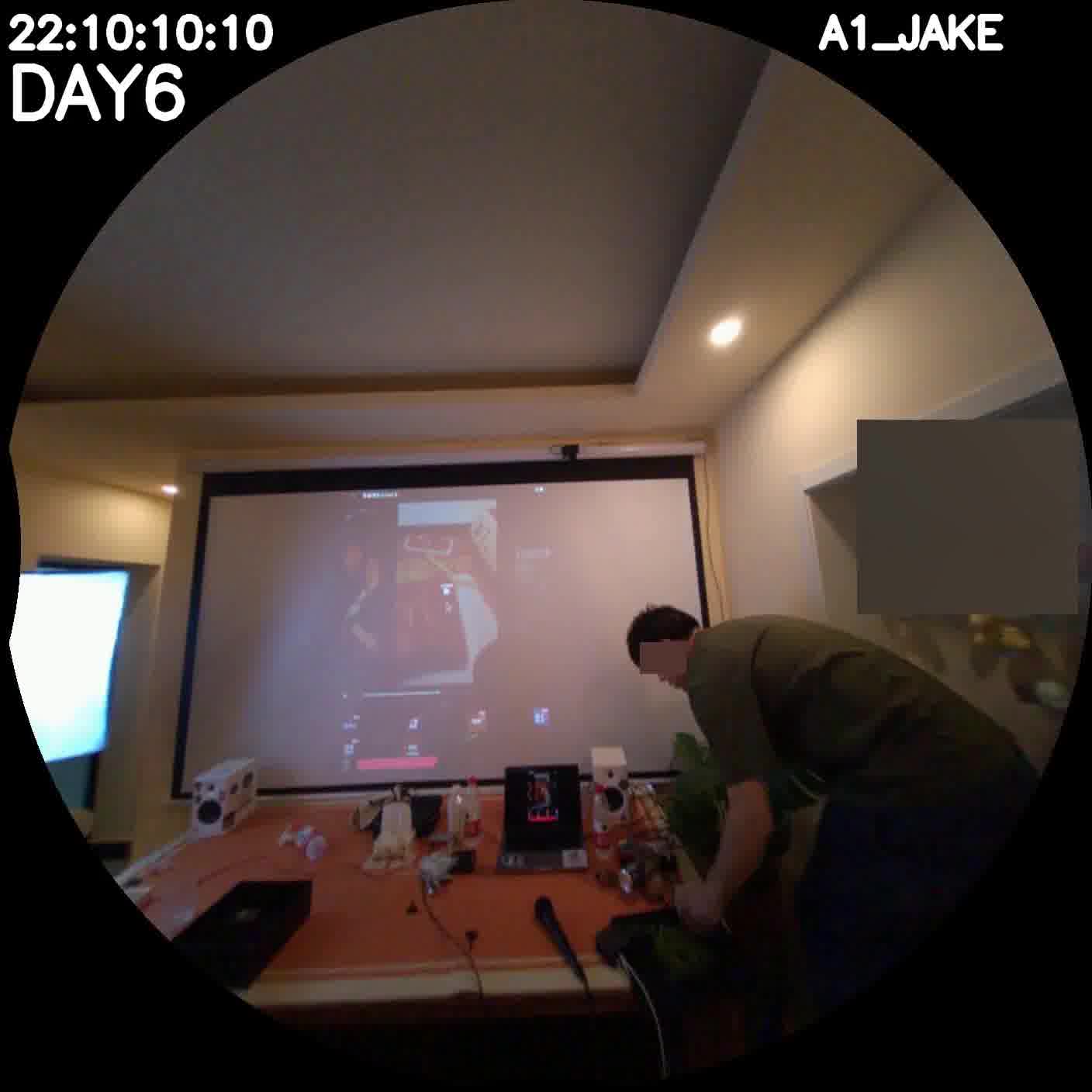} \hspace{1pt}
        \includegraphics[width=0.18\linewidth]{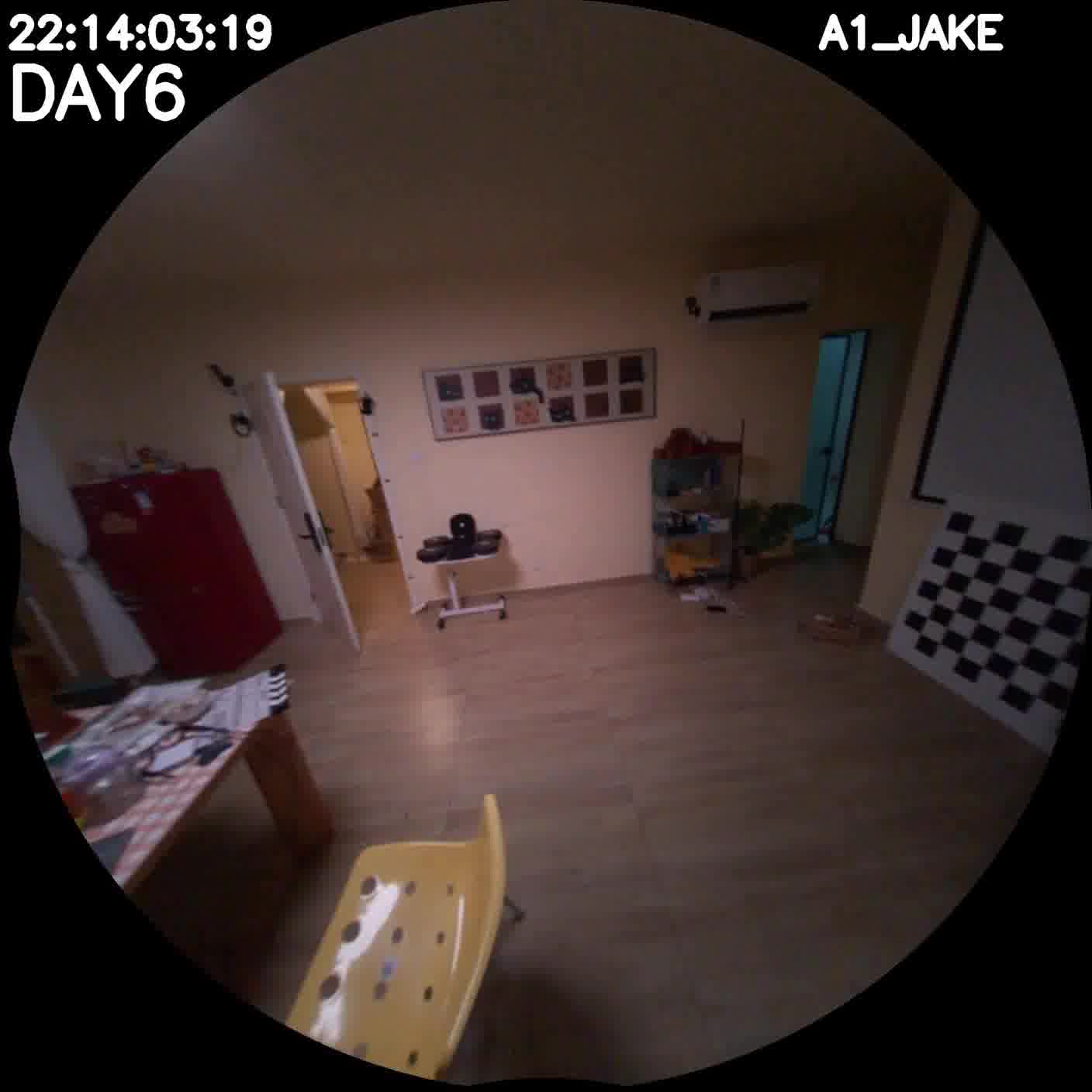} \hspace{1pt}
        \includegraphics[width=0.18\linewidth]{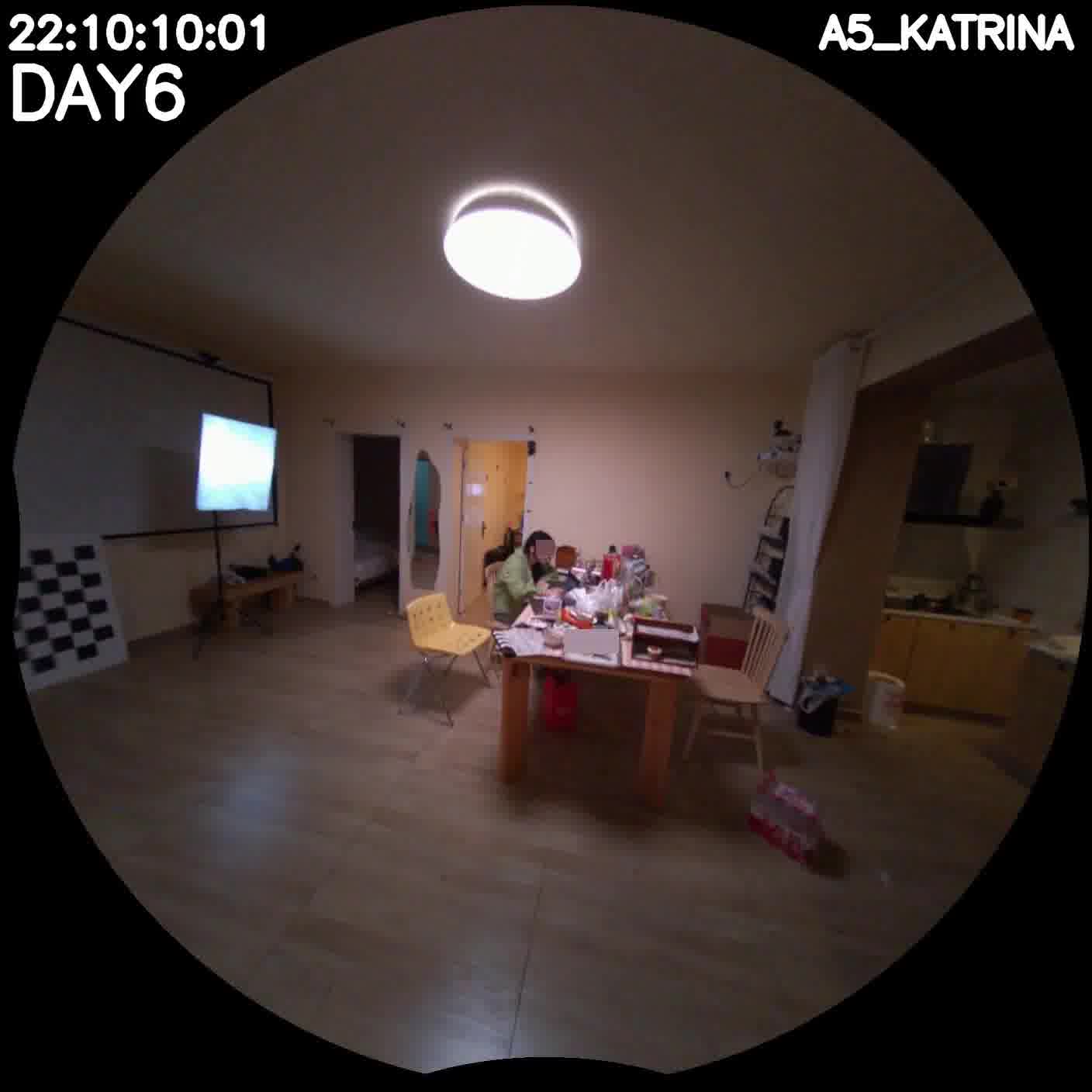} \hspace{1pt}
        \includegraphics[width=0.18\linewidth]{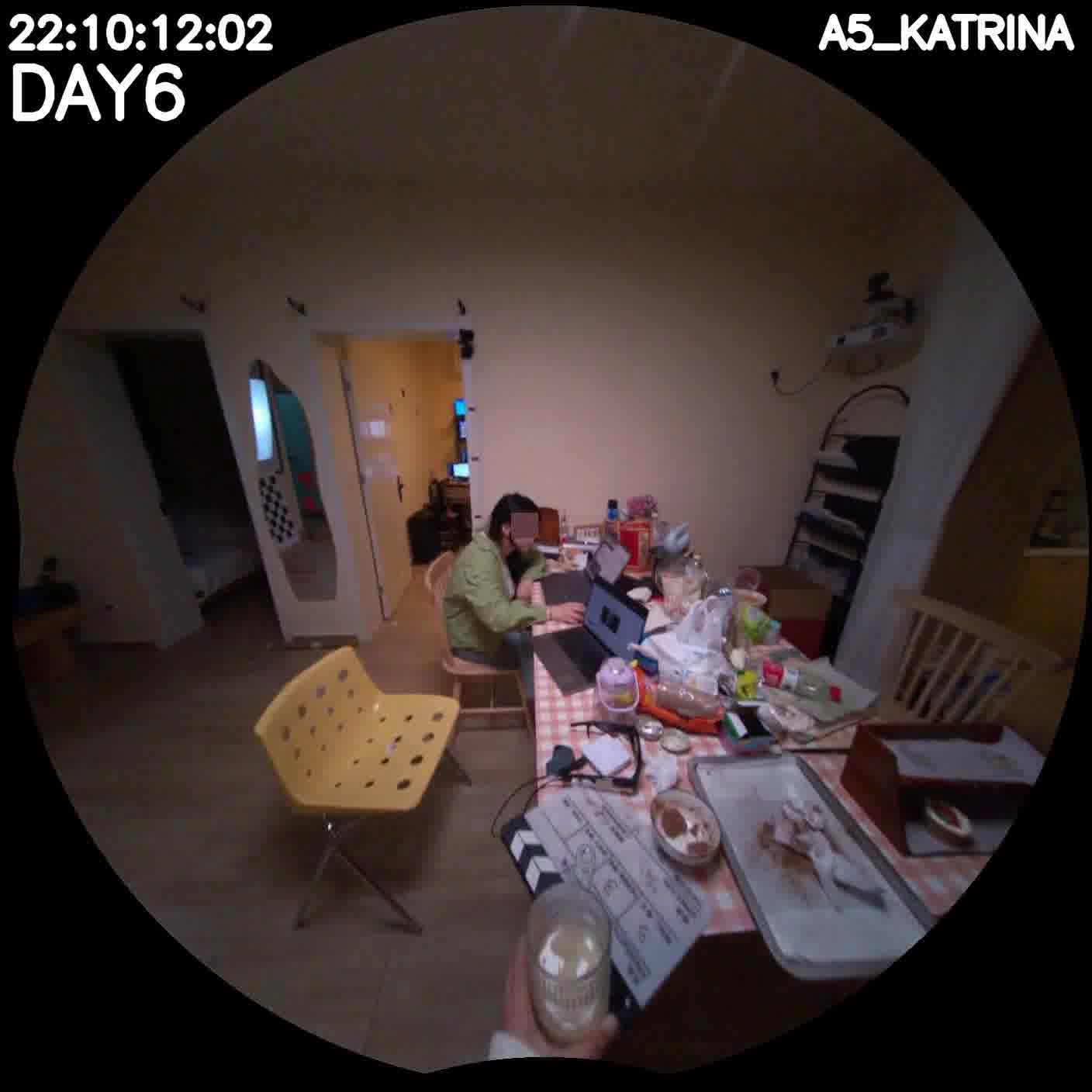} 
        } \\
        \midrule
        \textbf{Question} & \textbf{Why did Shure think Choiszt might have misinterpreted the coffee instructions?} \\
        ~ & (A) Choiszt thought they needed a latte machine instead of an espresso machine. \\
       ~ & (B) Choiszt added too much water and ignored exact measurements. \\
       ~ & \textbf{(C) Choiszt confused the steps and thought milk was required.} \\
       ~ & (D) Choiszt started brewing with a wrong type of coffee grinder. \\
       ~ & (E) Choiszt assumed fruity coffee beans required special brewing techniques. \\
        \\
        \textbf{Evidence} & 
        \makecell[l]{
        Source: Shure (Day2 3PM) \\
        \includegraphics[width=0.18\linewidth]{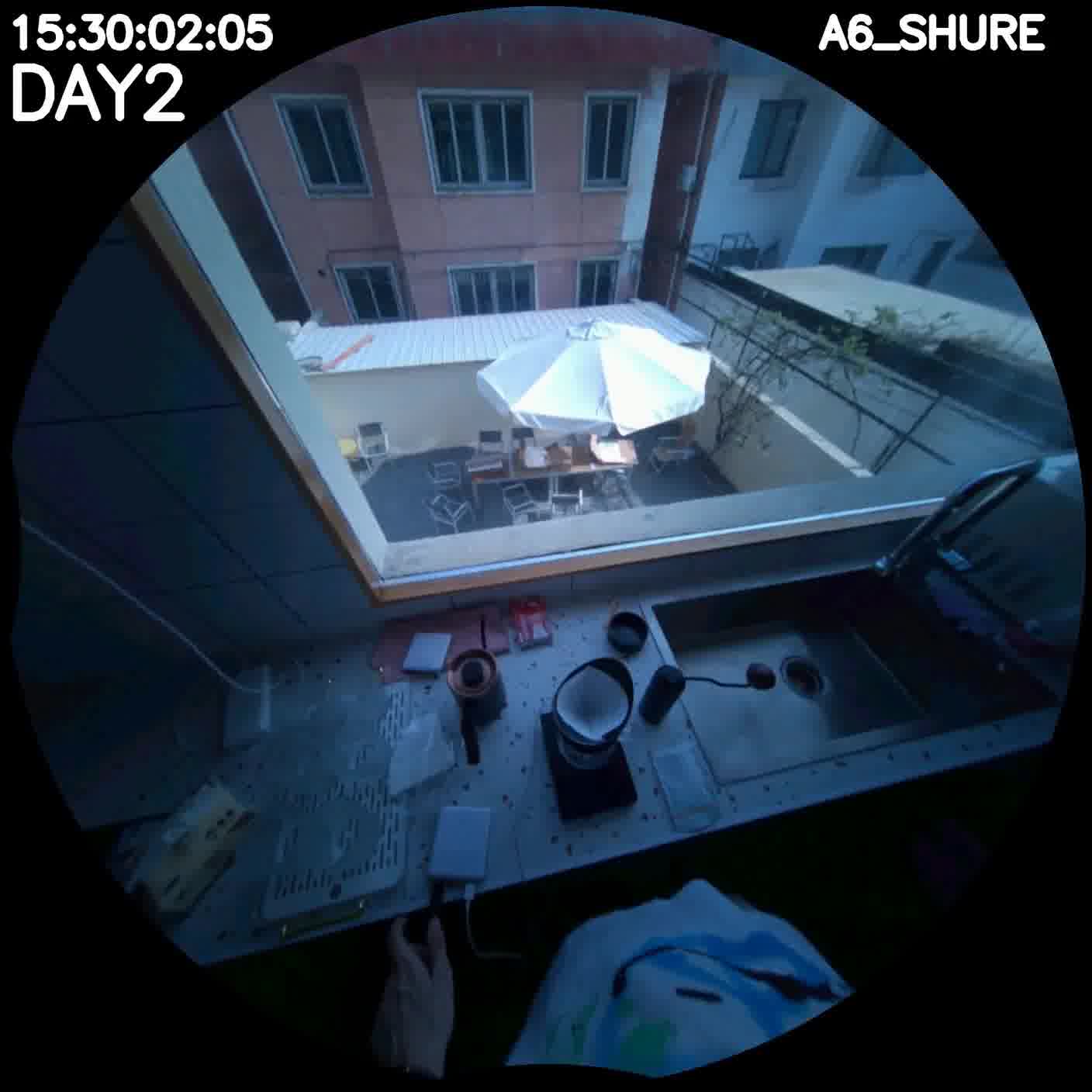} \hspace{1pt}
        \includegraphics[width=0.18\linewidth]{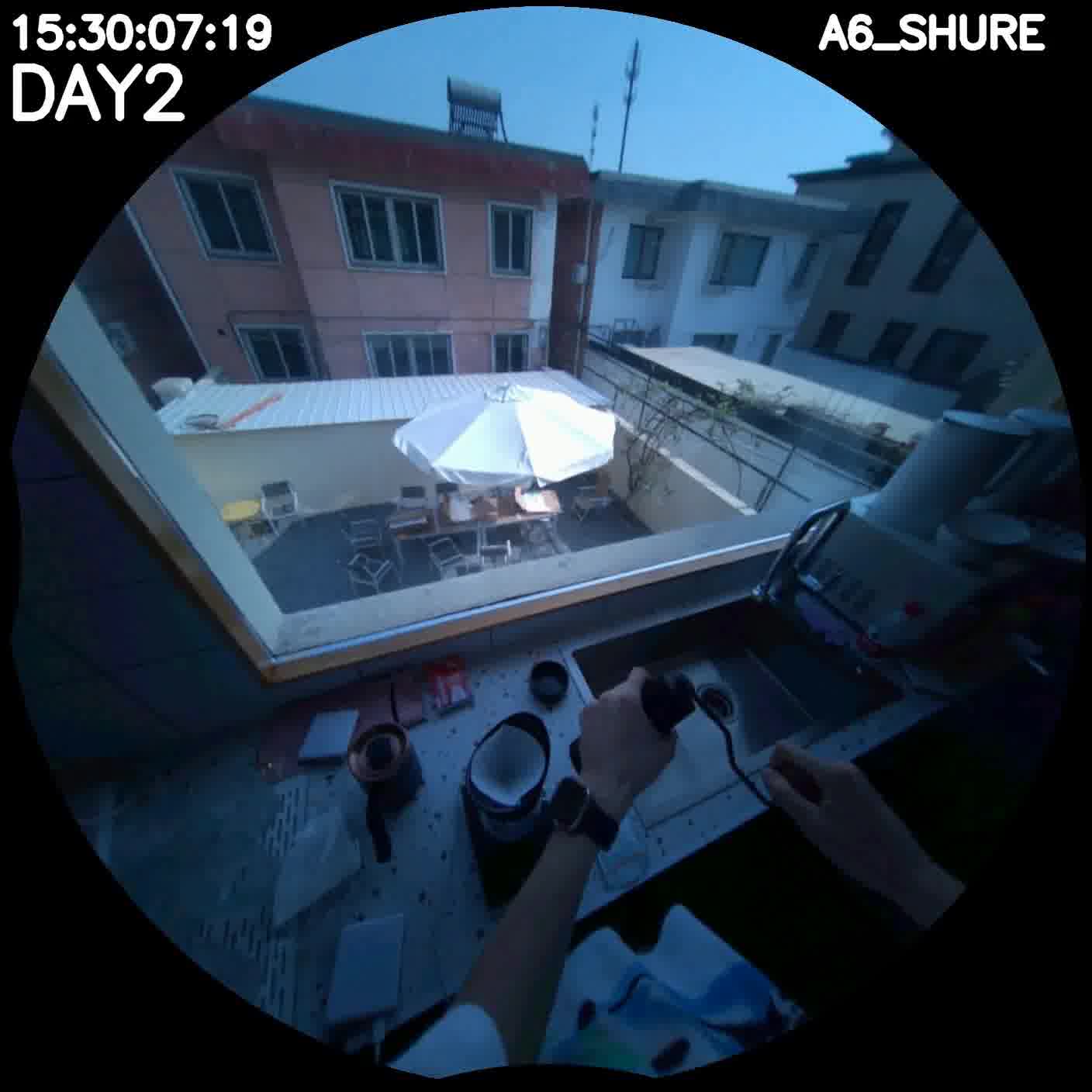}
        }
        \\
        \midrule
        \textbf{Question} & \textbf{Why didn't Katrina know where to put the big kitchen tools?} \\
        ~ & (A) Tasha was using the drawer for other purposes. \\
        ~ & (B) Katrina wasn't involved in cleaning the kitchen before.  \\
        ~ & \textbf{(C) Shure had placed things randomly, making it hard to locate spaces.} \\
        ~ & (D) The storage arrangement in the kitchen had recently changed. \\
        ~ & (E) Alice moved things while reorganizing the kitchen. \\
        \\
        \textbf{Evidence} & 
        \makecell[l]{
        Source: Katrina (Day4 11AM) \\
        \includegraphics[width=0.18\linewidth]{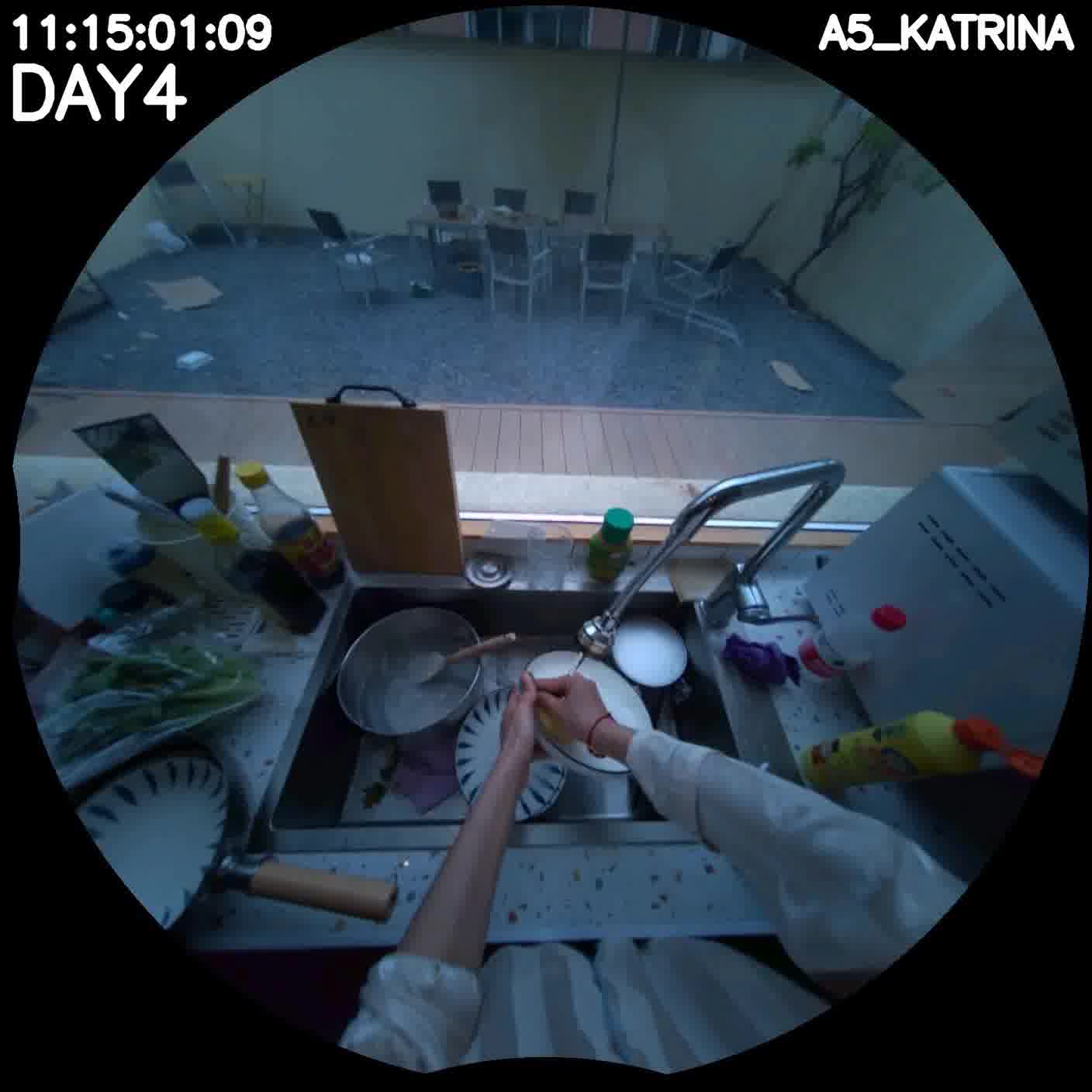} \hspace{1pt}
        \includegraphics[width=0.18\linewidth]{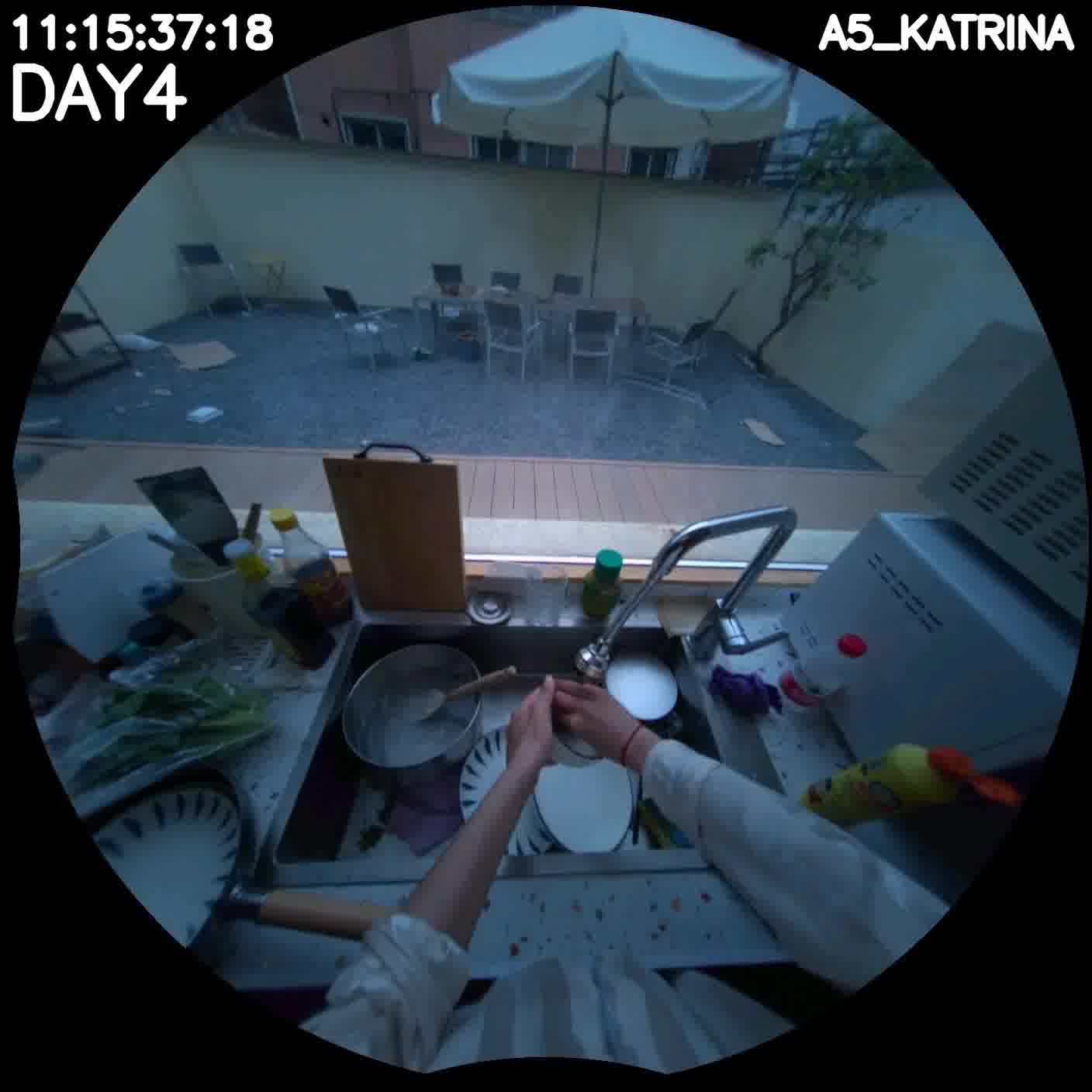} \hspace{1pt}
        \includegraphics[width=0.18\linewidth]{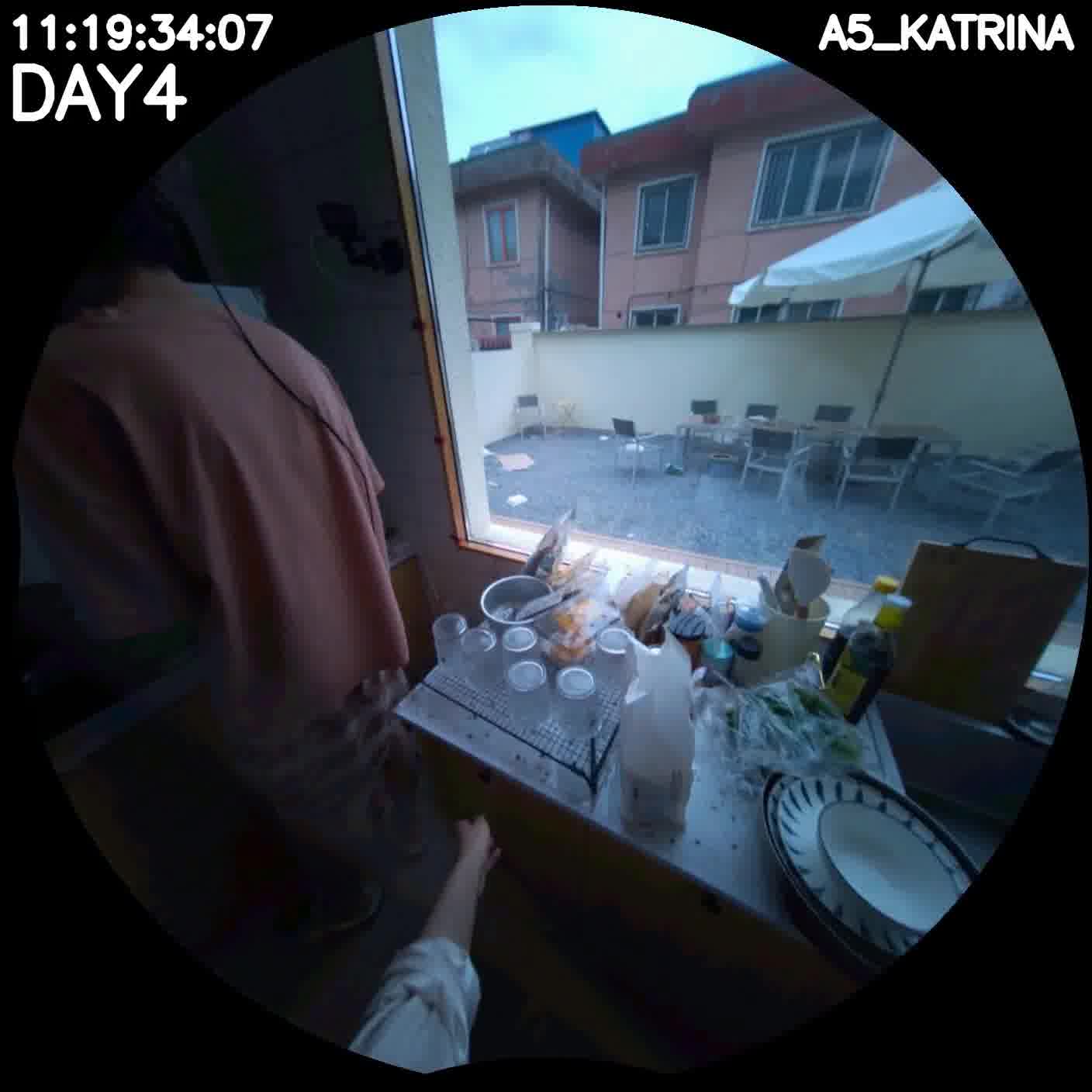} \hspace{1pt}
        \includegraphics[width=0.18\linewidth]{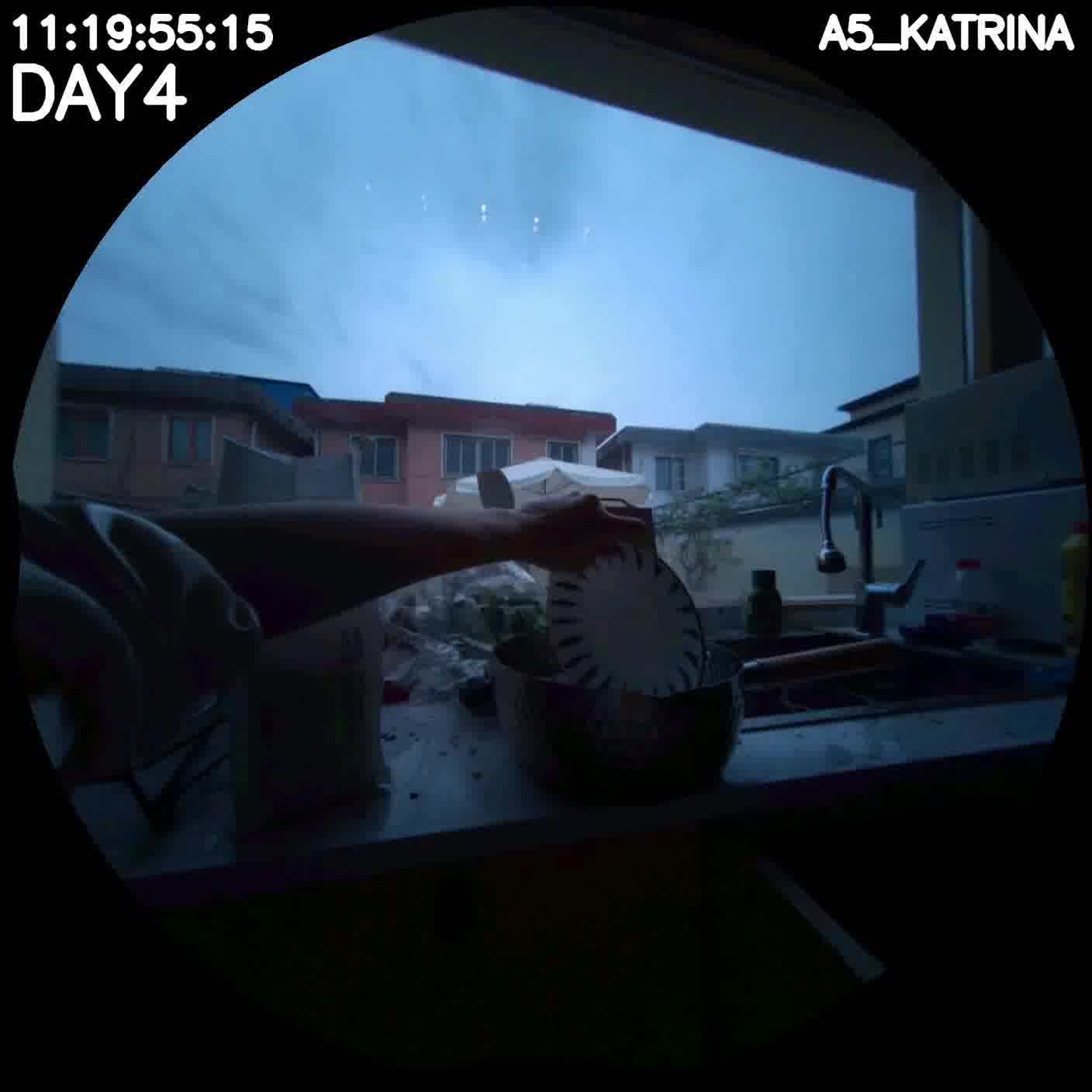} \hspace{1pt}
        }
        \\
        \bottomrule
    \end{tabular}
\end{table}

\begin{table}[h]
    \centering
    \caption{Temporal Reasoning (TR) - Comparison category samples}
    \label{tab:qasample_tr_com}
    \scriptsize
    \begin{tabular}{@{}p{0.11\linewidth} p{0.88\linewidth}@{}}
        \toprule
        \rowcolor{gray!20} \textbf{Category} & \textbf{Temporal Reasoning, Comparison} \\
        \midrule
        \textbf{Question} & \textbf{What happened between when Lucia led the puzzle completion and supply planning and when Alice deep-cleaned the kitchen after hot pot?} \\
        ~ &  (A) Shure repaired a drone at dawn before anyone arrived. \\
       ~ & (B) Tasha oversaw final errands and device data consolidation discussions. \\
       ~ & (C) Lucia ran a midnight karaoke session after the meeting. \\
       ~ & \textbf{(D) Jake coordinated outdoor setup and takeout distribution in the courtyard.} \\
       ~ & (E) Katrina led themed room planning and decor decisions. \\
        \\
        \textbf{Evidence} & 
        \makecell[l]{
        Source: Jake (Day1 2PM), Alice (Day2 8PM), Lucia (Day1 1PM) \\
        \includegraphics[width=0.18\linewidth]{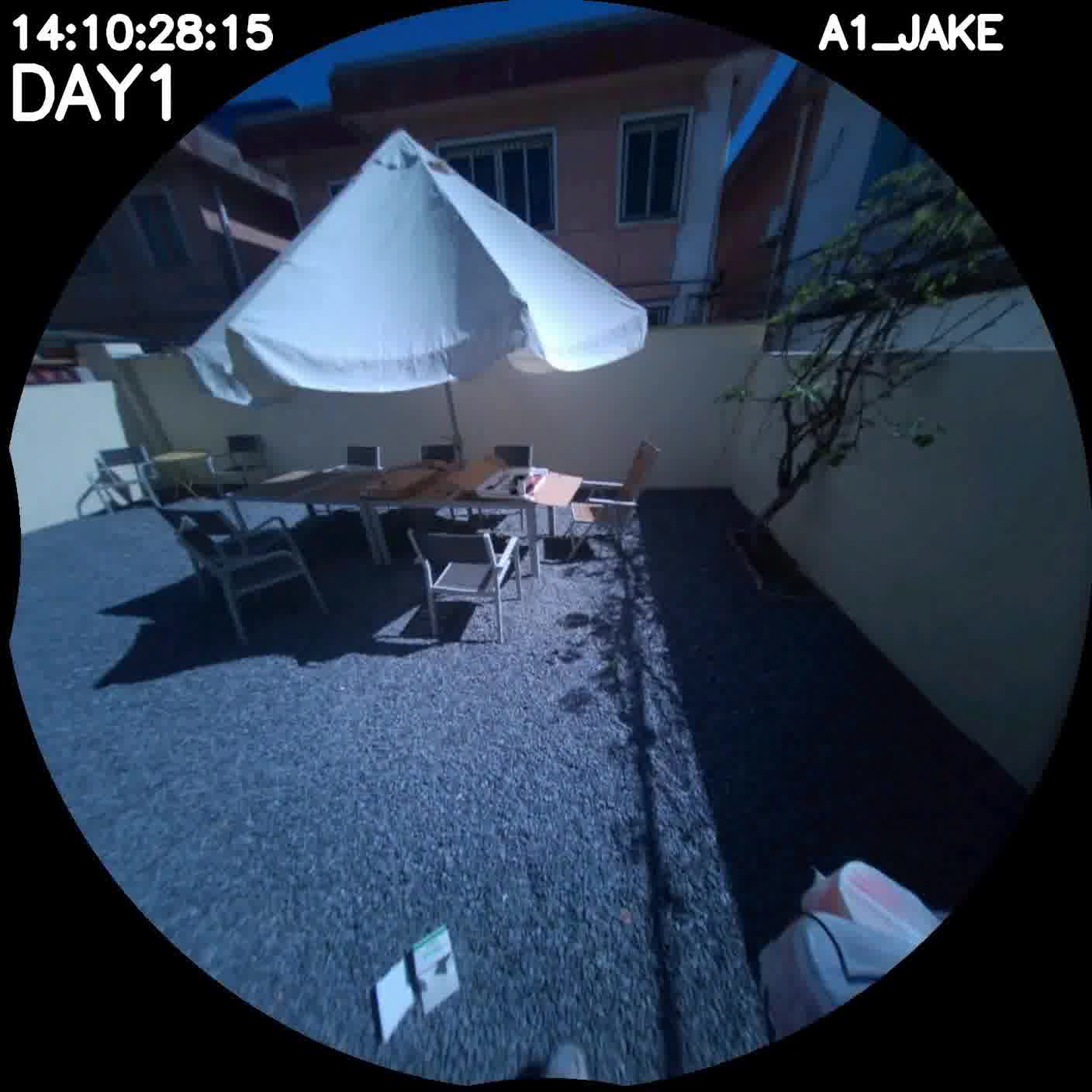} \hspace{1pt}
        \includegraphics[width=0.18\linewidth]{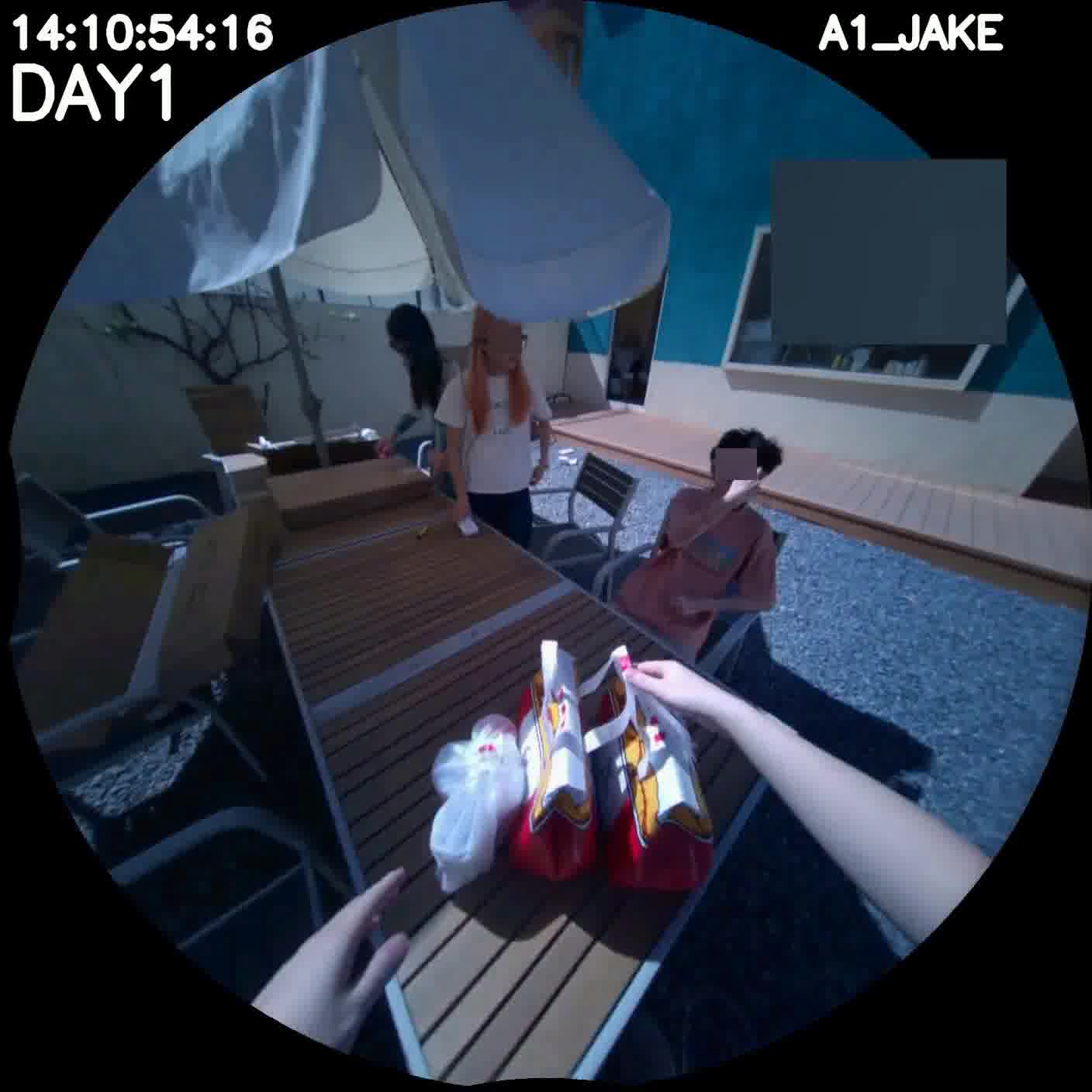} \hspace{1pt}
        \includegraphics[width=0.18\linewidth]{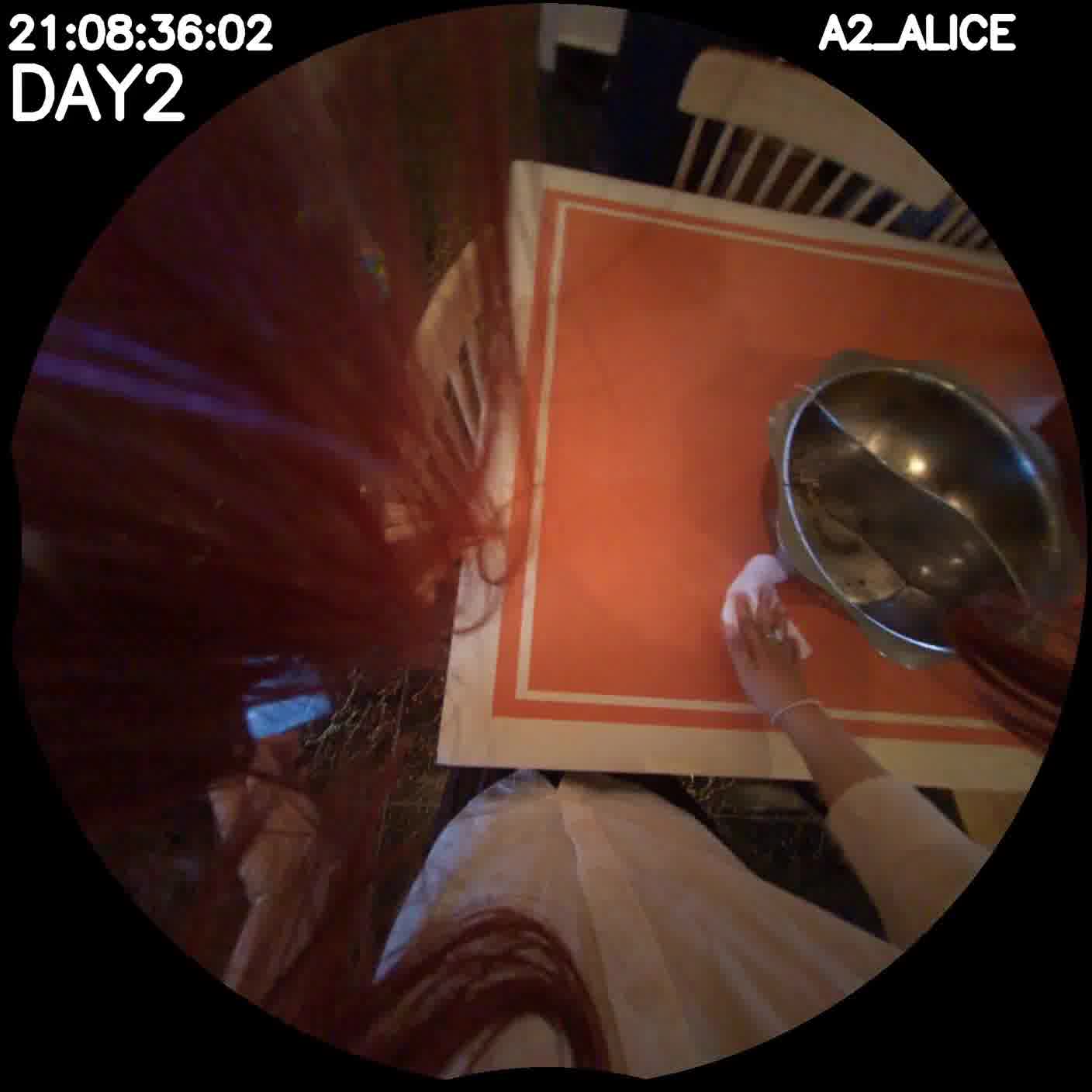} \hspace{1pt}
        \includegraphics[width=0.18\linewidth]{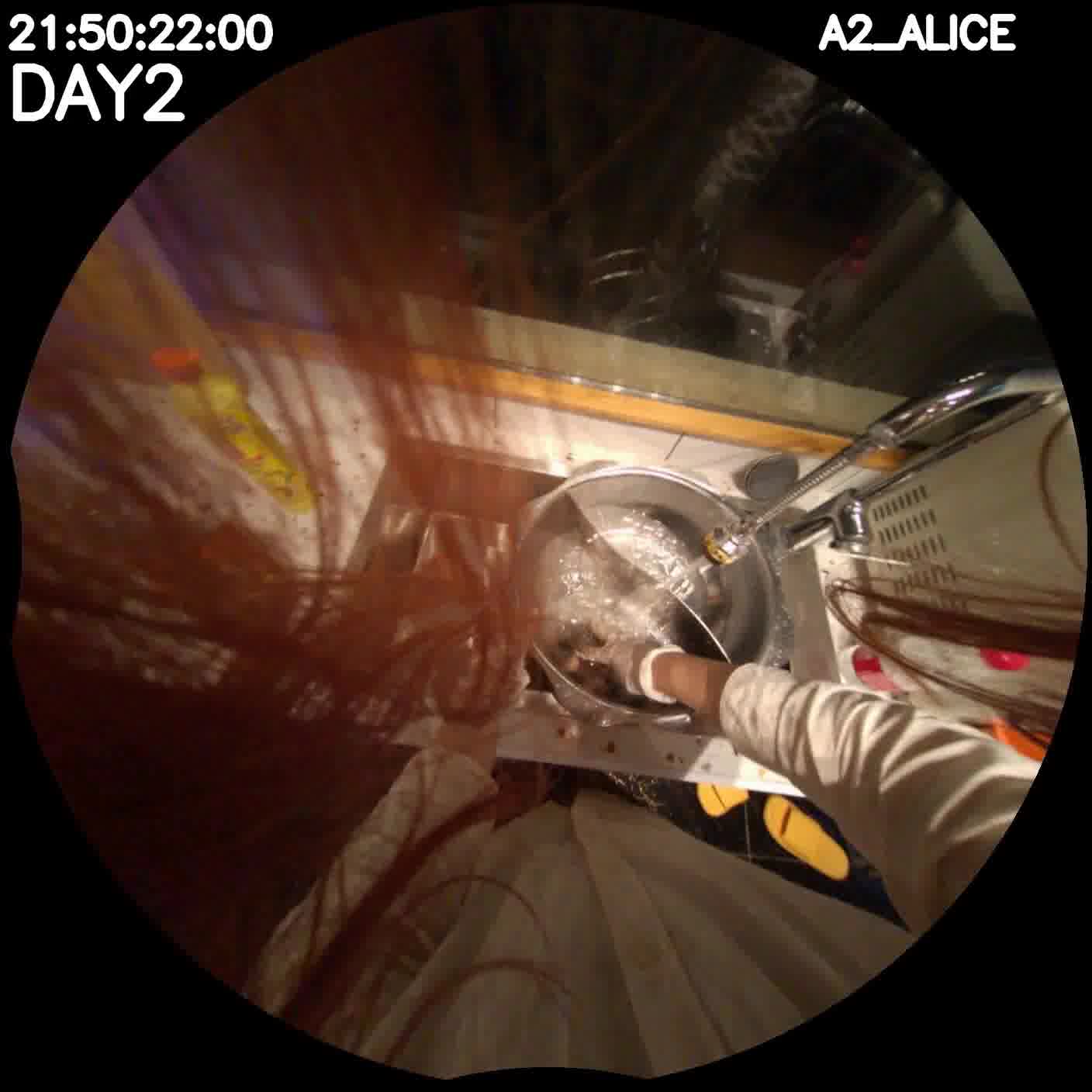} \hspace{1pt}
        \includegraphics[width=0.18\linewidth]{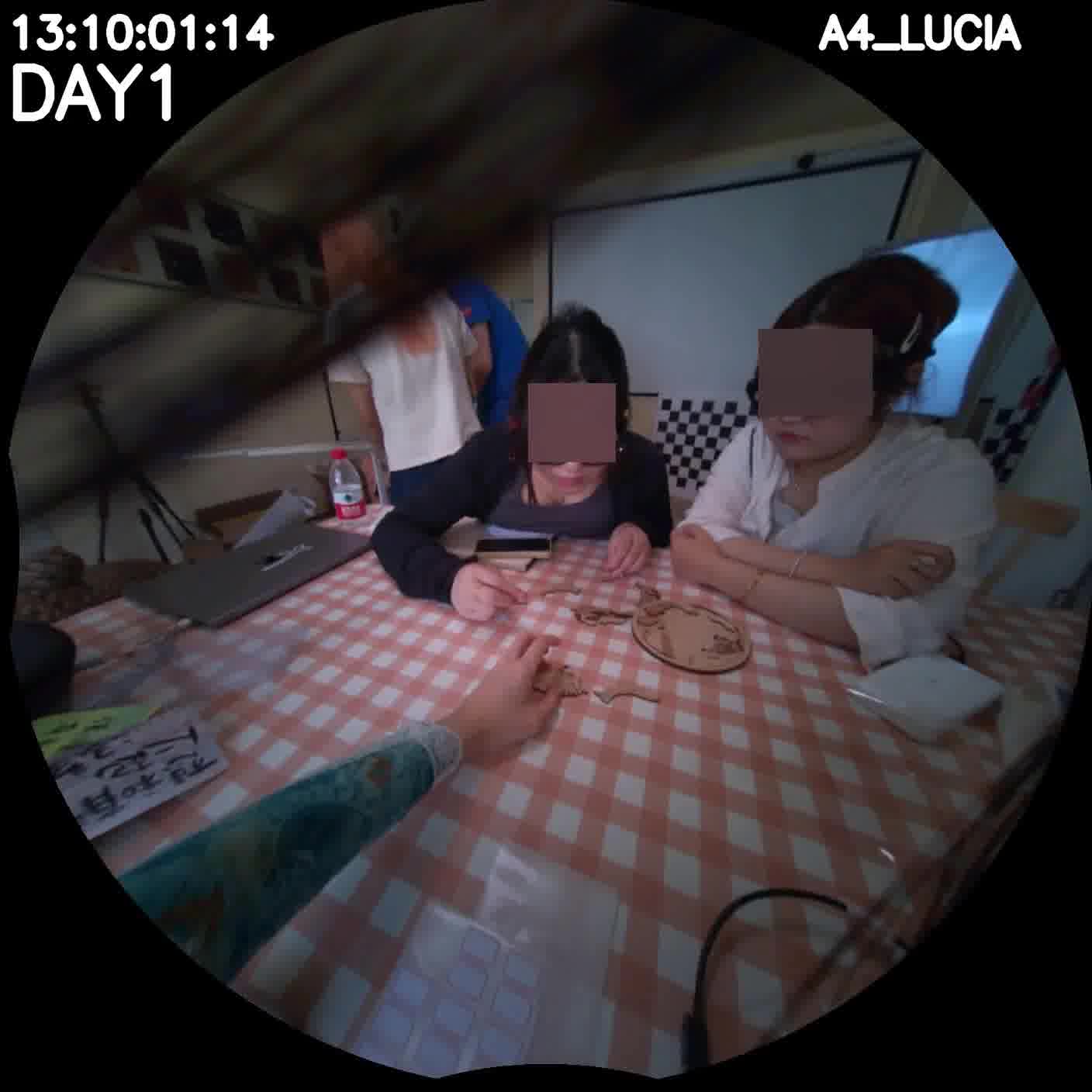}
        }
        \\        
        \midrule
        \textbf{Question} & \textbf{Which is the correct sequence of events?} \newline A) Alice retrieved a power bank, doubled back for a forgotten item, and then took a ride-hailing car. \newline B) Tasha wrapped up checking tickets and led a group discussion on May Day travel, even demonstrating a bus ticket booking. \newline C) Jake set up lighting and equipment for meal prep and then worked on computer file transfers. \\
        ~ & (A) A-C-B \\
       ~ & (B) B-C-A \\
       ~ & (C) A-B-C \\
       ~ & \textbf{(D) C-A-B} \\
       ~ & (E) B-A-C \\
        \\
        \textbf{Evidence} & 
        \makecell[l]{
        Source: Jake (Day2 6PM), Alice (Day4 9PM), Tasha (Day2 11AM) \\
        \includegraphics[width=0.18\linewidth]{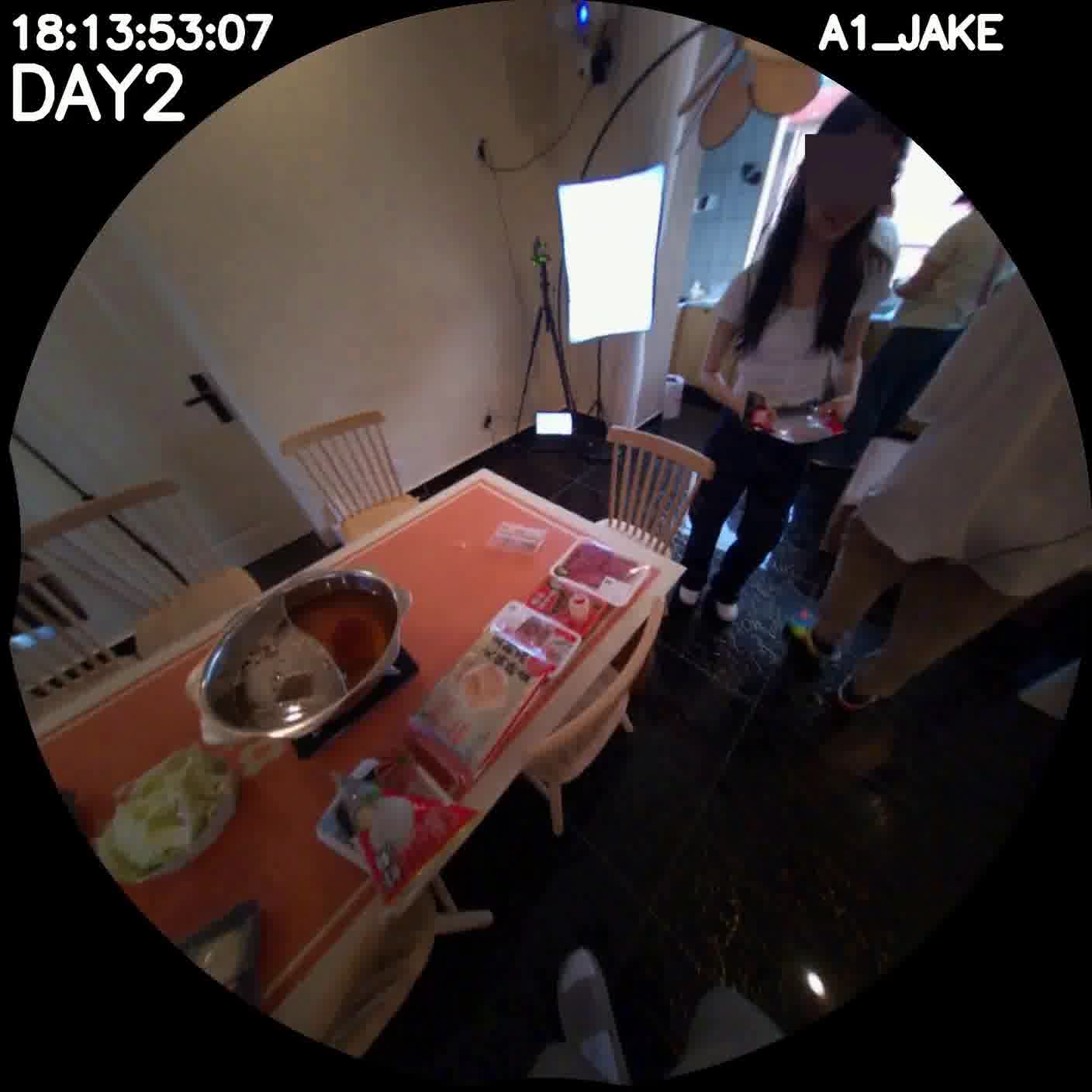} \hspace{1pt}
        \includegraphics[width=0.18\linewidth]{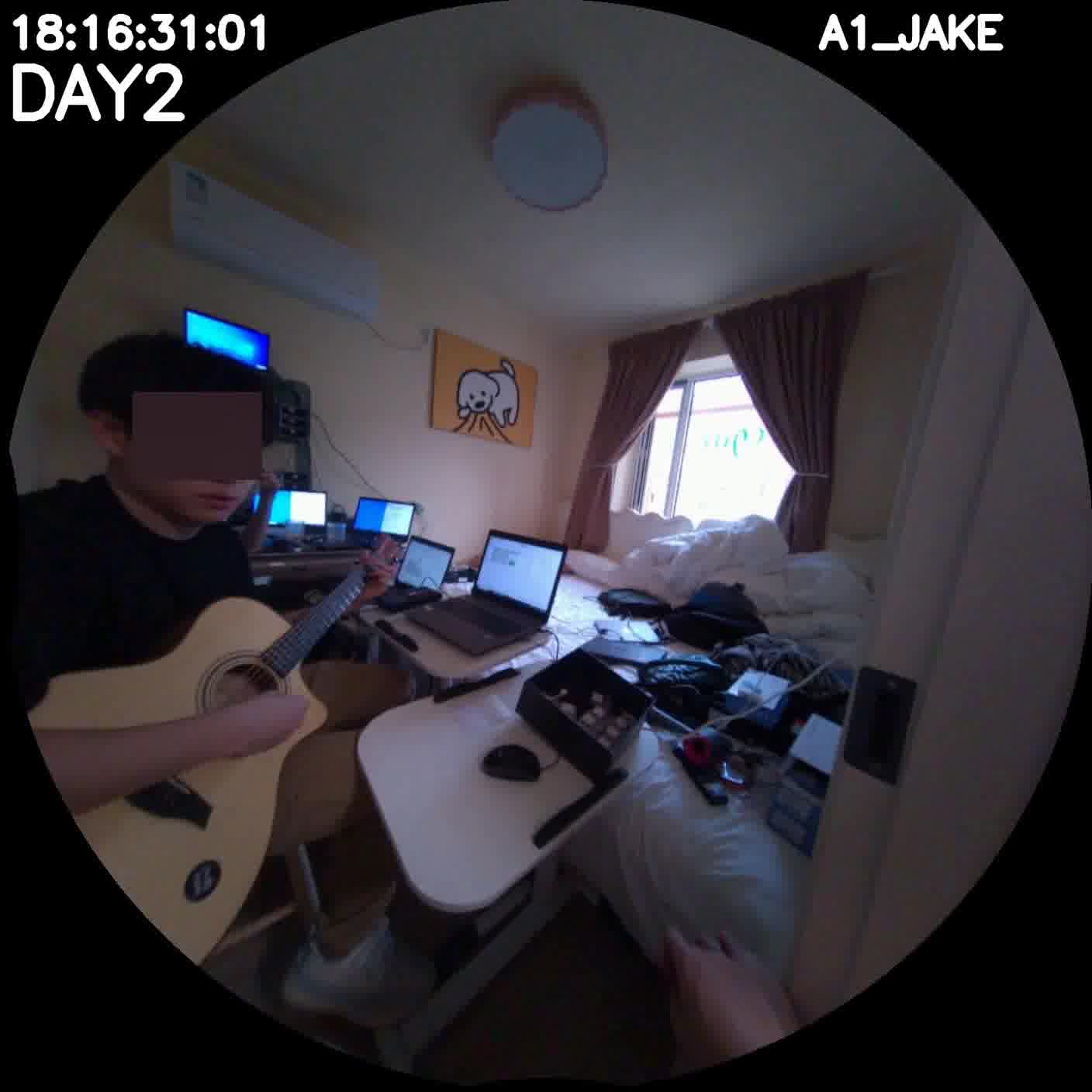} \hspace{1pt}
        \includegraphics[width=0.18\linewidth]{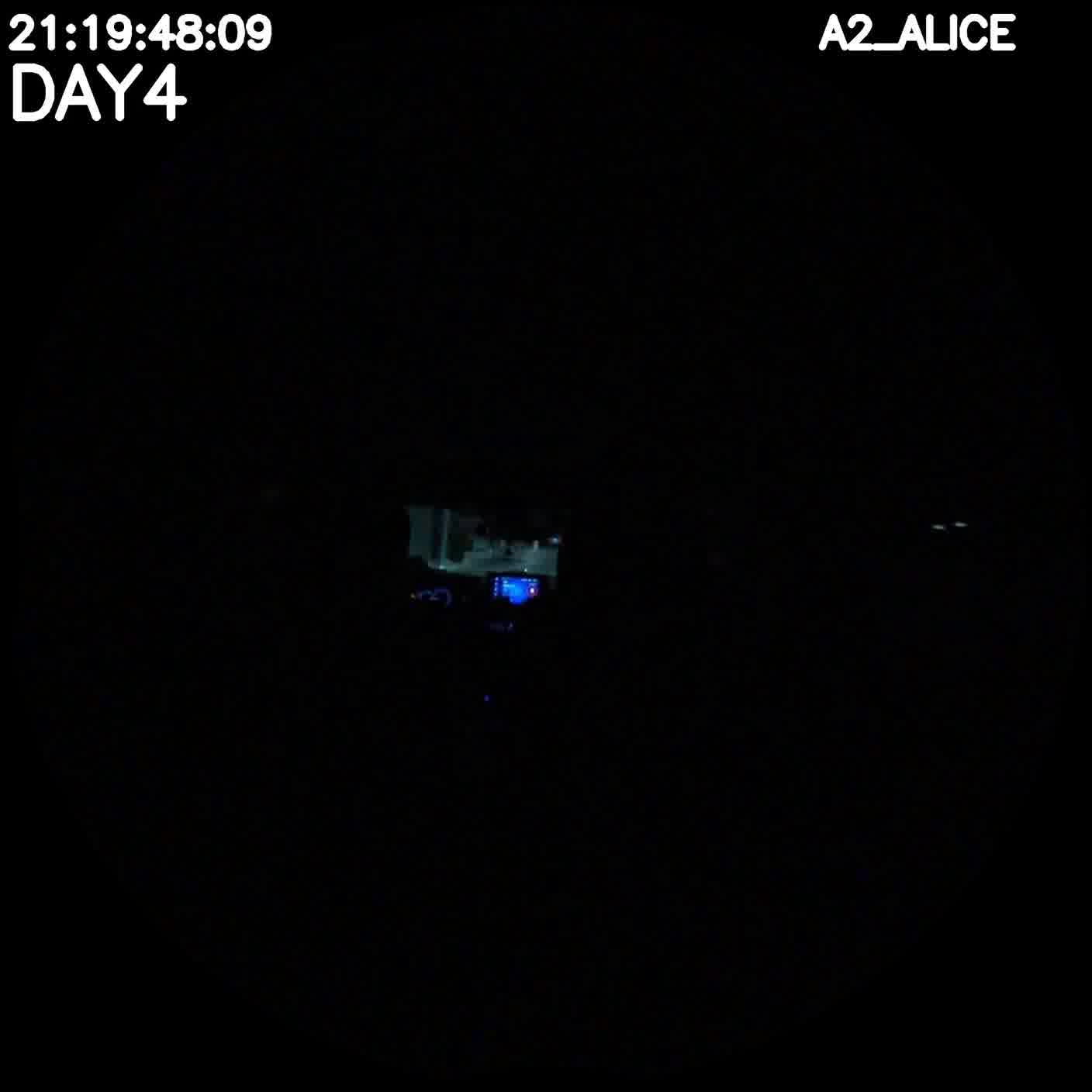} \hspace{1pt}
        \includegraphics[width=0.18\linewidth]{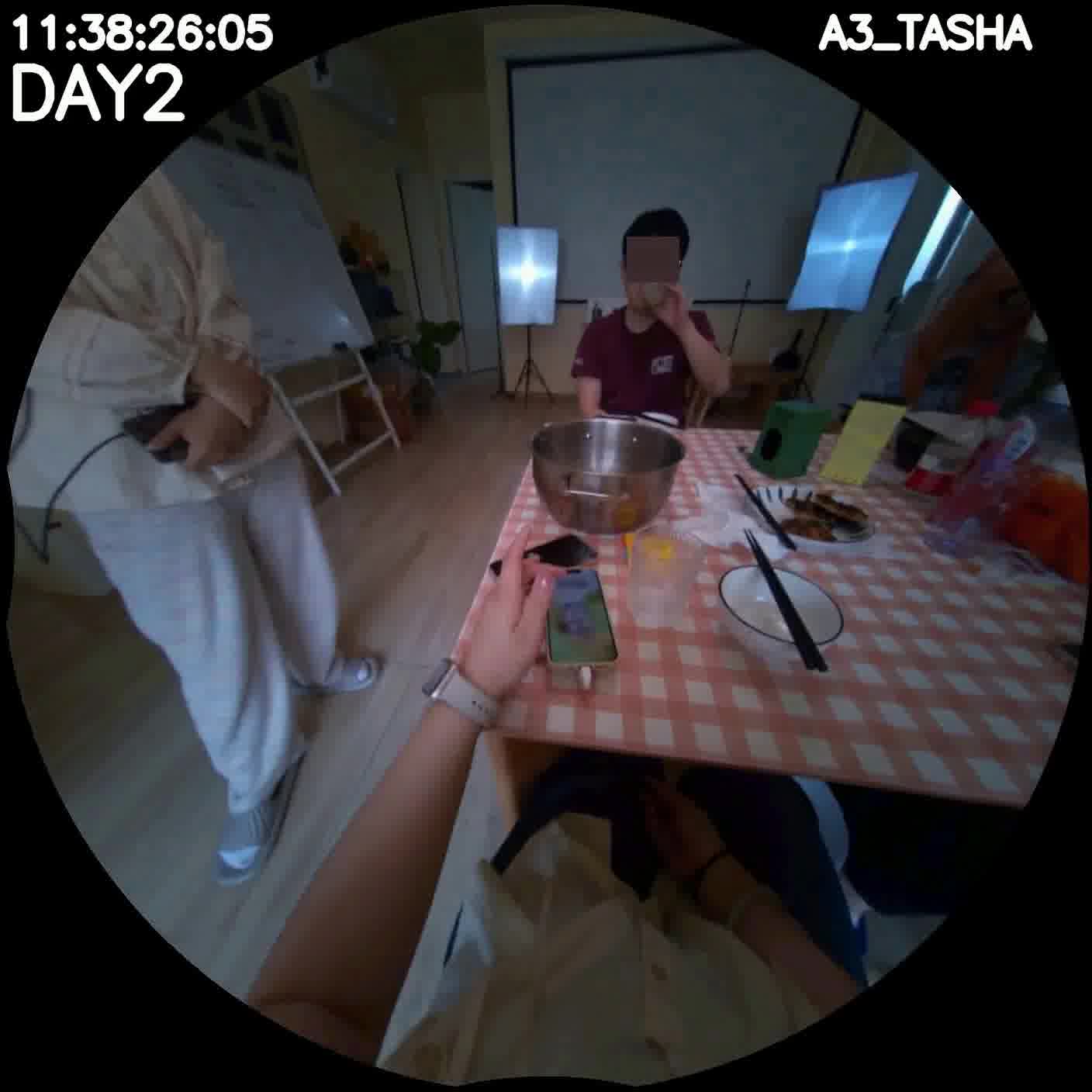} \hspace{1pt}
        \includegraphics[width=0.18\linewidth]{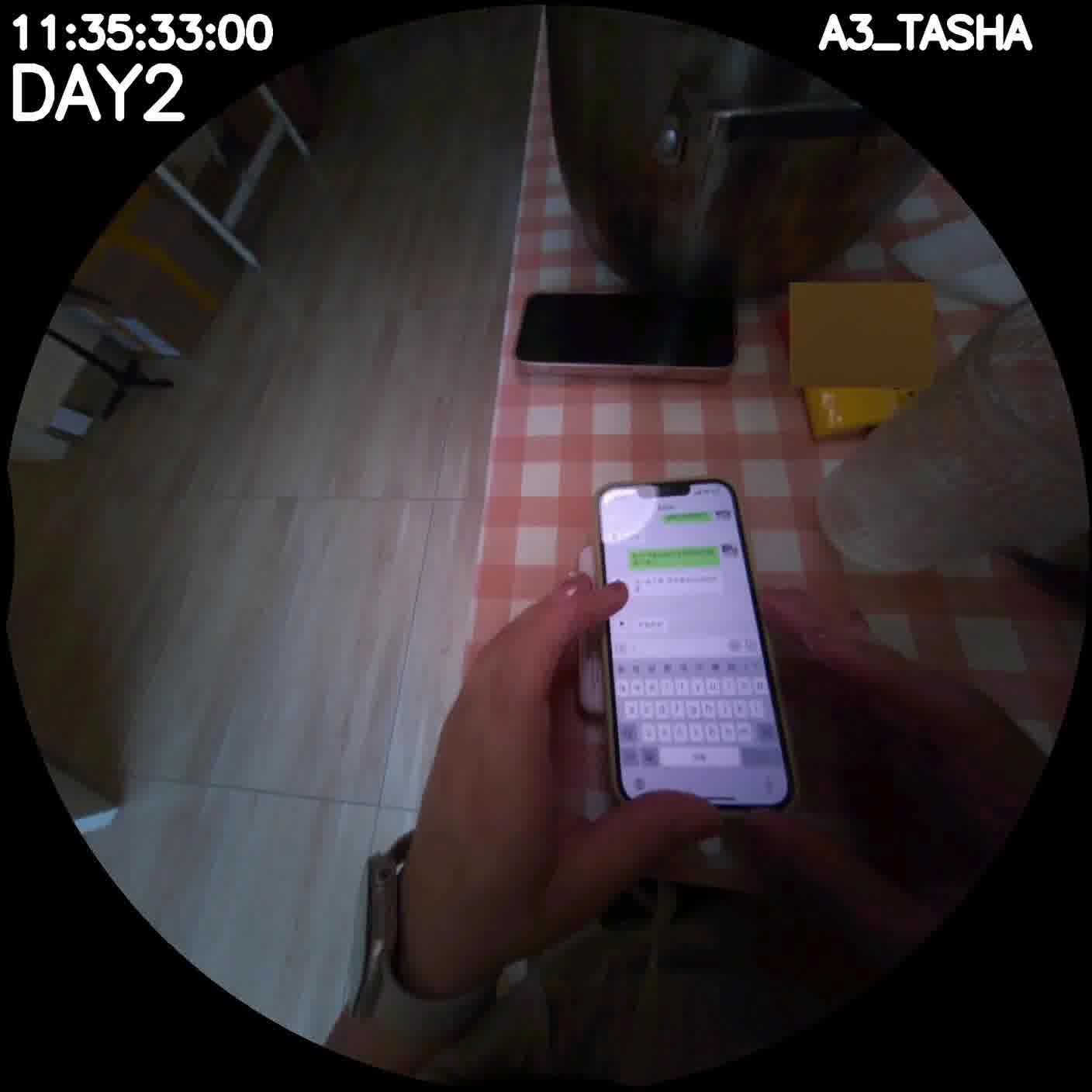} \hspace{1pt}
        }
        \\
        \bottomrule
    \end{tabular}
\end{table}

\begin{table}[h]
    \centering
    \caption{Temporal Reasoning (TR) - Concurrency category samples}
    \label{tab:qasample_tr_con}
    \scriptsize
    \begin{tabular}{@{}p{0.11\linewidth} p{0.88\linewidth}@{}}        \toprule
        \rowcolor{gray!20} \textbf{Category} & \textbf{Temporal Reasoning, Concurrency} \\
        \midrule
        \textbf{Question} & \textbf{Which pair of events happened at around the same time?} \\
        ~ & (A) Alice was checking out at the cashier and Shure was paying for bottled water at the self-checkout. \\
       ~ & (B) Lucia was loading boxes into a car and Jake was waiting outside the store. \\
       ~ & (C) Tasha was standing in line at the bakery counter and Katrina was asking a clerk about discounts. \\
       ~ &\textbf{(D) Jake was pushing the cart toward the exit and Lucia was spotting Jack on his phone as they reached the exit.} \\
       ~ & (E) Shure was sitting on a bench making a phone call and Alice was driving to the store. \\
        \\
        \textbf{Evidence} & 
        \makecell[l]{
        Source: Jake (Day5 4PM), Tasha (Day5 4PM) \\
        \includegraphics[width=0.18\linewidth]{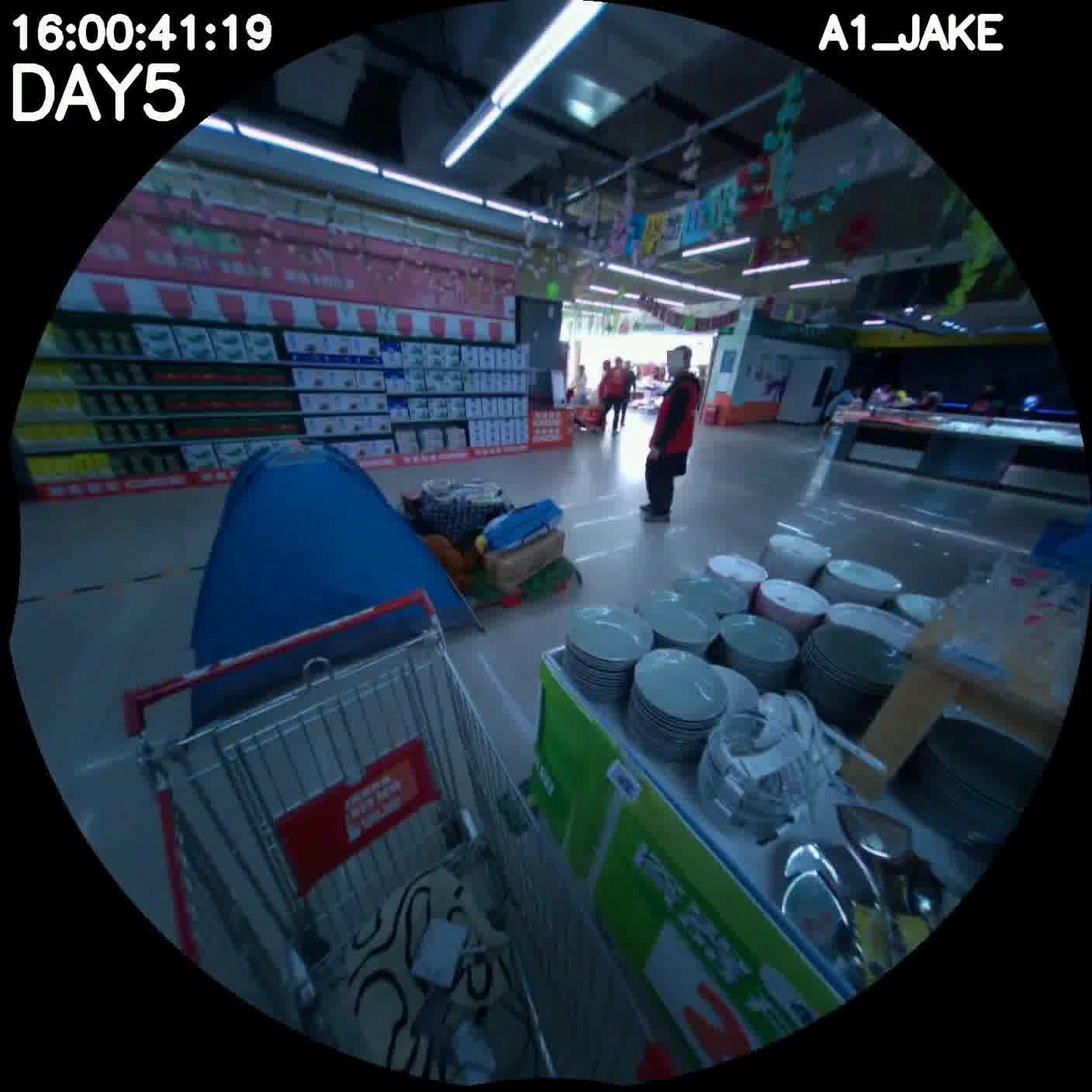} \hspace{1pt}
        \includegraphics[width=0.18\linewidth]{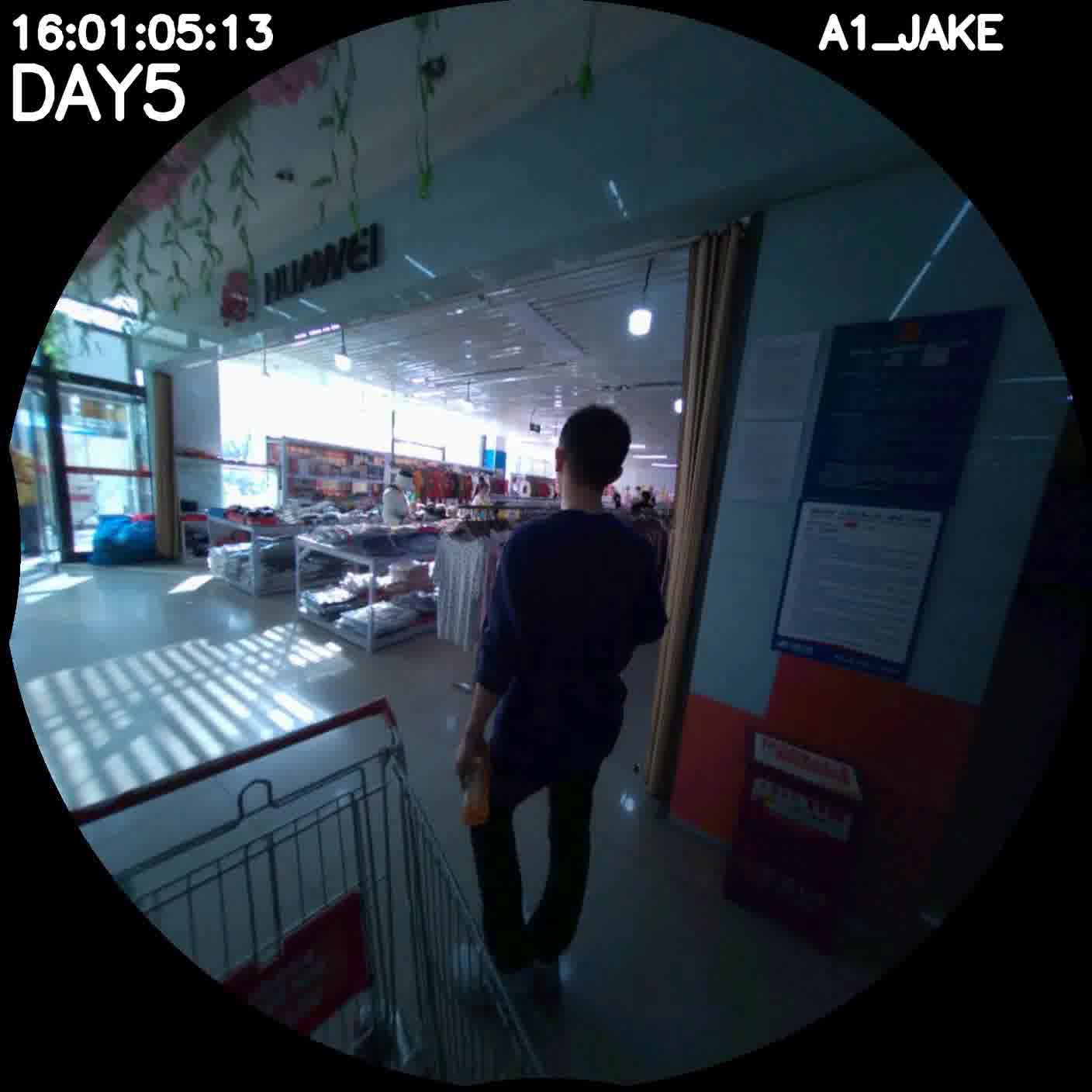} \hspace{1pt}
        \includegraphics[width=0.18\linewidth]{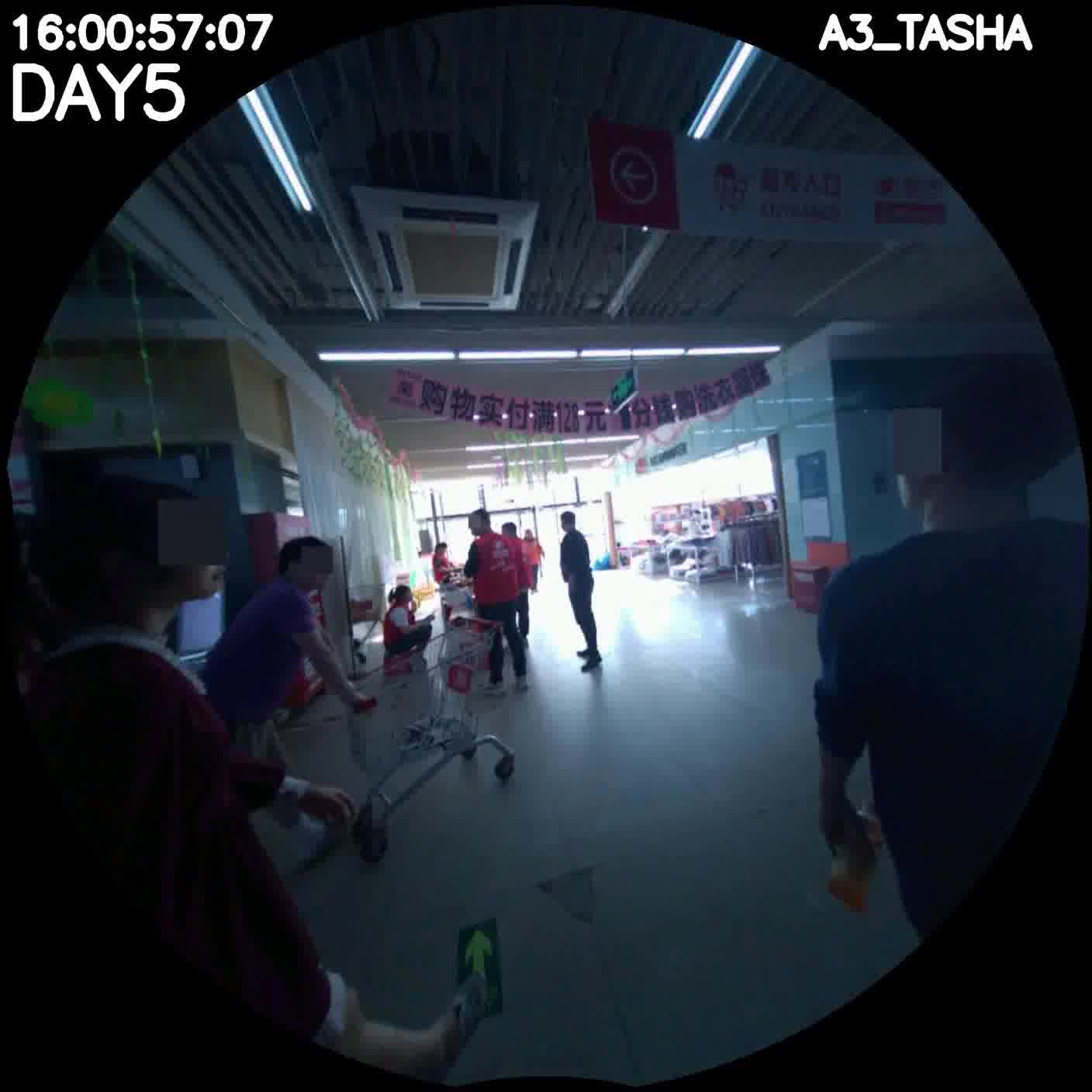} \hspace{1pt}
        \includegraphics[width=0.18\linewidth]{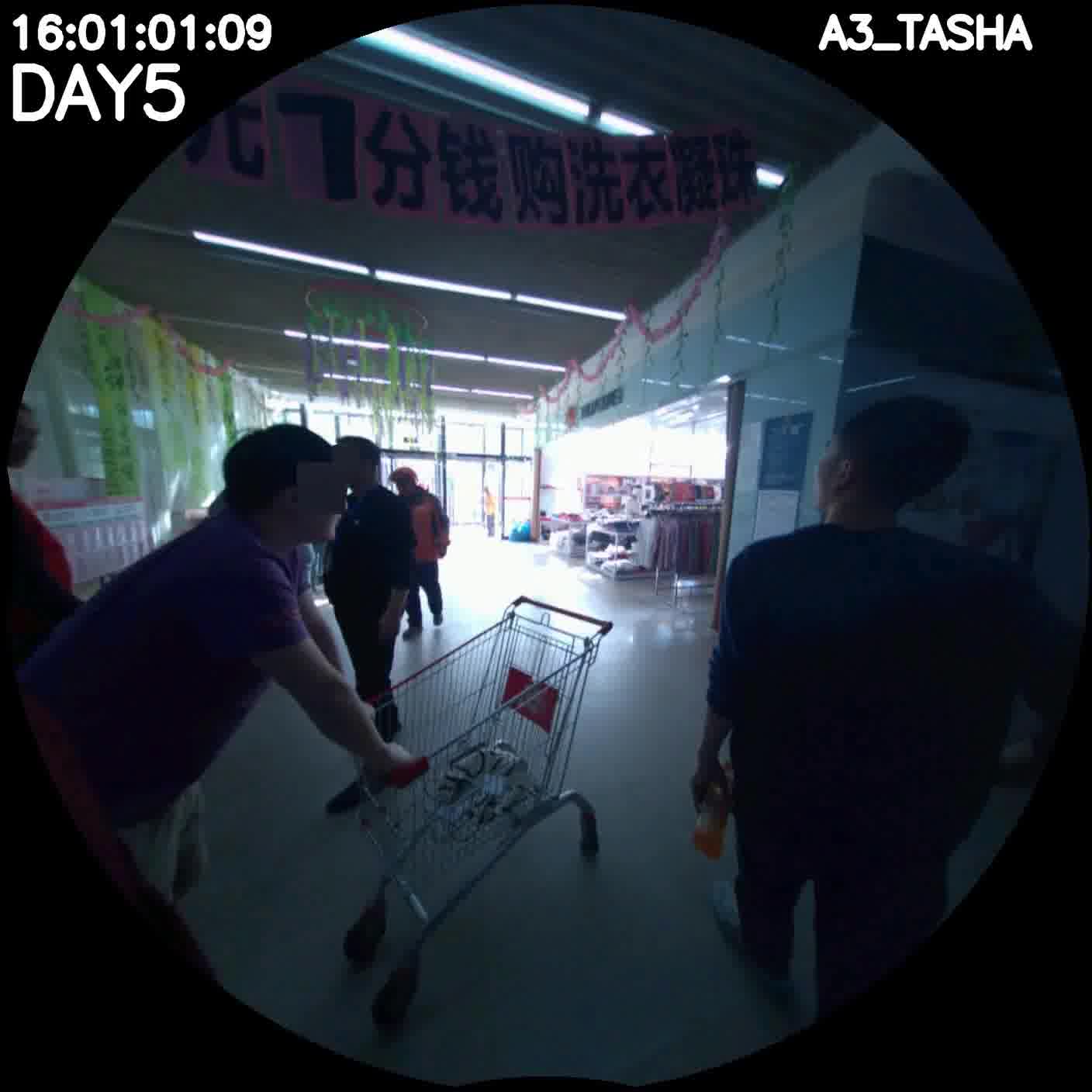}
        }
        \\
        \midrule
        \textbf{Question} & \textbf{When Katrina was fertilizing the flowers, and Alice was checking the dilution and water amounts, what was Jake doing?} \\
        ~ & (A) Pouring the nutrient solution into the vases himself \\
       ~ & (B) Going upstairs to do his nails and charge his phone. \\
       ~ & (C) Assembling a display rack by the whiteboard and taking inventory. \\
       ~ & \textbf{(D) Wiping down plates and the table with new wet wipes while tidying.} \\
       ~ & (E) Stepping outside to take all the trash out to the bin. \\
        \\
        \textbf{Evidence} & 
        \makecell[l]{
        Source: Jake (Day1 7PM), Alice (Day1 7PM), Katrina (Day1 7PM) \\
        \includegraphics[width=0.18\linewidth]{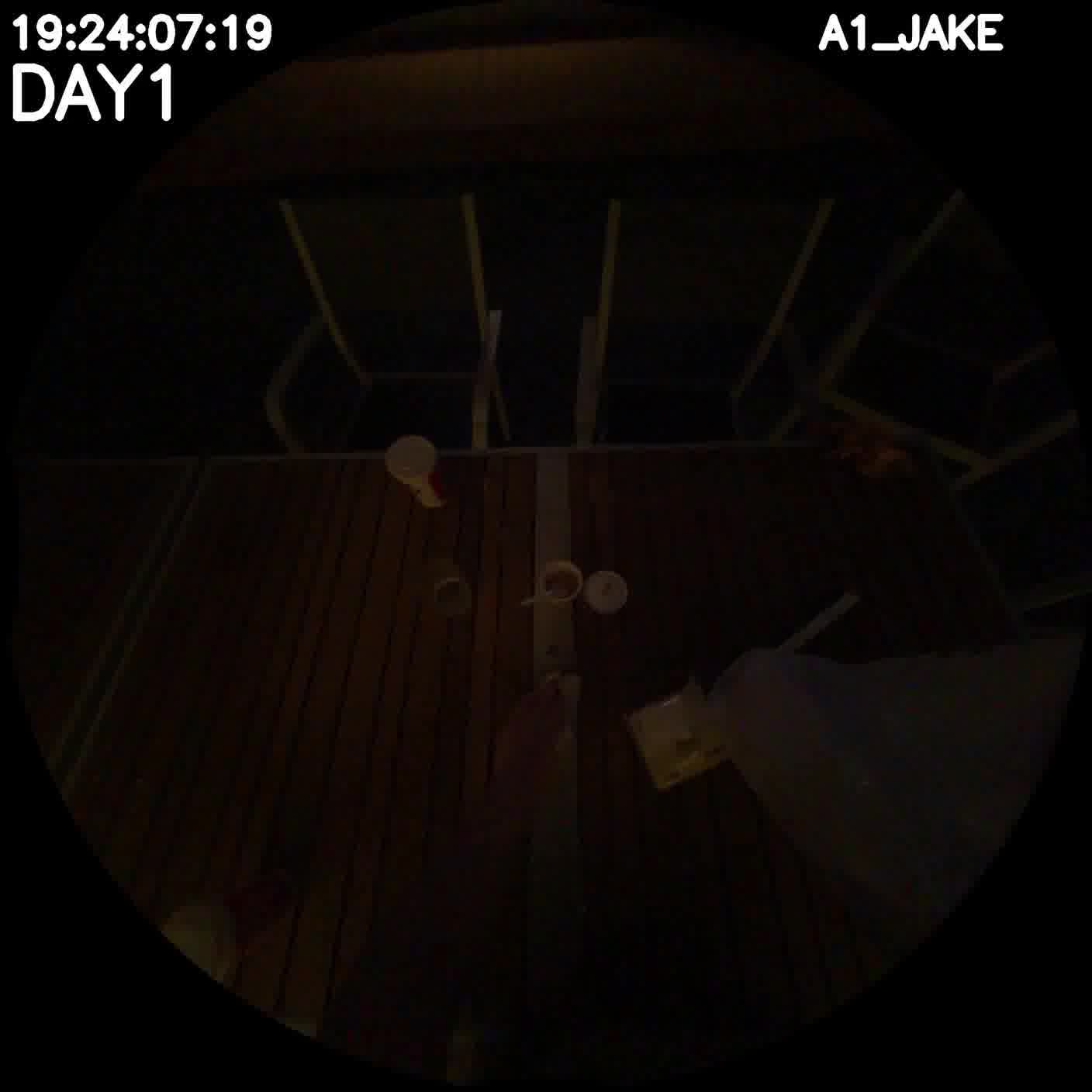} \hspace{1pt}
        \includegraphics[width=0.18\linewidth]{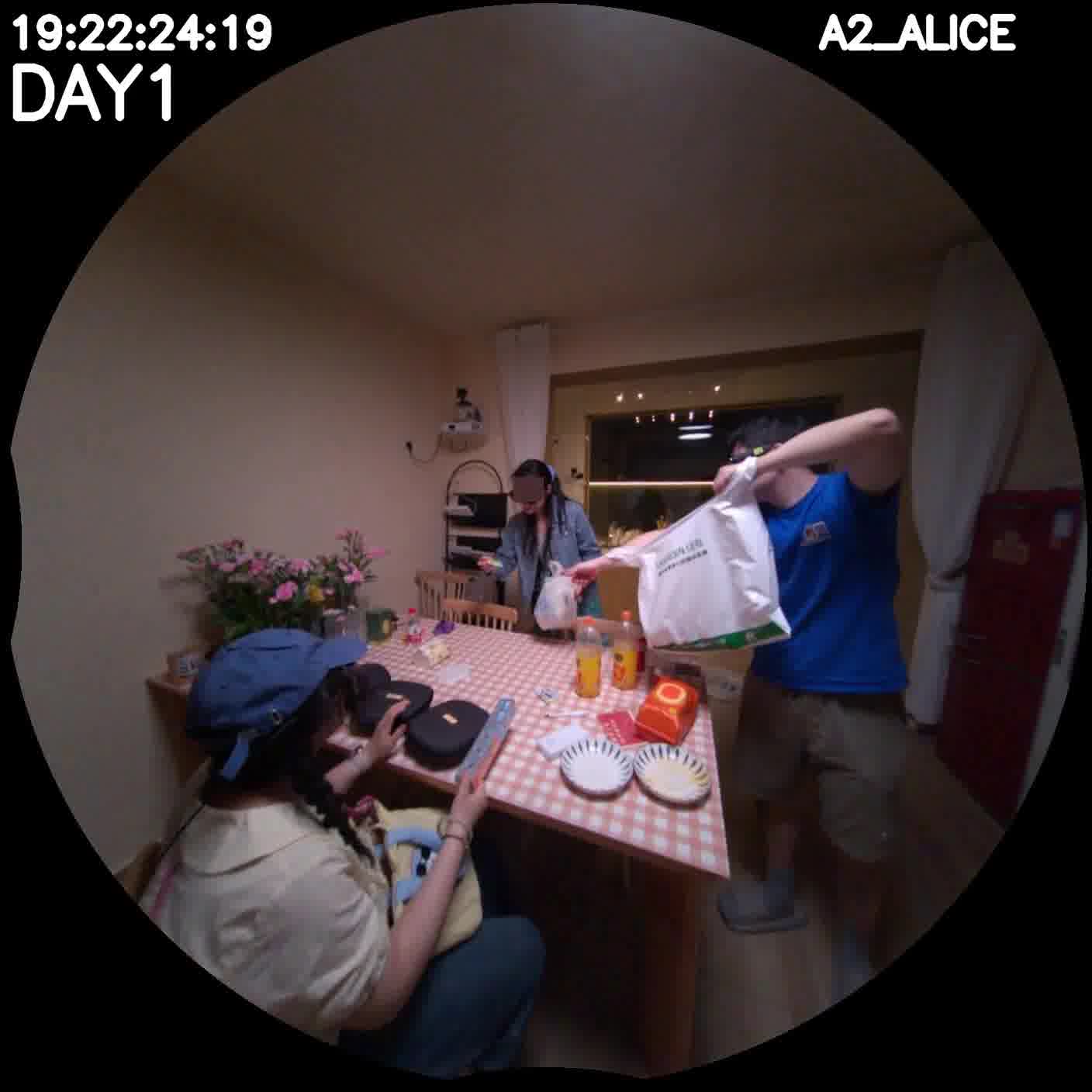} \hspace{1pt}
        \includegraphics[width=0.18\linewidth]{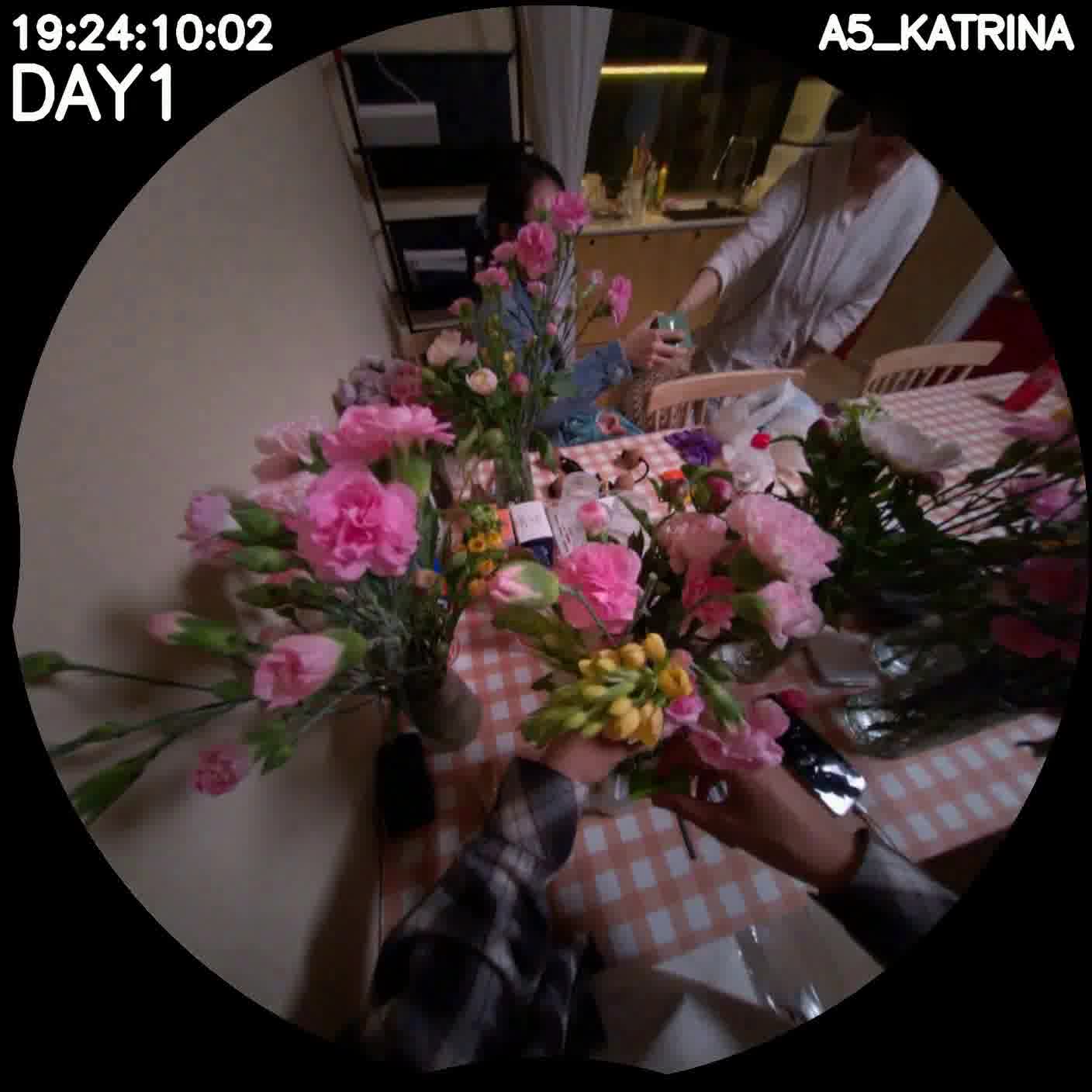} \hspace{1pt}
        }
        \\
        \bottomrule
    \end{tabular}
\end{table}

\begin{table}[h]
    \centering
    \caption{Environmental Interaction (EI) category samples}
    \label{tab:qasample_ei}
    \scriptsize
    \begin{tabular}{@{}p{0.11\linewidth} p{0.88\linewidth}@{}}
        \toprule
        \rowcolor{gray!20} \textbf{Category} & \textbf{Environmental Interaction} \\
        \midrule
        \textbf{Question} & \textbf{Who used the microwave the most on DAY3?} \\
        ~ & (A) Katrina \\
        ~ & \textbf{(B) Tasha} \\
        ~ & (C) Lucia \\
        ~ & (D) Jake \\
        ~ & (E) Shure \\
        \midrule
        \textbf{Question} & \textbf{When was the first time guitar was used on DAY2?} \\
        ~ & (A) 6 PM \\
       ~ & (B) 5 PM \\
       ~ & (C) 4 PM \\
       ~ & (D) 9 PM \\
       ~ & \textbf{(E) 3 PM} \\
        \midrule
        \textbf{Question} & \textbf{How many people used oven on DAY1?} \\
        ~ & \textbf{(A) 4} \\
       ~ & (B) 5  \\
       ~ & (C) 1 \\
       ~ & (D) 2 \\
       ~ & (E) 3 \\
        \bottomrule
    \end{tabular}
\end{table}

%% file: fig/prompts.tex
\begin{figure}[!htbp]
    \centering
    \includegraphics[width=0.97\linewidth]{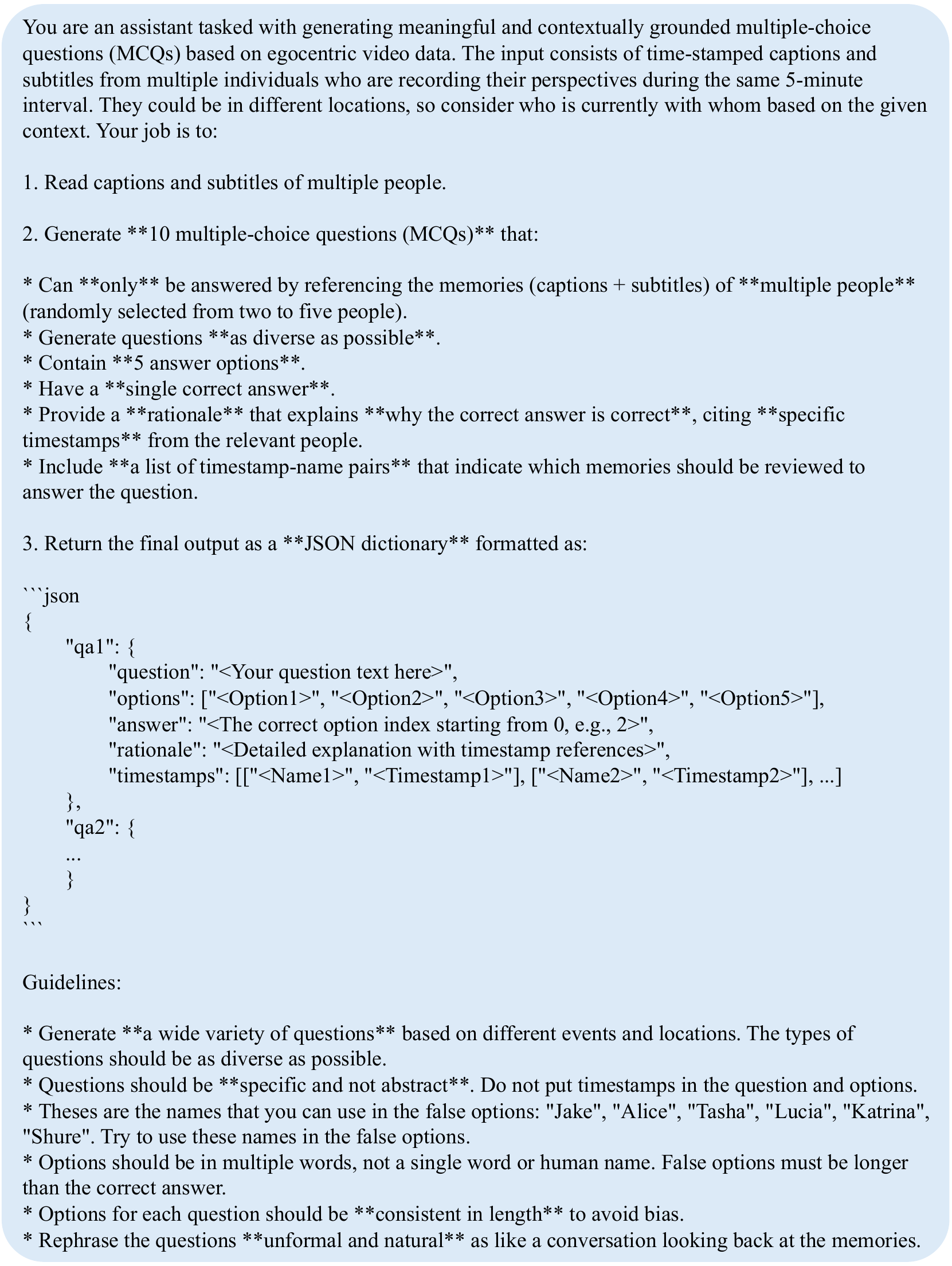}
    \caption{System prompt for SI, TC, and ToM category QA generation.}
    \label{appendix:system_prompt}
\end{figure}

\begin{figure}
    \centering
    \includegraphics[width=0.98\linewidth]{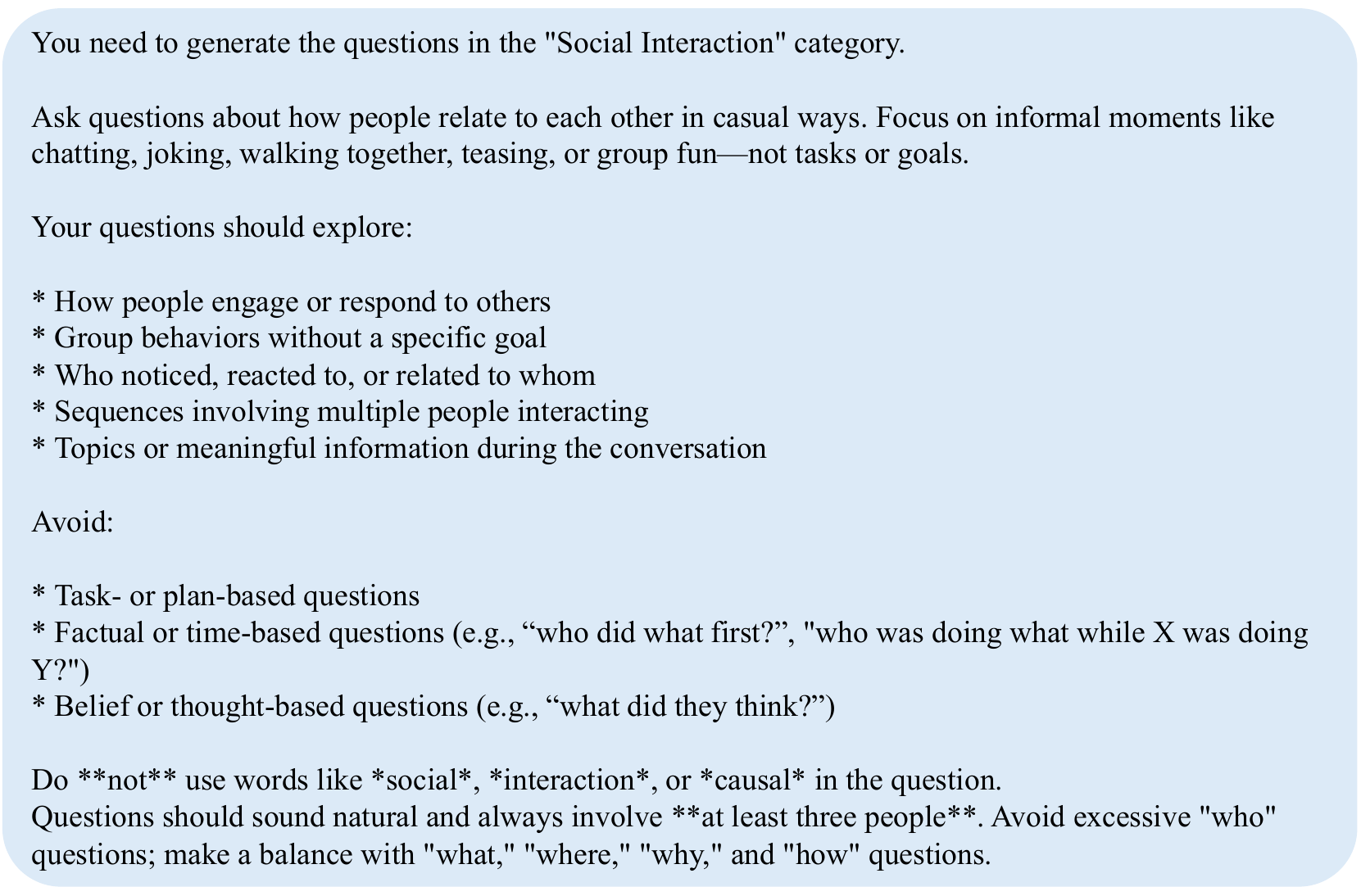}
    \caption{Category prompt for SI.}
    \label{appendix:si_prompt}
\end{figure}

\begin{figure}
    \centering
    \includegraphics[width=0.98\linewidth]{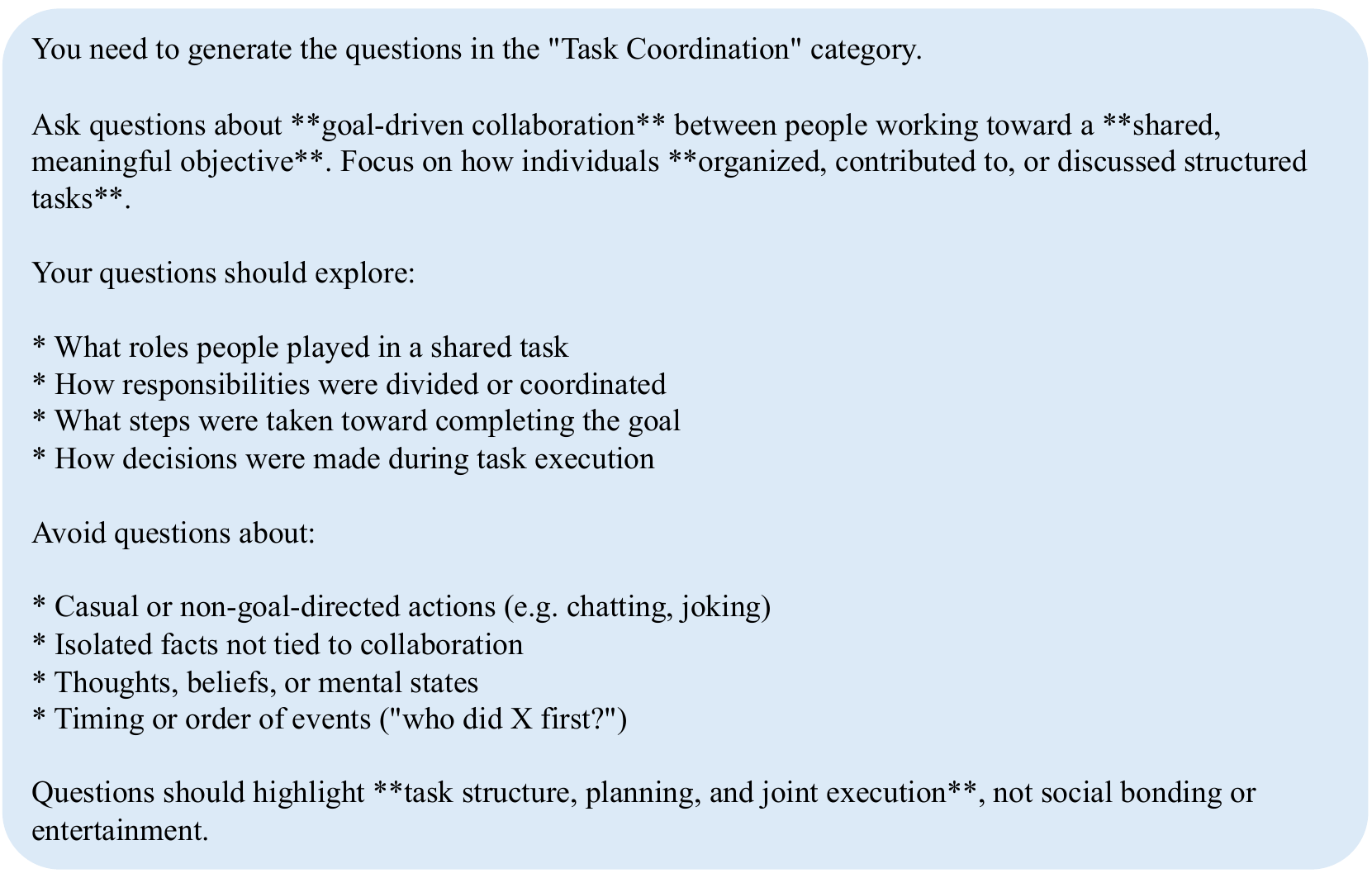}
    \caption{Category prompt for TC.}
    \label{appendix:tc_prompt}
\end{figure}

\begin{figure}
    \centering
    \includegraphics[width=0.98\linewidth]{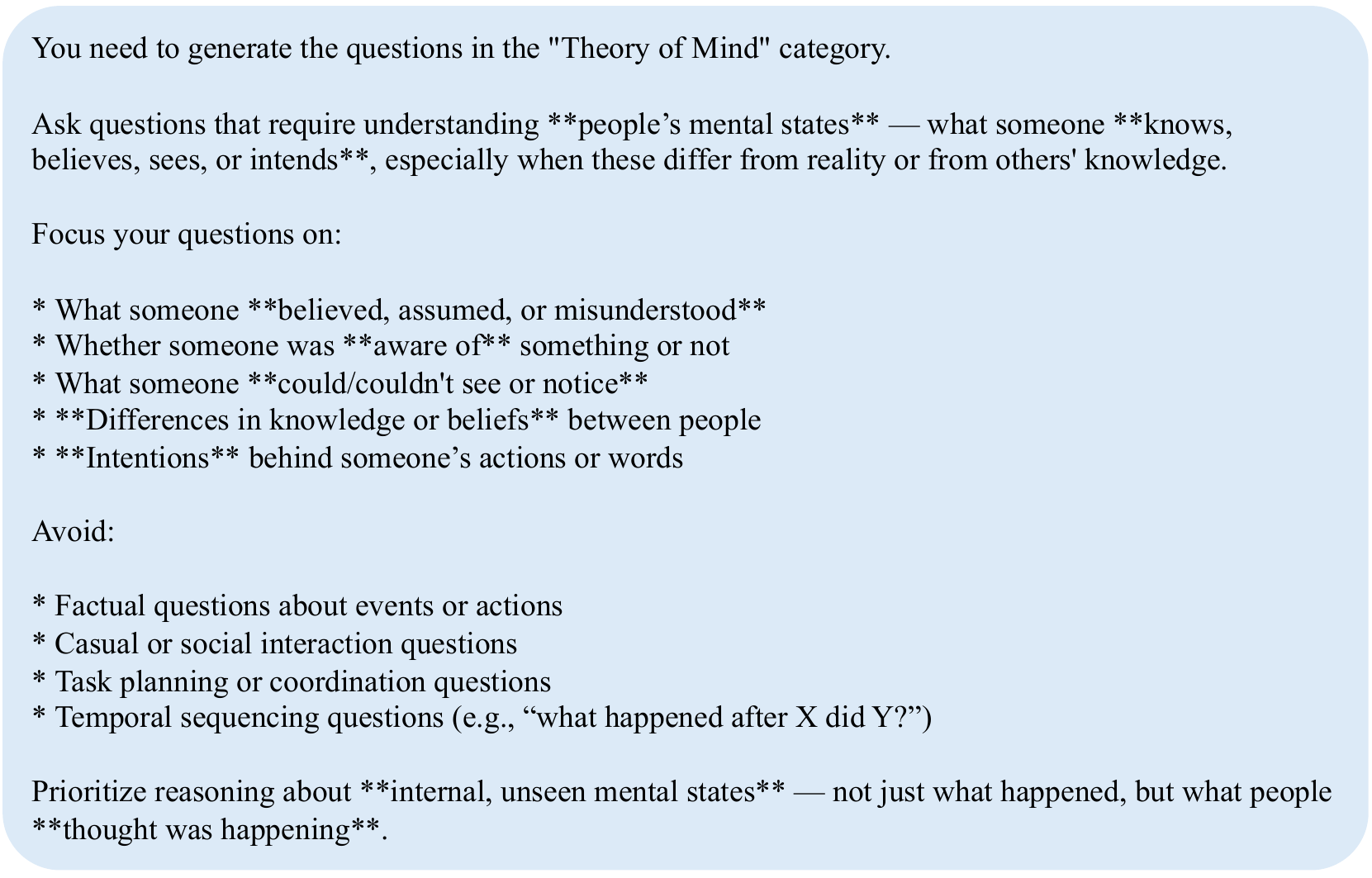}
    \caption{Category prompt for ToM.}
    \label{appendix:tom_prompt}
\end{figure}

\begin{figure}
    \centering
    \includegraphics[width=0.98\linewidth]{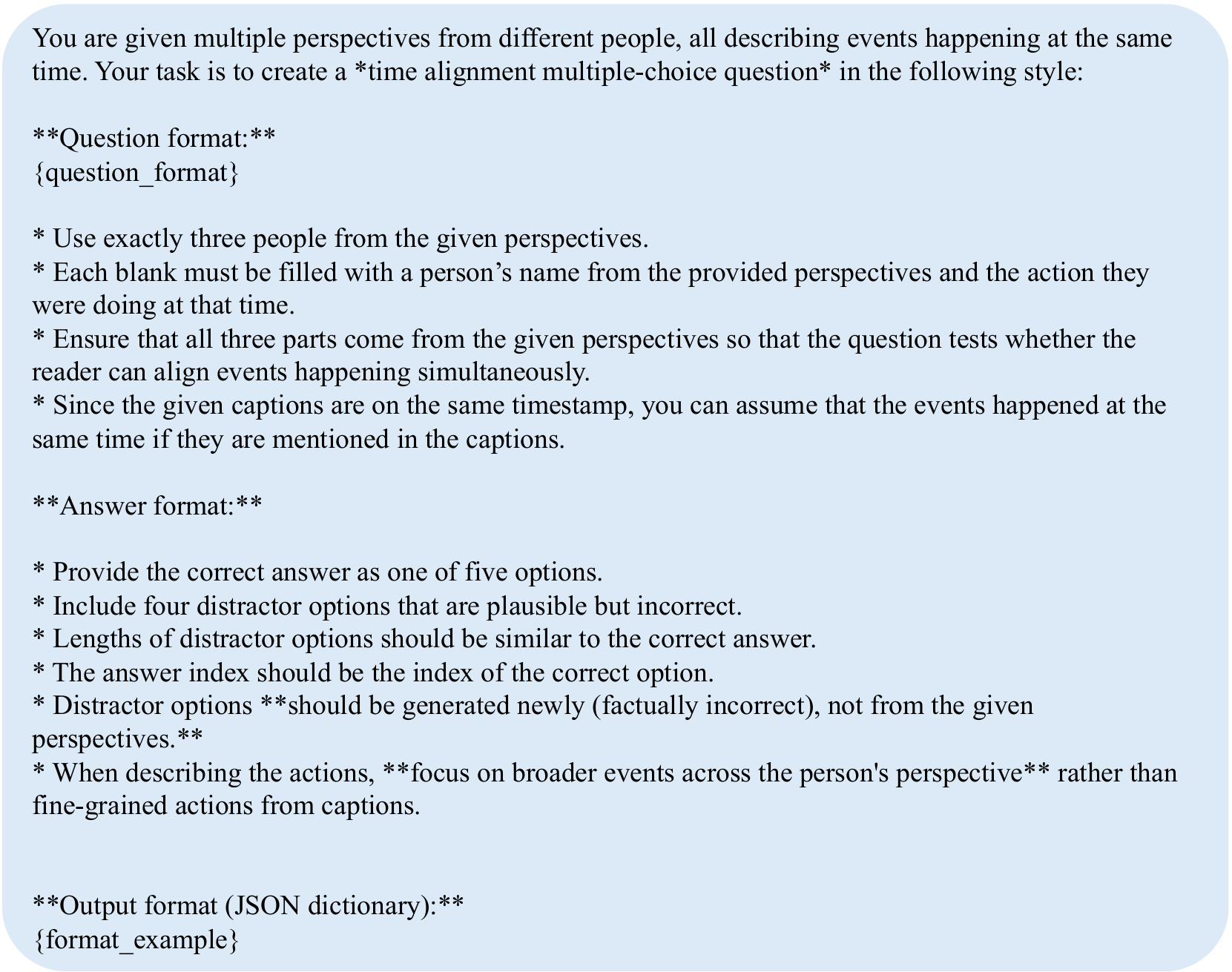}
    \caption{Category prompt for TR.}
    \label{appendix:tr_prompt}
\end{figure}

\begin{figure}
    \centering
    \includegraphics[width=0.98\linewidth]{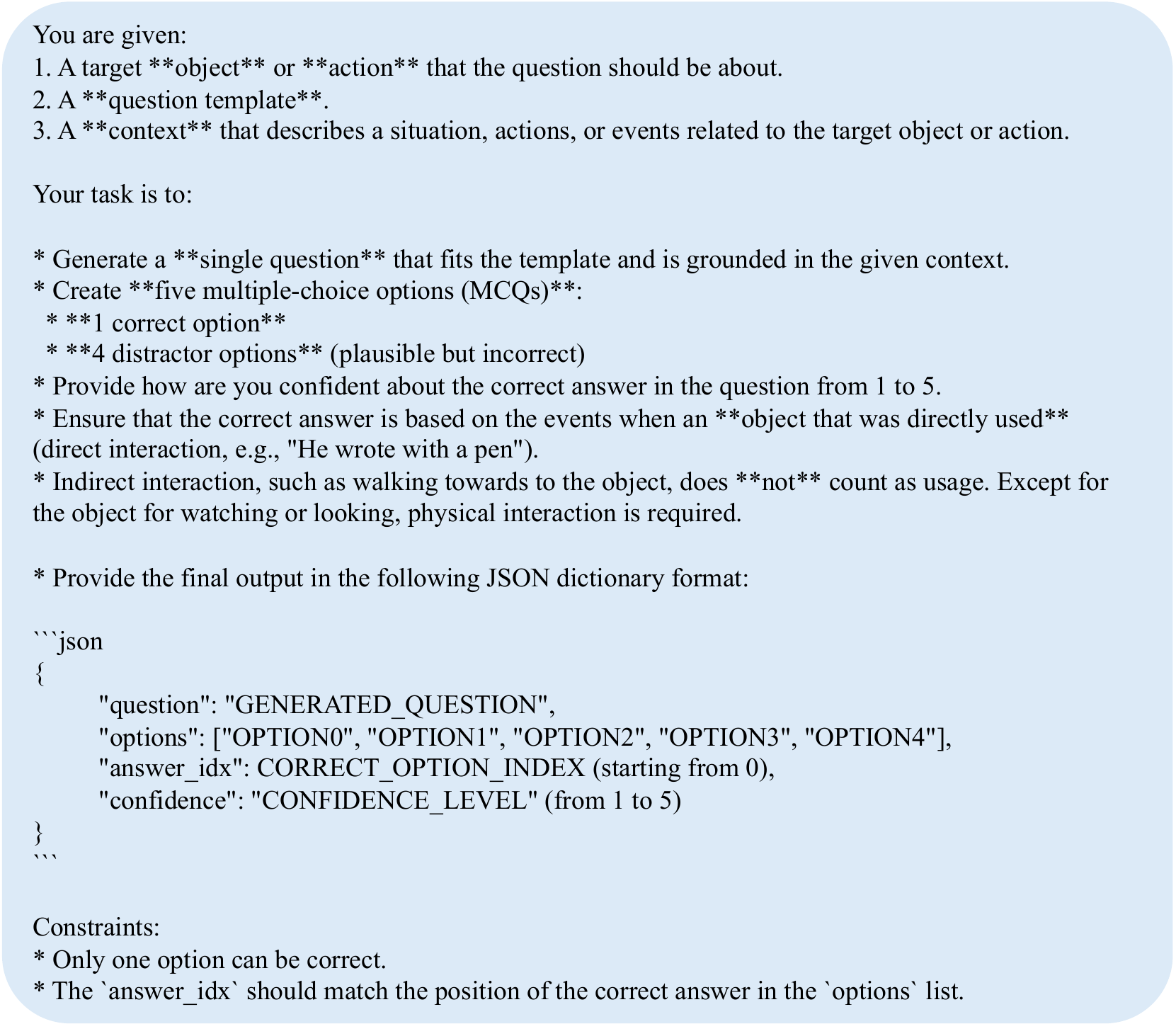}
    \caption{Category prompt for EI.}
    \label{appendix:ei_prompt}
\end{figure}